**Suez Canal University**
**Faculty of Engineering**
**Port Said, Egypt**

# *Combined Classifiers for Invariant Face Recognition*

**A Thesis**
submitted for partial fulfillment for
the requirements of the degree of
Master of Science in Electrical Engineering

**Submitted by**

**Eng.\ Ahmad Hosney Awad Eid**
B.Sc., Electrical Engineering, 1999
Computer Division
Faculty of Engineering
Suez Canal University

**2004**

**Suez Canal University**
**Faculty of Engineering**
**Port Said, Egypt**

# *Combined Classifiers for Invariant Face Recognition*

**A Thesis**
submitted for partial fulfillment for
the requirements of the degree of
Master of Science in Electrical Engineering

## Submitted by

**Eng.\ Ahmad Hosney Awad**
B.Sc., Electrical Engineering, 1999
Computer Division
Faculty of Engineering
Suez Canal University

## Supervised by

**Prof. Dr. Abd El Hay A. Sallam**
Dept. of Electrical Engineering
Faculty of Engineering
Suez Canal University

**Dr. Kamel A. El Serafi**
Dept. of Electrical Engineering
Faculty of Engineering
Suez Canal University

**2004**

**Suez Canal University**
**Faculty of Engineering**
**Port Said, Egypt**

# *Combined Classifiers for Invariant Face Recognition*

**A Thesis**
submitted for partial fulfillment for
the requirements of the degree of
Master of Science in Electrical Engineering

## Submitted by

**Eng.\ Ahmad Hosney Awad**
B.Sc., Electrical Engineering, 1999
Computer Division, Faculty of Engineering
Suez Canal University

## Approved by

**Prof. Dr. Hany Selim Girgis**
Prof. of Computer & Electronics Engineering
Faculty of Engineering
Assute University

**Prof. Dr. Nadder Hamdy Ali**
Head of Dept. of Electronics
Arab Academy for Science & Technology and Maritime Transport, Alexandria

**Prof. Dr. Abd El Hay Ahmad Sallam**
Prof. of Electrical Engineering
Faculty of Engineering
Suez Canal University

**2004**

| | |
|---|---|
| **Author** | **Ahmad Hosney Awad Eid** |
| **Title** | **Combined Classifiers for Invariant Face Recognition** |
| **Faculty** | **Engineering** |
| **Department** | **Electrical Engineering** |
| **Location** | **Port-Said** |
| **Degree** | **Master of Science** |
| **Date** | **30 / 6 / 2004** |
| **Language** | **English** |
| **Supervision Committee** | **Prof. Dr. Abd El Hay A. Sallam**<br>**Dr. Kamel A. El Serafi** |

## English Abstract

No single classifier can alone solve the complex problem of face recognition. Researchers found that combining some base classifiers usually enhances their recognition rate. The weaknesses of the base classifiers are reflected on the resulting combined system. In this work, a system for combining unstable, low performance classifiers is proposed. The system is applied to face images of 392 persons. The system shows remarkable stability and high recognition rate using a reduced number of parameters. The system illustrates the possibility of designing a combined system that benefits from the strengths of its base classifiers while avoiding many of their weaknesses.

# ACKNOWLEDGEMENT

I would like to express my gratitude to all the people who helped me complete this thesis. I offer my thanks to Prof. Dr. Abd El Hay Sallam for his kindness, patience and great support. I also would like to thank Prof. Dr. Ahmad Tolba for his generosity and his help that made this work possible in the first place. I would like to thank Dr. Kamel El Serafi for his continuing support and cooperation. Finally, I would like to thank my friend and colleague Eng. Adel El Sayed for his valuable remarks and pleasant discussions.

# SUMMARY

Among all the problems of pattern recognition, face recognition is one of the most difficult ones. The special nature of this problem required the researchers to investigate many classification approaches to solve this problem. No single classifier is able to perform equally well under all the various forms of face recognition applications. The only proof of the existence of such a classifier is the human brain with its enormous recognition capabilities. Since the idea of combining multiple classifiers appeared, it triggered a huge number of attempts to apply it to many pattern recognition problems. In this technique, a number of base classifiers are separately trained on the problem and their decisions are then combined using some combination strategy. Although there is no combined system able to remove all the difficulties of the face recognition problem, the combined classifiers techniques proved to possess very interesting attributes that can eventually remove many obstacles in the way of obtaining a good solution for such a complex problem. Nevertheless, the weaknesses of the base classifiers are reflected on the final combined systems. Base classifiers that are complex to design, possess low recognition rate, or have low stability can greatly affect the complexity, performance, and stability of the resulting combined system. In this work, a system for combining unstable, low performance classifiers is proposed. The system is applied to classify face images from a face database containing 392 persons. The proposed system shows remarkable stability and a high recognition rate using a reduced number of design parameters. The system is better than many of the combined classifiers systems reported in literature in its simplicity, stability, recognition rate, and scaling with increased input size. The proposed system can be implemented on an ordinary PC and is suitable for multimedia human-computer interaction applications. The proposed system can also be implemented using parallel processing techniques to handle a large number of persons. This work illustrates the possibility of designing a combined system that benefits from the strengths of its base classifiers while effectively avoiding their weaknesses.

# Table of Contents

# List of Tables

# List of Figures

# List of Abbreviations

| | |
|---|---|
| 2D: | Two-Dimensional |
| 3D: | Three-Dimensional |
| BKS: | Behavior-Knowledge Space |
| CLVQ: | Controlled Learning Vector Quantization |
| CMV: | Constrained Majority Voting |
| DT: | Decision Tree |
| ECOC: | Error Correcting Output Coding |
| FEC: | Front-End Classifier |
| gcd: | Greatest Common Divisor |
| HMM: | Hidden Markov Models |
| LDA: | Linear Discriminant Analysis |
| LVQ: | Learning Vector Quantization |
| MLP: | Multi-Layer Perceptrons |
| NP hard: | Non-Polynomial hard |
| OC: | Output Coding |
| OCR: | Optical Character Recognition |
| ORL: | Olivetti Research Laboratory |
| PC: | Personal Computer |
| PCA: | Principle Component Analysis |
| RBF: | Radial-Basis Functions |
| SFS: | Shape From Shading |
| SVM: | Support Vector Machine |

# *Chapter 1*
# Introduction

One of Allah's most precious gifts to humanity is the human brain. The pattern classification capabilities of the human brain are not only fascinating but also hard to understand by modern science. Among these capabilities, our ability to recognize each other from our faces has caught the attention of scientists and engineers for a long time. Adding this recognition capability to the ability of computers to store and exchange huge amounts of data all over the world will result in a revolution in the way humans interact with computers. Many algorithms, therefore, appeared during the last two decades for automatic face recognition. Using combinations of such algorithms seem to produce better solutions to this complex problem in various face recognition applications. The main difficulty with this approach is that the resulting combined system will inherit some of the weaknesses of the underlying algorithms. For instance if the combined algorithms are not stable enough the combined system may suffer from such instability to some degree. Another difficulty is the lack of a unified theory for combining such algorithms. This results in many experimental combination techniques many of which are very complex. A combination algorithm that needs many parameters to be selected in advance will be annoying to any face recognition systems designer.

## *1.1 Face Recognition using Combinations of Classifiers*

Face recognition is a special branch of biometrics. In biometrics the aim is at identifying a human individual among a human population through one or more biological metrics (hence the name biometrics). As described in [1], an ideal biometric system should have the following attributes:

- All members of the population possess the characteristic that the biometric identifies.
- Each biometric signature differs from all others in the controlled population.
- The biometric signatures do not vary under the conditions in which they are collected.
- The system resists countermeasures.

There are two types of biometric systems: identification systems and verification systems. The main differences as described in [1] are that in identification systems, a biometric signature of an unknown person is presented to a system. The system compares the new biometric signature with a database of biometric signatures of known individuals. Based on the comparison, the system reports (or estimates) the identity of the unknown person from this database. In verification systems, a user presents a biometric signature and a claim that a particular identity belongs to the biometric signature. The algorithm either accepts or rejects the claim. Alternatively, the algorithm can return a confidence measurement of the claim's validity. In this work, the term *face recognition* is used mainly to refer to face identification.

### *1.1.1 The Face Recognition System*

Having a population of humans (persons), one or more images are taken for each of their faces. These images are called training images. It is desired to design and implement a machine capable of classifying a new face image by assigning a label representing one of the humans in the population. The phase in which signatures are extracted from training images and encoded into the system is called the training phase. The phase in which a signature is extracted from a new image and compared to the encoded signatures is called the classification phase. Usually, time is not a critical factor during the training phase. On the other hand, many applications require the classification phase to be in real time (online face recognition) so; the classification algorithm should be as fast as possible. In addition, storage requirements for both training and classification should not be too much for a computer to handle. Many present face recognition systems rely on common PCs (Personal Computers) found in any computer store. The system should ideally classify all new face images correctly. A new image should be correctly classified as either belonging to some person, or not belonging to the population on which the system was trained. An error occurs when the system falsely accepts an image as belonging to the population when it belongs to a person on which the system was not trained at all. This error is called a false-acceptance error. Another type of error occurs when the system falsely rejects an image when it actually represents some person in the population. This is called a false-rejection error. In an ideal system, both of these errors should not occur at all. Figure 1.1 shows the general layout of a typical face recognition system during classification. The environment surrounding a face recognition system can cover a wide spectrum from a well-controlled environment to an uncontrolled one. In a controlled environment, frontal and profile photographs of human faces are taken complete with a uniform background and identical poses among the participants. These face images are commonly called mug shots. Each mug shot can be cropped (manually or automatically) to extract a normalized subpart called a canonical face image, as shown in figure 1.2. In a canonical face image, the size and position of the face are normalized approximately to the predefined values and the background region is minimal. Face recognition techniques for canonical images have been successfully applied to many face recognition systems.

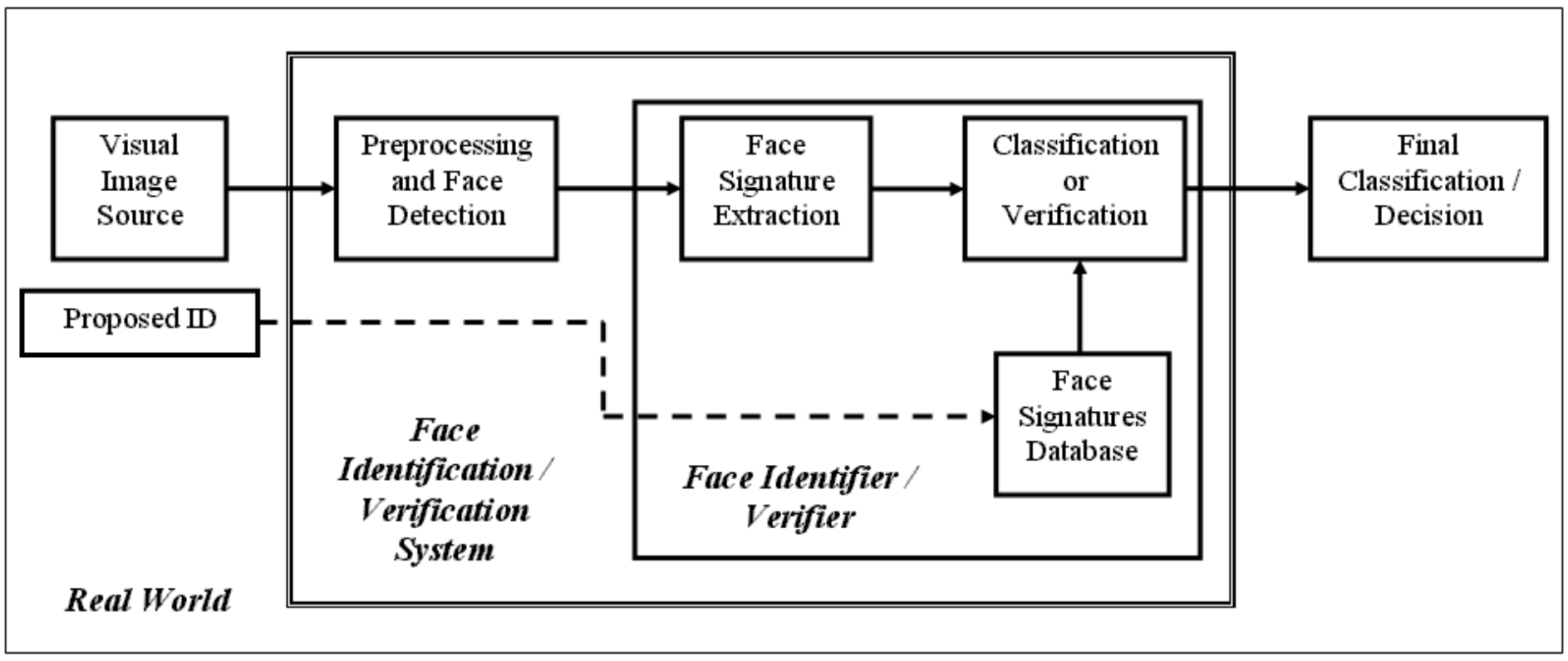

Figure 1.1. The general layout of a typical face recognition system during classification

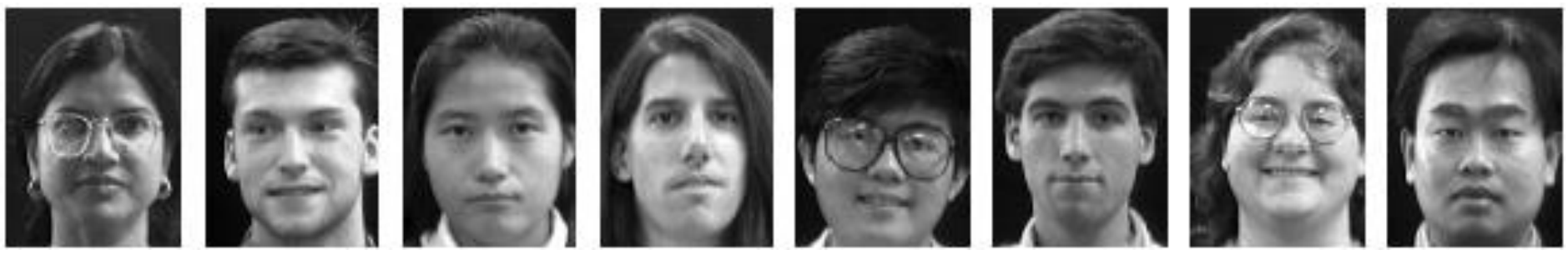

Figure 1.2. Some examples of canonical face images

The visual image source is simply an image or sequence of images containing one or more human faces. Depending on the application, the visual data may be a video containing a moving person or persons (like in a supermarket surveillance camera), a single photographic image of a person (like in the systems the police use to identify people from mug shots), an infra-red image of a person (figure 1.3) [2], or even range images holding 3D information about the face (figure 1.4) [3]. If the system is a verification system, a proposed identity is presented to the system to be verified or rejected. These are the main inputs to a biometric face-based system. Inside the system itself, the visual data are preprocessed to be suited for the recognition / verification process (for example sampled and digitized for a digital-computer based face recognition system).

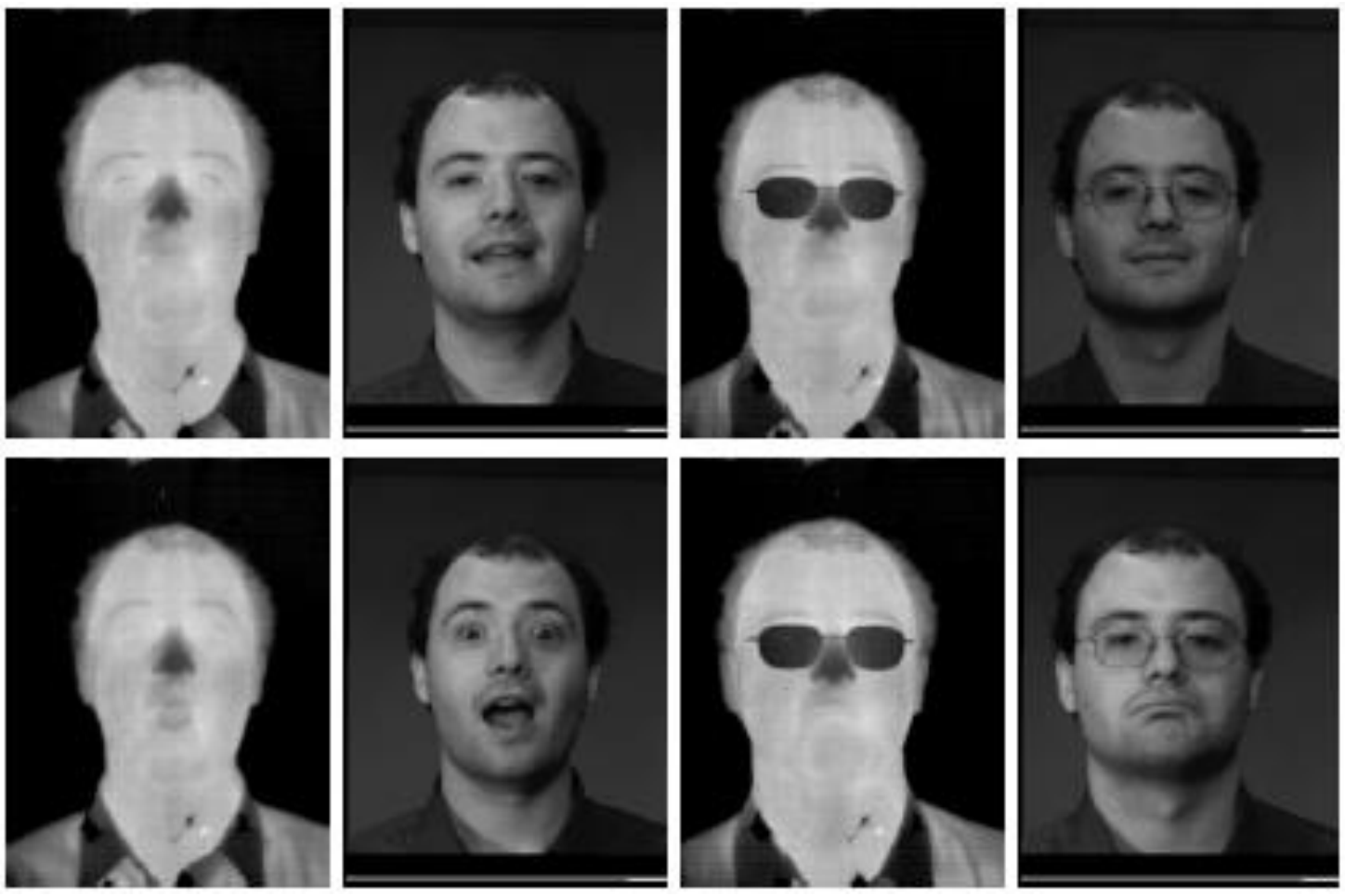

Figure 1.3. Examples of infrared images taken for a person

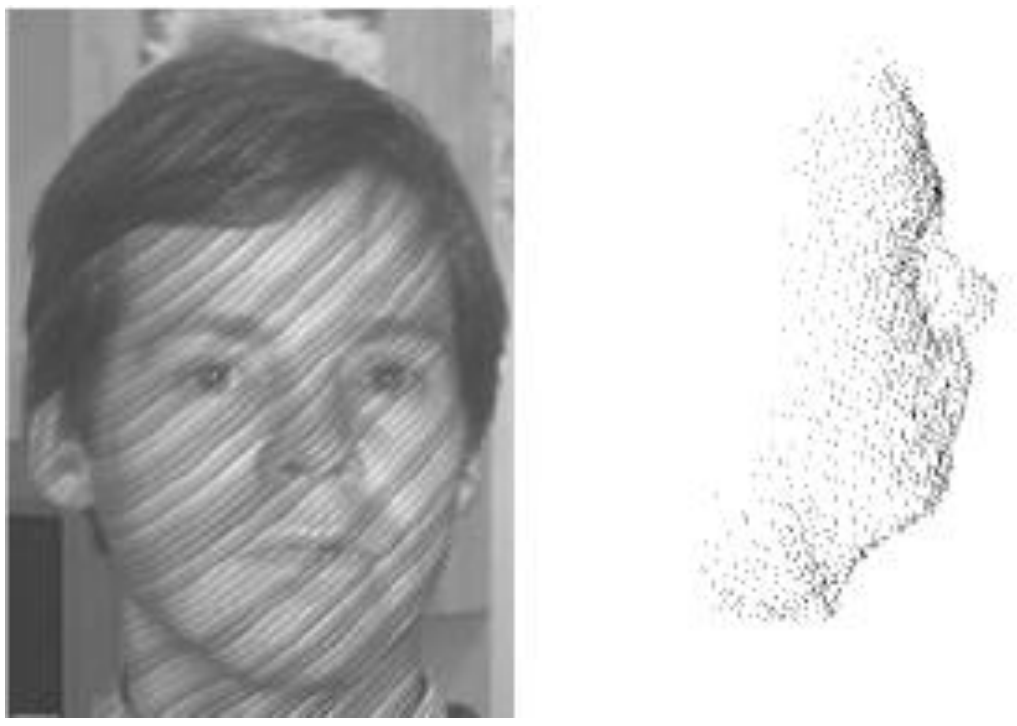

Figure 1.4. Structured light acquisition system and the constructed 3D model

If the visual data contain multiple faces or contain a single face that can be anywhere in the image area, a face detection phase is necessary to focus on the actual biometric face data and ignore the background of the captured scene. By this point the system has an image of a human face (or a sequence of images of one or more human faces) having most of its data content relevant to the recognition / verification problem. Other preprocessing may be then required to compensate for some undesirable effects like noise from cameras or even some illumination problems [4].

Often a process called normalization (consisting of scaling, rotation, and other operations) is also needed to present the face data in a suitable standard form to the next phase. Next, a biometric signature is extracted from the data and compared with other known biometric signatures for known people already stored in the system. In a recognition system, the extracted signature is compared against all the signatures of the database to find a number of people identities that mostly resembles the extracted signature. In verification systems, the extracted signature is compared to only one signature from the database to verify or reject its proposed identity. The final output of the system is then different between identification and verification systems. In identification systems, the system gives a list of possible matches where in verification systems the system gives a single number expressing the validity of the given claim.

In order to design a computer algorithm capable of performing face recognition, the designer must first decide the following:

- The method of extracting signatures from the training images and from the new image to be classified.
- The form in which the signatures of the training images are stored or encoded into the system.
- The algorithm to be used to store or encode the signatures of the training images into the system.
- The algorithm used to compare the signature of a new image with the signatures already stored or encoded into the system.

### *1.1.2 Face Recognition Applications*

Applications of face recognition are very diverse. Face recognition can be used as an access control method, as if it was a biometric password, to protect sensitive entities (like information, money, military facilities, and the like). Such a system must be very accurate and must have a false-acceptance rate near zero. Such systems usually operate inside controlled environments. Nevertheless, the high accuracy demand renders them very expensive. Another important application is to identify criminals or lost people from their photos. This application also requires a relatively accurate system. Unfortunately, the input images might be taken under random conditions like airport surveillance cameras, outside buildings under open sky, from un-normalized angles, and so on. Other applications include multimedia applications such as video games and other interactive entertainment applications. In this type of applications, the system is required to deal with the user / customer based on his / her identity. Accuracy is not critical in such applications. Thus, inexpensive and simple to use face recognition systems are suitable. Another application might be software protection against piracy. The programmer could protect his / her software by allowing a restricted number of users to use the software using their faces as identifiers. Any other people trying to execute the software will not be able to. This type of applications requires a fast online face recognition system that is both simple to use and inexpensive. Generally speaking, there is no single face recognition system that can be universally used in all kinds of applications. Every category of applications will require a suitable face recognition system. The system presented in this work requires a relatively controlled environment. It is very stable and relatively fast even for a large population of people (around 400 persons). Moreover, it can be used on an ordinary PC. It is not suitable for protection of critical entities. However, it could be very useful to multimedia applications, software protection applications, and the like.

### *1.1.3 Combining Classifiers*

To identify a face is to classify a face image to be belonging to some specific person. For such a task to be performed, a classifier is needed that accepts the face image and produces the classification decision. The percentage of correctly classified face images is called the recognition rate of the classifier and is the main performance measure for any classifier. In the past two decades, many classifiers were designed to be used in face recognition applications. No single classifier was ever found to be suitable for all applications. Some applications acquire face images in a controlled environment in which lighting, pose, facial expression and other variables are strictly controlled. Other applications obtain face images in the form of a video taken in the outside world for a passing person. Even in a single application, a classifier may be confused about classifying some persons more than others. Researchers found combining a number of classifiers helpful for reducing such disadvantages. Since the idea of combining classifiers appeared, it was used to solve many problems in pattern recognition. As stated in [5], with these combination methods the focus of pattern recognition shifted from the competition among classification approaches to the integration of such approaches as potential contributing components in a combined system. The classifiers to be combined are called the base classifiers or the component classifiers. Each classifier produces a classification decision similar to or different from other base classifiers for a given input. These decisions are combined using a combination method to produce the final decision. In many cases, the recognition rate of the combined system is higher than the recognition rate of the best component classifier.

### *1.1.4 The Difficulties of Combining Classifiers*

Although combining classifiers improves the overall recognition rate in many cases, this approach faces some difficulties. These difficulties are mainly results of the combination process itself. The weaknesses and faults of the base classifiers are sometimes magnified in the combined system. For instance if the recognition rates of the base classifiers are too low, the recognition rate of the combined system may not be sufficiently high to be useful. Another example is when using complex base classifiers. If the base classifiers are hard to design or require many parameters to be set before training, the combined system will inherit such complexity in a multiplied form. This may render the combined system to be impractical. A third example is when the base classifiers are unstable; in this case, the combined system may not be stable enough to be reliable. A classifier is said to be stable if its recognition performance on some dataset is not too sensitive to its initial conditions and parameters. The performance of a stable classifier should not be too sensitive to parameters such as number of training iterations, the order of presenting training inputs, exact details of its underlying structure. A typical example of unstable classifiers is an LVQ neural network. To train an LVQ net some parameters are required to be set in advance. For instance the number of hidden units, the maximum number of training iterations, the value of the learning rate, the initial values of the network weights and the order at which the inputs are presented during training. Any small change in such parameters may result in a large variation in the final recognition rate after training is completed. Such instability is completely undesirable in any practical face recognition system.

### *1.2 The Problem Statement*

If a combination of a set of classifiers is used to solve a face recognition problem, the designer will face some or all of the difficulties in section 1.1.4. If the designer decides to use unstable classifiers, such as LVQ neural networks, for their desirable properties like efficiency and simplicity, the weaknesses of such classifiers must be prevented from affecting the final combined system. Sine the base classifiers are unstable, the recognition rate of each classifier can be high or low depending on its selected parameters. In other words, the instability of the base classifiers results in two problems in the final combined system:

- The combined system may become unstable.
- The recognition rate of the combined system may become less than expected.

These two problems must be solved in order to obtain a useable combined classifiers system for face recognition that relies on unstable base classifiers.

### *1.3 Objectives*

The objectives of this work can be stated as follows:

- To design a combined classifiers system capable of performing face recognition with a high recognition rate.
- To illustrate the possibility of designing a combined classifiers system that is both simple and effective.
- To illustrate the possibility of designing a highly stable combined classifiers system based on unstable base classifiers

In other words, the main objective is to illustrate the ability of designing a combined system that benefits from the strengths of its base classifiers while avoiding many of their weaknesses.

### *1.4 Contents*

This thesis consists of seven chapters organized as follows:

Chapter 1: An introduction to face recognition using systems of combined classifiers.

Chapter 2: Literature review on the various techniques used to design face recognition systems with the focus on combined systems.

Chapter 3: Introduces the problem of pattern recognition, its main concepts and terms, and the difficulties of solving its sub-problems.

Chapter 4: Focuses on the subject of combining multiple classifiers to solve pattern recognition problems. It illustrates the importance of this subject and some of the many methods of combining classifiers.

Chapter 5: Presents the proposed combined system for face recognition.

Chapter 6: The results of the proposed system are presented along with a discussion of these results.

Chapter 7: The conclusions and future work are presented.

*Chapter 2*

# Literature Review

In order to illustrate the use of combined systems in face recognition, the use of single classifier systems is first presented in section 2.1. The intension is to illustrate the advantages and disadvantages of the classifiers that act as base classifiers in combined systems. In section 2.2, some of the attempts to apply combined systems to face recognition are presented along with their apparent disadvantages. The conclusions are then presented in section 2.3.

## *2.1 Single Classifier Systems*

Face recognition techniques based on a single classifier can be divided into two main categories: Statistical and Non-statistical approaches. Statistical approaches include Eigenfaces ([6], [7], and [8]), Hidden Markov Models (HMM) ([9], [10], [11], and [12]), and Fisherfaces [13]. Non-statistical approaches include Neural Networks ([7], [14], [15] and [16]), Elastic Matching [17], and many others. Some of the previous techniques are biologically inspired techniques [18]. In what follows a general idea is presented about the more popular techniques along with some comparison results reported in the literature.

### *2.1.1 Statistical Approaches*

This type of approach originated from an image representation task where a face image is treated as a high dimensional vector, each pixel being mapped to a component in that vector. The Karhunen-Loeve projection is used on the corresponding vector space for face image characterization. The idea of representing the intensity image of a face by a linear combination of the principle component vectors can also be used for recognition. This technique relies on what is called principle component analysis (PCA). Turk and Pentland used this technique for face recognition problem [6]. Using this image vector representation, the Linear Discriminant Analysis (LDA) has been independently used for face recognition by several research groups, including [19], [20], and [21], among many other groups. Such statistical methods derive features directly from intensity images, using statistical techniques. They do not require humans to write explicit procedures to detect facial features, such as eyes, nose, and mouth. A major limitation of such statistical methods is that they require that the input face images are canonical. To deal with variation in the position and size of the faces in an input image, a pixel-based scan window has been used. The size of the window changes within an expected range. For each size, the scan window scans the input image by centering it at each pixel. Each position with each size of the scan window determines a subimage. Such a subimage is scaled to the standard input size for face recognition. Many statistical methods use such a scan method to deal with position and size variation of face in a static input image. These statistical methods are well understood and easy to implement. Various versions of this class of methods have been implemented by many research groups and have been tested extensively in the blind FERET tests with a large number of images [22]. A large portion of commercial systems is based on this class of algorithms. Another popular statistical classifier is the HMM classifier. Referring to [10], [11], and [12] it is apparent that in HMM approaches, the resulting classifier is sensitive to the selection of parameters such as the number of iterations used during training [23]. In addition, the resulting recognition rates reported in literature are low compared to other methods with the exception of the work presented in [9].

### *2.1.2 Non-Statistical Approaches*

Neural networks are found to be popular in the field of face recognition. Many papers use different kinds of neural networks as face recognizers: Radial-Basis Functions (RBF) ([24], [7], and [14]), Support Vector Machines (SVM) ([15], [25], [26], and [27]), and Learning Vector Quantization (LVQ) [24] are some examples. A survey on the application of neural networks in the field of face recognition can be found in [16]. One of the advantages of using neural networks is that, for some designs, no feature extraction phase is required. The feature extraction is left to the network and the face data is presented in its raw form (or may be after some simple preprocessing). The main disadvantages are in the fact that in order for the neural network to be used in real life applications, it needs to be trained on images taken under all possible illumination changes, poses, facial expressions, …etc. of the persons to be recognized, which is not practically available. In addition, many neural networks are unstable classifiers like LVQ, Multi-Layer Perceptrons (MLP), and Back Propagation networks. Other non-statistical methods are also present. Shape From Shading (SFS) ([28], [29], and [30]), Elastic Bunch-Graph Matching [17], Dynamic Link Matching [31], and Optical flow [32] are some examples. Most of these methods address the problems of illumination changes, pose variation, and facial expressions. Generally, they are less efficient in terms of processing time.

### *2.1.3 Comparisons*

In [8], a comparative study has been performed for three face recognition techniques, namely, eigenface, autoassociation and classification networks, and elastic matching. First, these techniques were analyzed under a statistical decision framework. Then they were evaluated experimentally on four different databases of moderate subject size and a combined database of more than 100 subjects. The results indicate that the eigenface algorithm, which is essentially a minimum distance classifier, works well when lighting variation is small. Its performance deteriorates significantly as lighting variation increases. The reason for this deterioration is that lighting variation introduces biases in distance calculations. When such biases are large, the image distance is no longer a reliable measure of face difference. The elastic-matching algorithm, on the other hand, is insensitive to lighting, face position, and expression variations and therefore is more versatile. This owes to the Gabor features, which are insensitive to lighting variation, rigid, and deformable matching, which allows for position and expression variation, and the fact that only features at key points in the image, rather than the entire image, are used. The performance of the autoassociation and classification nets is upper bounded by that of the eigenface and is more difficult to implement in practice. The main disadvantage of elastic matching is its low computational efficiency. In [18] a comparison of two biologically motivated techniques, eigenfaces and graph matching, is presented. Although the biologically inspired models are very useful for neuroscientists, ultimately, when building a commercial face recognition system, one should use the algorithm with the highest performance, regardless of biological relevance. However, for specialized applications, such as witness face reconstruction, in which human perception of similarity is relevant to the task, models developed using human psychophysical evidence might outperform other algorithms. Face recognition, especially in a cluttered dynamic environment, is a difficult problem; most of the published results have been obtained on static, high quality, frontal facial images. Better algorithms are needed to overcome the problems of out-of-plane facial rotation, lighting variations, occlusion, and viewpoint changes. Many researchers have derived inspiration from the biological study of face recognition, but it is unclear whether these techniques succeed either as physiological models or as effective algorithms. This leads to the following conclusions:

- The neurophysiological evidence is sufficiently ambiguous to permit several plausible models; with the addition of pre- or post-processing steps, almost any model can be adjusted to fit the available evidence.

- Although PCA based techniques appear computationally elegant, they suffer from the flawed assumption that reprojecting images into an eigenvector basis will improve the separability of the image classes.
- One of the most promising areas for computer based face recognition algorithms is the development of systems that correlate well with human ratings of similarity. Current computer algorithms, such as PCA and graph matching, correlate well with each other, but are less good as predictors of human perception.
- Standard face image datasets are typically inadequate for measuring the true performance of algorithms, since they lack illumination and background variation. As shown by ARENA [33], even relatively simple approaches, such as nearest-neighbor classifiers, can excel on such a test set.

### *2.2 Combined Classifiers Systems*

In many cases, a number of classifiers (may or may not be of the same nature) are combined to enhance recognition. In this case, the face recognition system is called a combined or hybrid system as in [24], [23], [34], [35], [36], and [37]. A survey on hybrid systems in face recognition can be found in [38]. In [24] two neural classifiers (RBF and LVQ) are combined to enhance classification. The classifier implemented in [24] combines the generalization characteristics of both the LVQ and RBF classifier networks. The investigations described in [24] were performed using facial images of the ORL database [39]. The whole set of images is resampled to three different sizes: 24x24, 32x32 and 64x64. A low pass filter is applied to the image before interpolation using the nearest neighbor interpolation method. This reduces the effect of Moiré patterns and ripple patterns that result from aliasing during resampling. After resampling all images will have the same size. Many experiments were performed on the individual classifiers to obtain the best performer. Although different, nearly the two individual classifiers (LVQ, and RBF networks) agree upon the final decision of classifying the faces in the ORL database. This means that the diversity criterion is not satisfied because both classifiers misclassified the same test patterns. This situation means that these patterns have special attributes, which cause their interference with other classes. The classes of these patterns together with the interfering classes are designated as Familiar Classes (FC). The other correctly recognized classes are designated as Distinctive Classes (DC). Considering the above diversity problem, designing a special type of classifier (figure 2.1) is necessary to resolve the confusion problems caused by the familiar faces. The classifier design steps are:

1- Train the best performing individual classifier (LVQ) on the whole training set (200 faces from 40 persons).

2- Test the LVQ classifier of step 1 on the whole test set (other 200 faces from 40 persons).

3- Separate the testing set into two groups: the correctly classified group (DC), and the misclassified group together with the classes with which they interfere (FC).

4- Train two new classifiers (LVQ and RBF) one on the DC faces and the other on the familiar classes of faces (FC).

5- Apply a front-end classifier (FEC) on the outputs of the DC and FC classifiers.

Figure 2.1 shows the architecture of the combined classifier.

The best performing classifiers are selected for building a combined classifier. A learning vector quantization network with 1200 hidden neurons resulted in 99% correct classification. A radial basis function network trained on the faces of the confused classes, resulted in only one misclassification. Combining the results of both classifiers, the system performance is improved to a recognition rate of 99.5% (table 2.1).

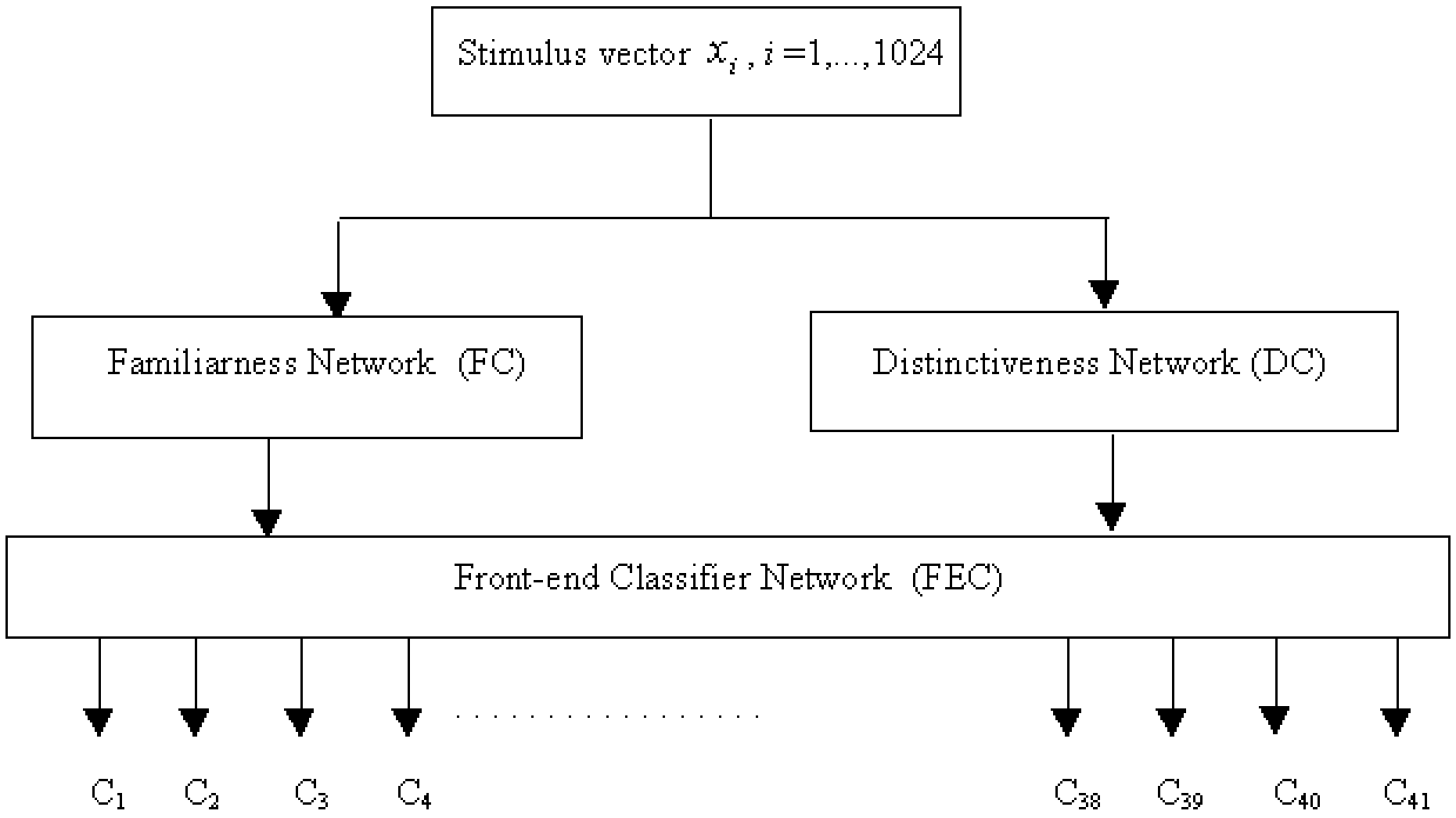

Figure 2.1. The combined classifiers system of [24]

Table 2.1. Performance of different classifiers in [24]

| Set | Recognition Rate % | | |
|---|---|---|---|
| | LVQ Classifier | RBF Classifier | Combined Classifier LVQ+RBF+FEC |
| Training | 100 | 100 | 100 |
| Test | 99.0 | 98 | 99.5 |

Despite achieving a high recognition rate, the technique described in [24] has some disadvantages:

- The training parameters of the base classifiers need to be pre-selected by trial and error to result in their high recognition rates. This leads to the complex problem of parameter selection that complicates the design process.

- The design process involves using the test images to construct the DC and FC sets. In other words, the information contained in the test images contribute in the design and training process. Thus, the test images should be in fact considered as training images. Hence, the reported recognition rates are not based on independent test images and cannot be used as a true measure of the system performance.

In [23] a system for person identification is presented. The system is based on the combination of three face classifiers: an eigen face classifier, a HMM classifier, and a profile classifier. Since the scores (outputs) of the three classifiers represent different measures in different ranges, a transformation is used to enable the combination of the three outputs. Three methods of combining the three outputs were used:

a) Voting: Each classifier gives a single vote equal to the other two. The class taking the majority votes is the final decision. If voting ends with a draw, the input pattern is rejected.

b) Ranking: By summing the ranks for every class in the combination set and taking the class with the lowest rank sum as the final decision.

c) Scoring: By summing the transformed outputs (scores) of the three classifiers and ranking in ascending order.

A face database for 30 persons is used. The face images are divided into two sets:

1) Frontal images: 10 (512 by 342) gray level images per person with varying head positions. The faces are cropped and normalized to the width of 40 pixels.

2) Profile images: 5 (512 by 342) binarized profile images per person with varying head positions.

The results indicate that by combining the three classifiers one obtains higher recognition rate (99.7%) than using one classifier or combining any two classifiers. This technique also has some disadvantages:

- The input images must come from two sources, frontal and profile. Each frontal image must be treated in two different ways to be suitable for the two frontal classifiers (HMM and PCA). This increases the storage and processing demands of the combined system.

- The HMM classifier requires optimal parameter selection or it will give a low recognition rate.
- The used face database is small and the results could be completely different for a larger database.
- The outputs of the three classifiers are very different. The combination strategy by scoring is the best in [23] but it also requires some parameters selection to be optimal.

In [36] the AdaBoost algorithm is used on a composite database of 137 individuals with 10 images per person. Since the AdaBoost is mainly used in two class problems, a method called the Constrained Majority Voting (CMV) is used to combine a reduced set from all the pairwise classification results without loosing the recognition accuracy. Principal Components Analysis (PCA) is used to construct the input to the AdaBoost classifiers as a feature vector of the principal components of the face images. The reported recognition rate is around 86%.

Boosting is one of the famous combination techniques in pattern recognition. Although Boosting succeeded in many cases, it failed in others. Some researchers criticize Boosting techniques for many reasons. As stated in [5], for example, there is no through understanding of how overfitting is avoided or controlled within the training process, thus there is no guarantee on the results. The empirical evidence does show that these techniques do not always work. Apart from the use of AdaBoost in [36], there are other disadvantages as well:
- Some parameters that affect the recognition rate are required to be pre-selected.
- The reported recognition rates are lower than other techniques.
- The stability of the final system is within 6%. This means that using the same system with the same parameters the resulting recognition rate might range from 80% to 86%. Thus, this system can be considered low in its stability.

The same disadvantages are present in the system of [35] where a pairwise classification framework for face recognition is developed. In [35], a C class face recognition problem is divided into a set of C(C-1)/2 two-class problems. Such a problem decomposition provides a framework for independent feature selection for each pair of classes. A simple feature ranking strategy is used to select a small subset of the features for each pair of classes. Furthermore, two classification methods under the pairwise comparison framework are evaluated: the Bayes classifier and the AdaBoost. Experiments on a face database with 1079 face images of 137 individuals indicate that 20 features (derived from PCA) are enough to achieve relatively high recognition accuracy (88% at most). As stated in [35], the overall recognition rates are improved consistently for the Bayes classifier. The performance of the AdaBoost method deteriorates as more features are presented. The authors interpreted this as the interior parameters of the AdaBoost should be adjusted more carefully for the special case of face recognition in order to get high accuracy constantly. Hence, the conclusion is that for the AdaBoost algorithm, further work should be done to improve its performance for face recognition.

In [37] a system for recognizing human faces with any view in the range of 30 degrees left to 30 degrees right out of plane rotation is presented. The system uses view specific eigenface analysis to extract the features that are fed to the next stage of view specific neural networks. The final stage is a neural network to combine the decisions of the view specific networks. The system is shown in the figure 2.2.

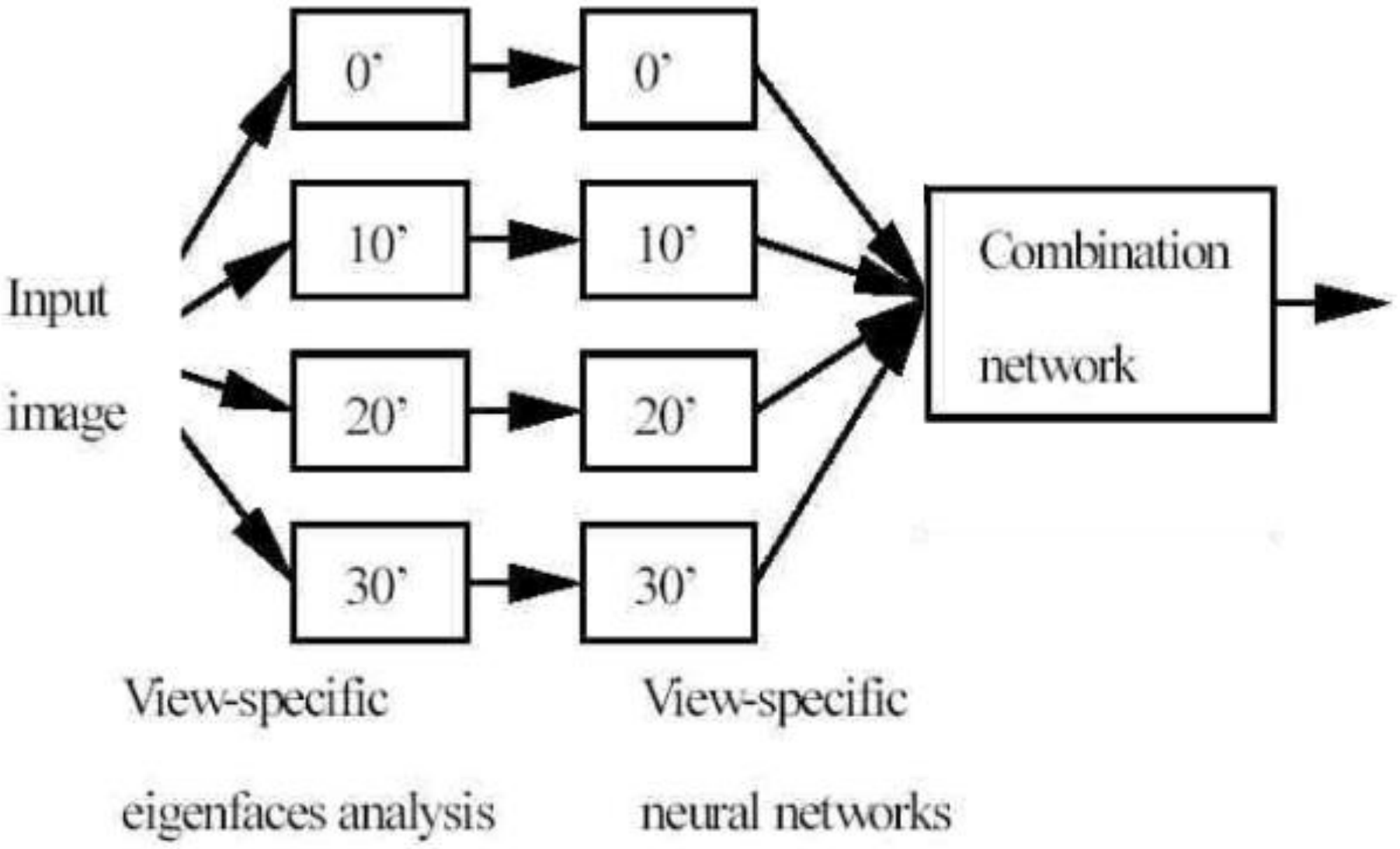

Figure 2.2. The system presented in [37]

The system is used to capture all the frames containing a specified person's face from a video sequence having attributes such as large face movement, out of plane rotation and scaling. The number of persons was kept low (10 only) with 5 to be the main classes and the other 5 to be used as a $6^{th}$ rejection class. The average recognition rate of the system is 98.75%.

The main disadvantage of the system in [37] is the small number of recognizable persons. This will lead to the following undesirable effects:

- If this number is increased, the neural networks will require to be largely expanded. This will certainly increase storage requirements and require a long training time (if training is possible at all).
- The reported recognition rate is unrealistic due to the small number of classes.

Finally, in [34] an approach for fully automated face recognition is described. The approach is based on a hybrid architecture consisting of an ensemble of connectionist networks - Radial Basis Functions (RBF) - and inductive Decision Trees (DT). The benefits of such architecture include robust face recognition using consensus methods over ensembles of RBF networks. Experiments carried out on a large database consisting of 748 images corresponding to 374 subjects yield on the average 87% correct recognition rate. The system automatically detects, normalizes, and recognizes faces. The recognition task is performed by the RBF ensemble of neural networks. No feature extraction phase is required for the recognition task. The recognition rate is lower than other reported methods. In addition, the recognition phase requires a highly normalized face image and hence relies completely on the accuracy of the face detection phase.

### *2.3 Conclusions*

From the discussions in chapter 1 and 2, the following points can be concluded:

- The possible applications of face recognition are alone a very powerful motivation to make this field an active field of research especially when the God-made solution, the human brain, is a proof of the possibility of a very powerful solution for the problem.
- Face recognition is not a simple task, its biological origins are not yet well understood and hence the field is wide open for new models and techniques.
- The nature of a face recognition application is the main factor that researchers can decide upon the best technique to use. There is no global solution for the problem yet.
- No universal agreement is present on how evaluating face recognition systems should exactly be performed. This is mainly due to the diversity of both the face recognition techniques and the possible applications of face recognition.
- An important conclusion of this chapter is that if there is an approach that is the closest to a global solution for the problem, it will be the combined systems approach. That is because combined systems capture the strengths of the combined classifiers and obtain an overall accuracy that is higher than that of any of the underlying base classifiers.

- Currently, combined classifiers techniques still suffer from problems caused by the weaknesses of their base classifiers.

*Chapter 3*

# Pattern Recognition

Referring to [40], this chapter aims at providing an overview of the general problem of pattern recognition. The disadvantages presented in the previous chapter are all results of the difficulties presented in this chapter since face recognition is a special case of the more general problem of pattern recognition. Section 3.1 is an introduction, section 3.2 presents the concept of machine perception, section 3.3 is an example of a pattern classification problem, section 3.4 presents the sub-problems of pattern classification, section 3.5 introduces the concepts of learning and adaptation, and finally section 3.6 presents the conclusions of this chapter.

## *3.1 Introduction*

The ease with which humans recognize a face, understand spoken words, read handwritten characters, identify their car keys in their pockets by feel, and decide whether an apple is ripe by its smell belies the astoundingly complex processes that underlie these acts of pattern recognition. Pattern recognition, the act of taking in raw data and taking an action based on the “category” of the pattern, has been crucial for our survival, and Allah has given us highly sophisticated neural and cognitive systems for such tasks.

## *3.2 Machine Perception*

It is natural that humans should seek to design and build machines that can recognize patterns. From automated speech recognition, fingerprint identification, optical character recognition and much more, it is clear that reliable, accurate pattern recognition by machine would be immensely useful. Moreover, in solving the myriad problems required to build such systems, researchers gain deeper understanding and appreciation for pattern recognition systems in the natural world, most particularly in humans. For some applications, such as speech and visual recognition, the design efforts may in fact be influenced by knowledge of how these are solved in nature, both in the algorithms employed and the design of special purpose hardware.

### *3.3 An Example*

To illustrate the complexity of some of the types of problems involved, the following example is given. Supposing that a fish packing plant wants to automate the process of sorting incoming fish on a conveyor belt according to species. As a pilot project it is decided to try to separate sea bass from salmon using optical sensing. A camera is set up, some sample images are taken and some physical differences between the two types of fish are noted: length, brightness, width, number and shape of fins, position of the mouth, and so on. These suggest features to explore for use in the classifier. Also noise or variations in the images are noticed: variations in lighting, position of the fish on the conveyor, even “static” due to the electronics of the camera itself. Given that there truly are differences between the population of sea bass and that of salmon, they are viewed as having different models - different descriptions, which are typically mathematical in form. The overarching goal and approach in pattern classification is to hypothesize the class of these models, process the sensed data to eliminate noise (not due to the models), and for any sensed pattern choose the model that corresponds best. Any techniques that further this aim should be in the conceptual toolbox of the designer of pattern recognition systems. The prototype system to perform this very specific task might well have the form shown in figure 3.1. First the camera captures an image of the fish. Next, the camera's signals are preprocessed to simplify subsequent operations without loosing relevant information. In particular, a segmentation operation in which the images of different fish are somehow isolated from one another and from the background might be used.

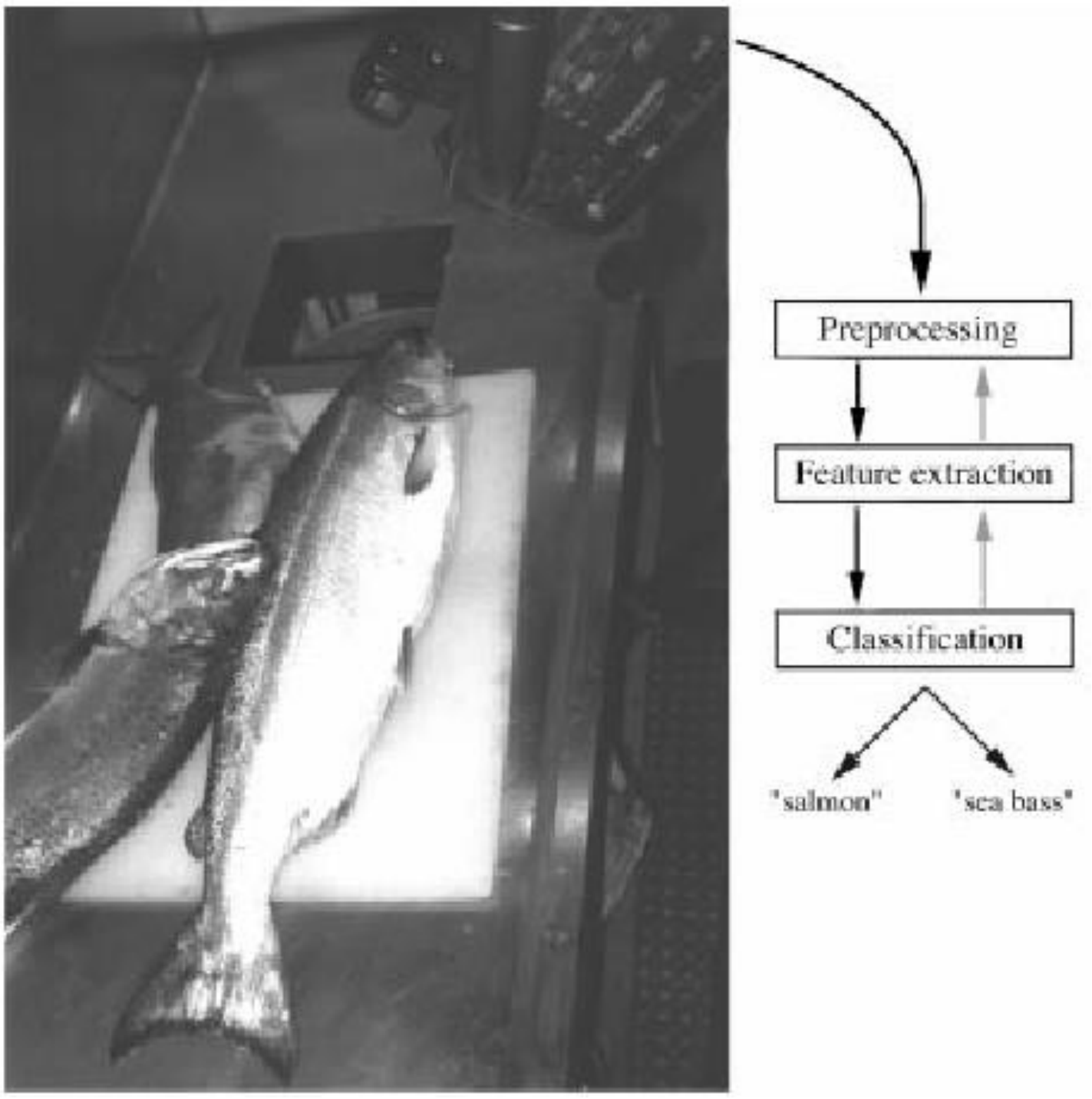

Figure 3.1. Pattern recognition steps

The information from a single fish is then sent to a feature extractor, whose purpose is to reduce the data by measuring certain “features” or “properties.” These features (or, more precisely, the values of these features) are then passed to a classifier that evaluates the evidence presented and makes a final decision as to the species. The preprocessor might automatically adjust for average light level or threshold the image to remove the background of the conveyor belt, and so forth. Supposing somebody at the fish plant states that a sea bass is generally longer than a salmon. These, then, give the tentative models for the fish: sea bass have some typical length, and this is greater than that for salmon. Then length becomes an obvious feature, and the fish might be classified merely by seeing whether or not the length l of a fish exceeds some critical value l*. To choose l* one could obtain some design or training samples of the different types of fish, make length measurements, and inspect the results. Supposing that this is done, the histograms shown in figure 3.2 are obtained. These disappointing histograms bear out the statement that sea bass are somewhat longer than salmon, on average, but it is clear that this single criterion is quite poor; no matter how l* is chosen, sea bass can not be reliably separated from salmon by length alone. Next another feature is tried: the average brightness of the fish scales. Variations in illumination are now carefully eliminated, since they can only obscure the models and corrupt the new classifier. The resulting histograms, shown in figure 3.3, are much more satisfactory; the classes are much better separated.

So far, it has tacitly been assumed that the consequences of these actions are equally costly: deciding the fish was a sea bass when in fact it was a salmon was just as undesirable as the converse. Such asymmetry in the cost is often, but not invariably the case. For instance, a fish packing company might know that its customers easily accept occasional pieces of tasty salmon in their cans labeled “sea bass,” but they object vigorously if a piece of sea bass appears in their cans labeled “salmon.” If the company wants to stay in business, it should adjust the decision boundary to avoid antagonizing the customers, even if it means that more salmon makes its way into the cans of sea bass. In this case, then, the decision boundary x* should be moved to smaller values of brightness, thereby reducing the number of sea bass that are classified as salmon (figure 3.3). The more the customers object to getting sea bass with their salmon - i.e., the more costly this type of error - the lower the decision threshold x* should be set in figure 3.3.

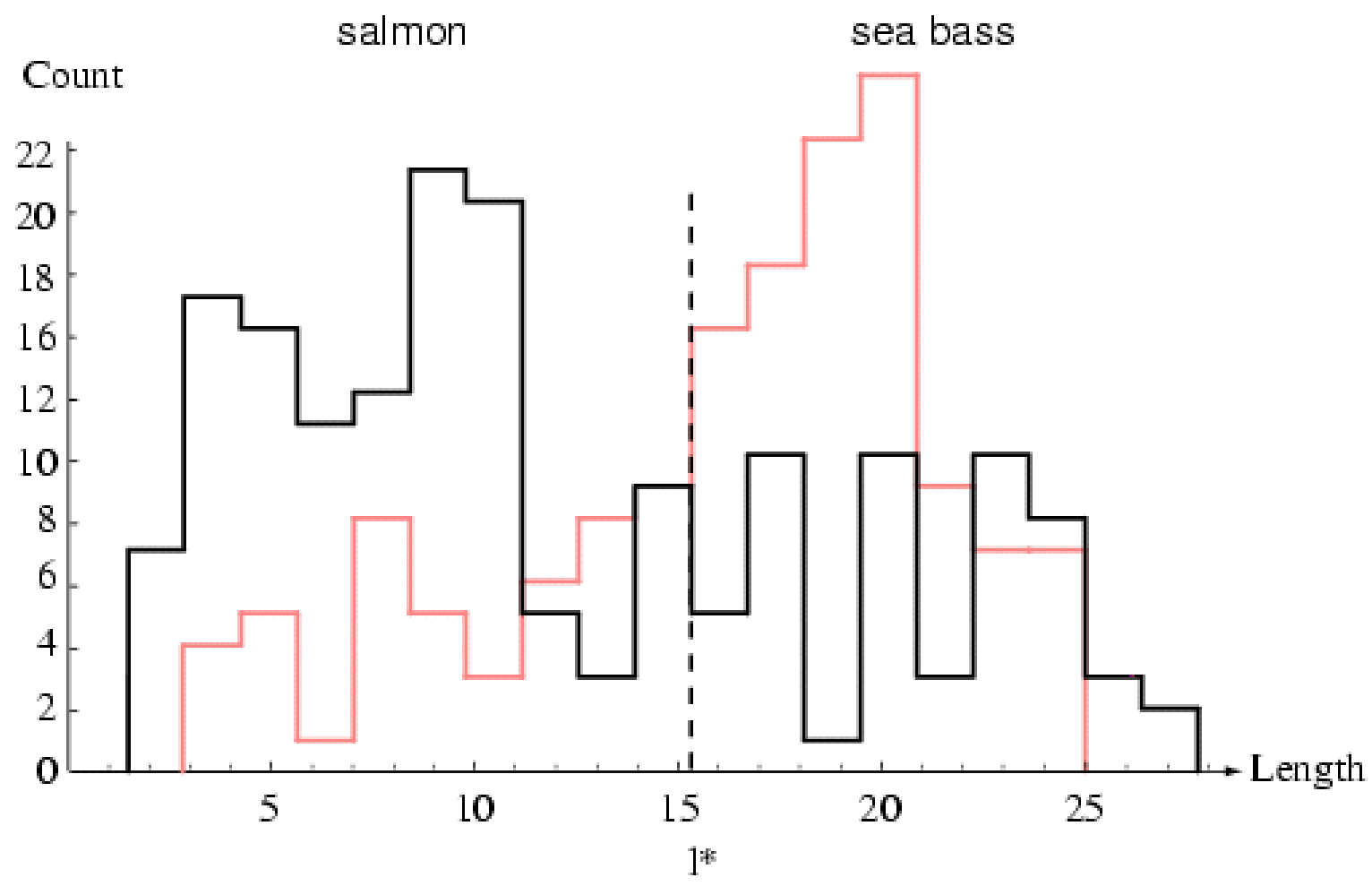

Figure 3.2. Histograms for the length feature for the two categories.

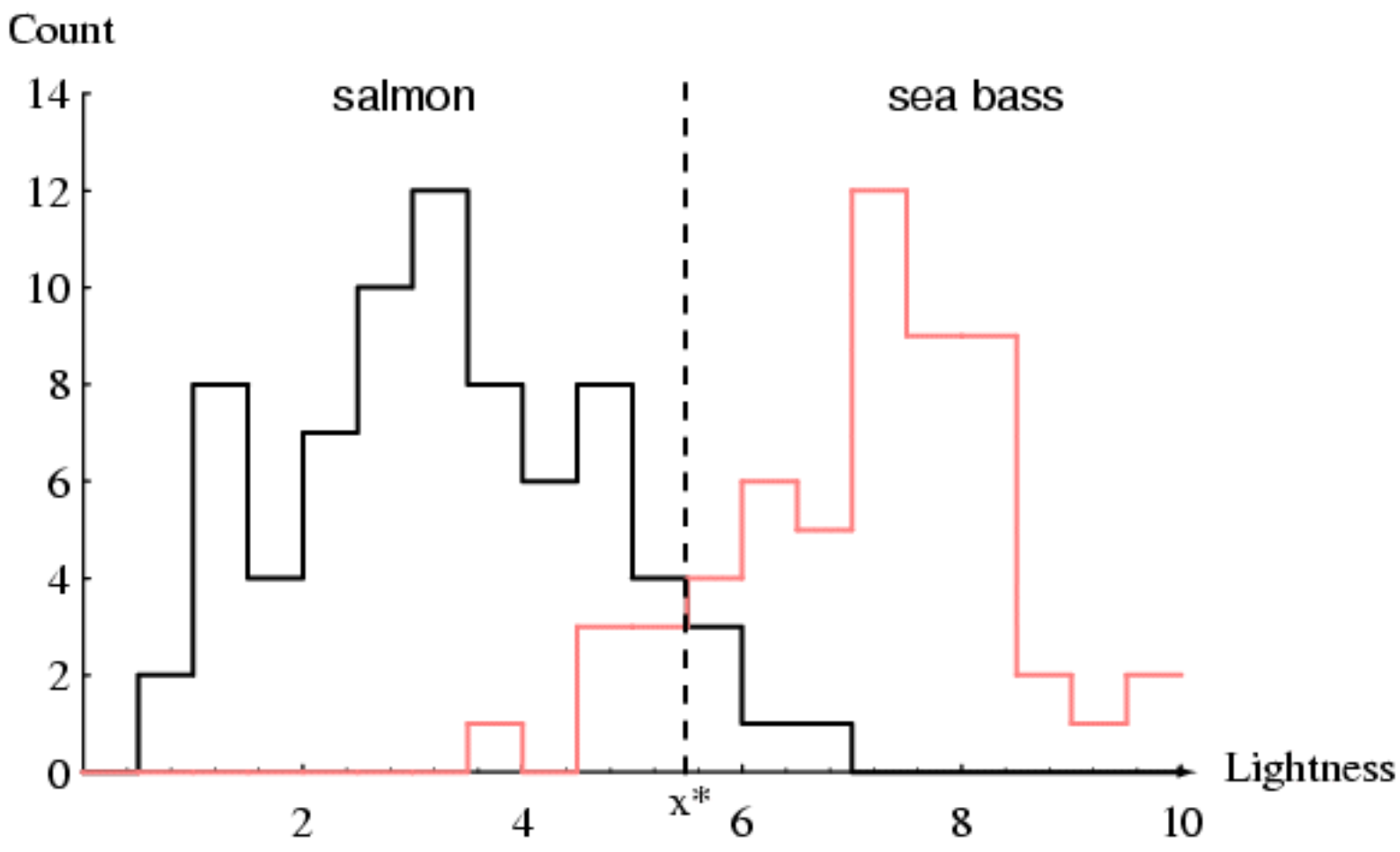

Figure 3.3. Histograms for the brightness feature for the two categories.

Such considerations suggest that there is an overall single cost associated with the decision, and the true task is to make a decision rule (i.e., set a decision boundary) so as to minimize such a cost. This is the central task of decision theory of which pattern classification is perhaps the most important subfield. Even if the costs associated with the decisions are known and the optimal decision boundary x* is chosen, the resulting performance might be dissatisfactory. The first impulse might be to seek yet a different feature on which to separate the fish. Assuming that no other single visual feature yields better performance than that based on brightness, then to improve recognition the use of more than one feature at a time is needed. In the search for other features, the observation that sea bass are typically wider than salmon can be relied on. Now two features for classifying fish are present: the brightness $x_1$ and the width $x_2$. Ignoring how these features might be measured in practice, the feature extractor has thus reduced the image of each fish to a point or feature vector x in a two-dimensional feature space, where:

$$\mathbf{x} = \begin{bmatrix} x_1 \\ x_2 \end{bmatrix} \quad \ldots(3.1)$$

The problem now is to partition the feature space into two regions, where for all patterns in one region the fish will be called a sea bass, and all points in the other will be called a salmon. Supposing that the feature vectors for the samples are measured, the scattering of points shown in figure 3.4 is obtained. This plot suggests the following rule for separating the fish: Classify the fish as sea bass if its feature vector falls above the decision boundary shown and as salmon otherwise. This rule appears to do a good job of separating the samples and suggests that perhaps incorporating yet more features would be desirable. Besides the brightness and width of the fish, some shape parameter might be included, such as the vertex angle of the dorsal fin, or the placement of the eyes (as expressed as a proportion of the mouth-to-tail distance), and so on. It is difficult to know beforehand which of these features will work best. Some features might be redundant: for instance if the eye color of all fish correlated perfectly with width, then classification performance need not be improved if eye color is included as a feature.

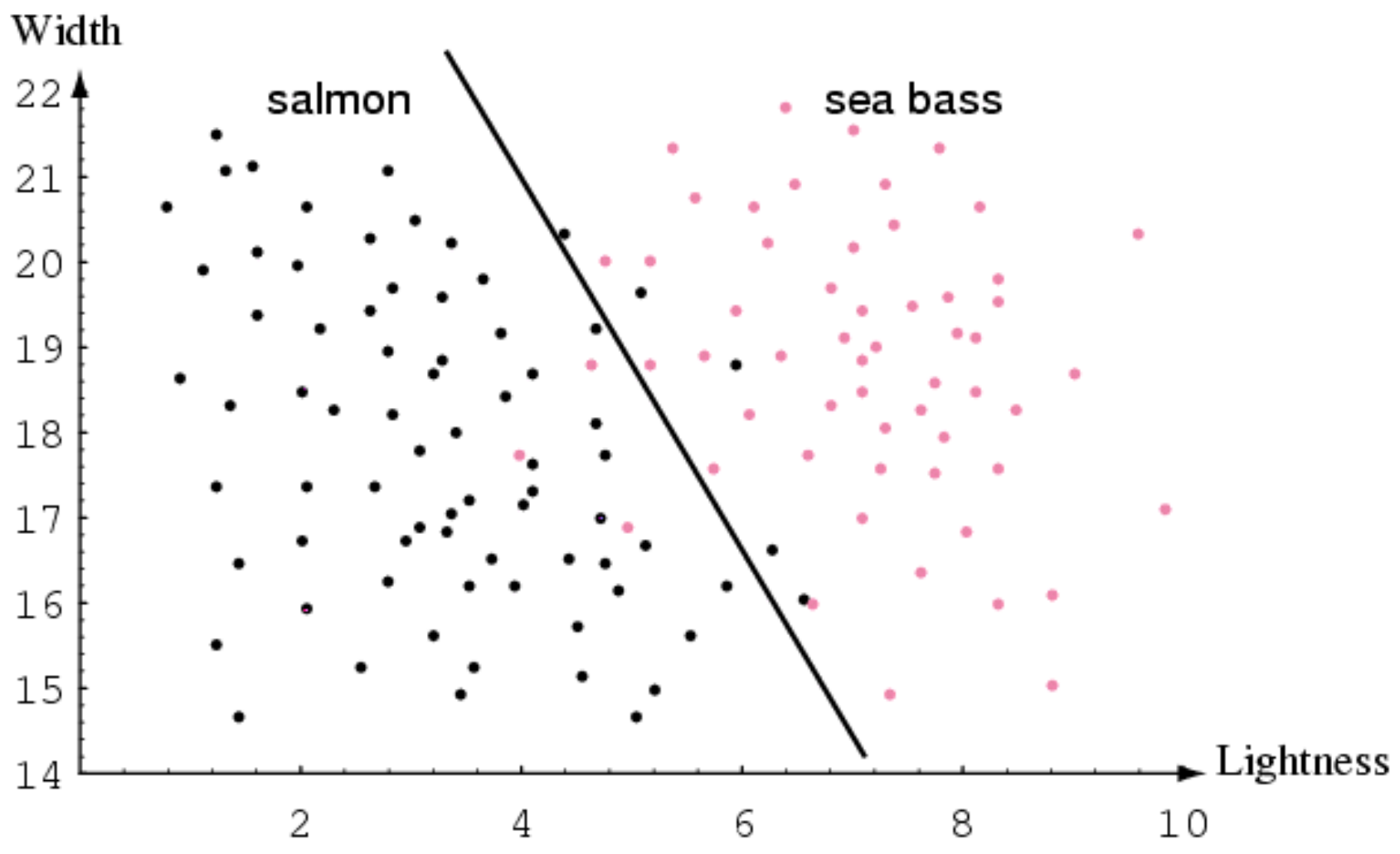

Figure 3.4. The two features of brightness and width for sea bass and salmon

Supposing that other features are too expensive to measure, or provide little improvement (or possibly even degrade the performance) in the approach described above, and that the classifier is forced to make the decision based on the two features in figure 3.4. If the models were extremely complicated, the classifier would have a decision boundary more complex than the simple straight line. In that case, all the training patterns would be separated perfectly, as shown in figure 3.5. With such a "solution," though, the satisfaction would be premature because the central aim of designing a classifier is to suggest actions when presented with novel patterns, i.e., fish not yet seen. This is the issue of generalization. It is unlikely that the complex decision boundary in figure 3.5 would provide good generalization, since it seems to be "tuned" to the particular training samples, rather than some underlying characteristics or true model of all the sea bass and salmon that will have to be separated. Naturally, one approach would be to get more training samples for obtaining a better estimate of the true underlying characteristics, for instance the probability distributions of the categories. In most pattern recognition problems, however, the amount of such data that can be obtained easily is often quite limited. Even with a vast amount of training data in a continuous feature space though, if the approach in figure 3.5 is followed, the classifier would give a horrendously complicated decision boundary, one that would be unlikely to do well on novel patterns. Rather, then, one might seek to "simplify" the recognizer, motivated by a belief that the underlying models will not require a decision boundary that is as complex as that in figure 3.5. Indeed, the slightly poorer performance on the training samples might be satisfactory if it means that the classifier will have better performance on novel patterns. One of the central problems in statistical pattern recognition is the ability of the system to automatically determine that the simple curve in figure 3.6 is preferable to the manifestly simpler straight line in figure 3.4 or the complicated boundary in figure 3.5. Assuming that one manages somehow to optimize this tradeoff, the ability to predict how well the system will generalize to new patterns is another central problem in statistical pattern recognition.

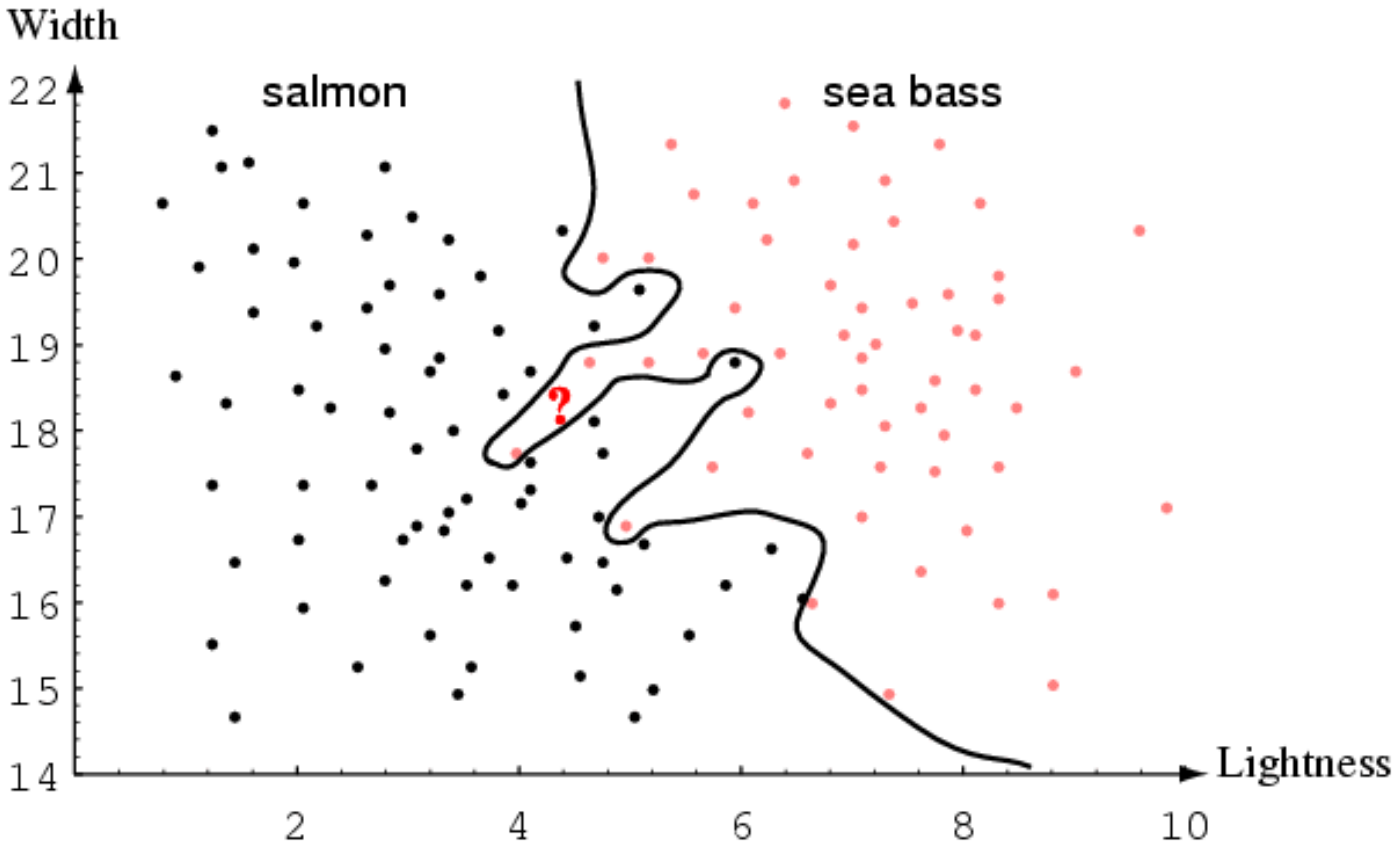

Figure 3.5. Overly complex models for the fish will lead to complicated decision boundaries

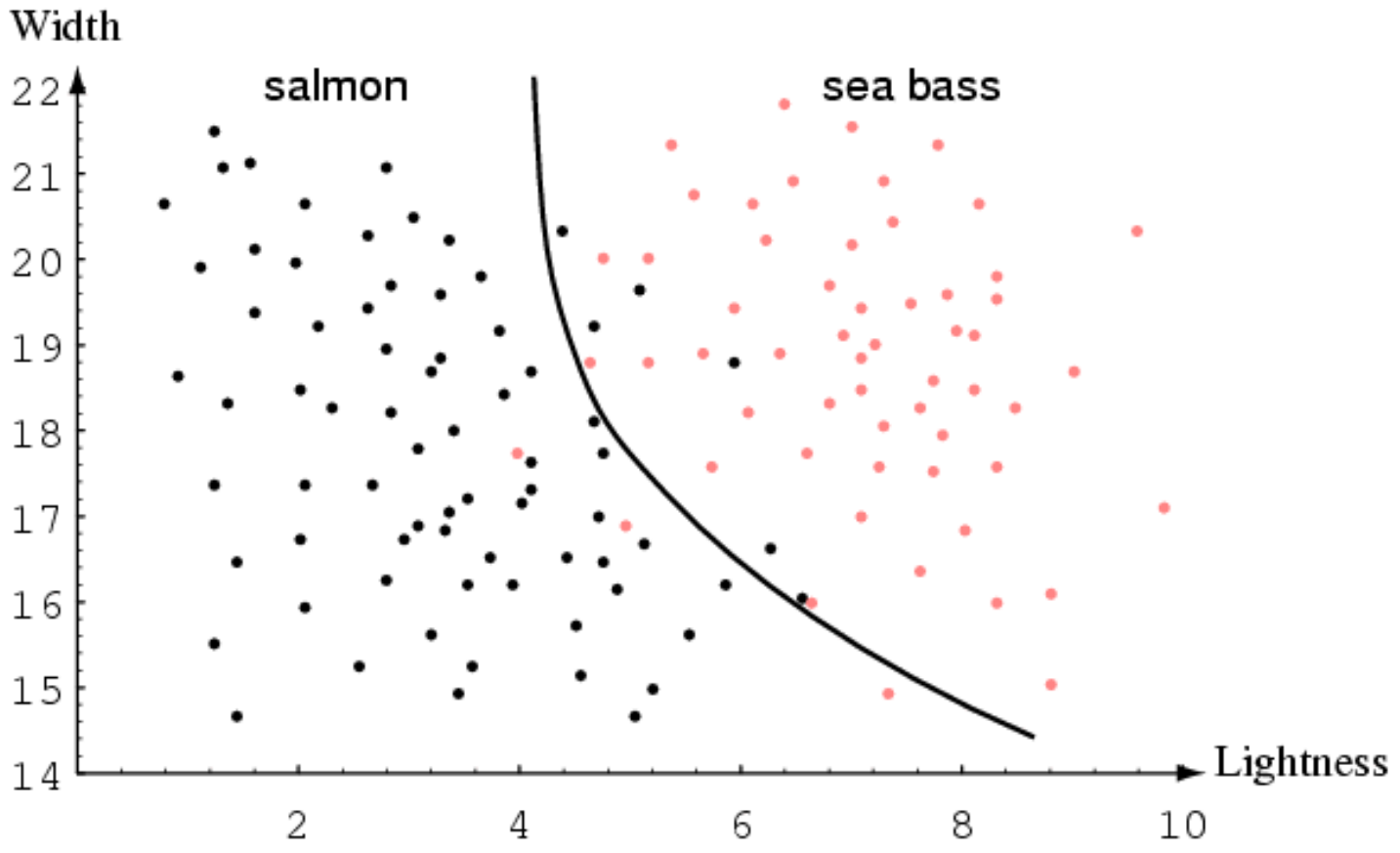

Figure 3.6. The optimal tradeoff between performance on the training set and simplicity of classifier

For the same incoming patterns, one might need to use a drastically different cost function, and this will lead to different actions altogether. For instance, one might wish, instead, to separate the fish based on their sex - all females (of either species) from all males if it is required to sell roe. Alternatively, one might wish to cull the damaged fish (to prepare separately for cat food), and so on. Different decision tasks may require features and yield boundaries quite different from those useful for the original categorization problem. This makes it quite clear that the decisions are fundamentally task or cost specific, and that creating a single general purpose artificial pattern recognition device - i.e., one capable of acting accurately based on a wide variety of tasks - is a profoundly difficult challenge. This, too, should raise the appreciation of the ability of humans to switch rapidly and fluidly between pattern recognition tasks.

Since classification is, at base, the task of recovering the model that generated the patterns, different classification techniques are useful depending on the type of candidate models themselves. In statistical pattern recognition the focus is on the statistical properties of the patterns (generally expressed in probability densities). Here the model for a pattern may be a single specific set of features, though the actual pattern sensed has been corrupted by some form of random noise. Occasionally it is claimed that neural pattern recognition (or neural network pattern classification) should be considered its own discipline, but despite its somewhat different intellectual pedigree, it should be considered a close descendant of statistical pattern recognition. If instead the model consists of some set of crisp logical rules, then the methods of syntactic pattern recognition are employed, where rules or grammars describe the decision. For example one might wish to classify an English sentence as grammatical or not, and here statistical descriptions (word frequencies, word correlations, etc.) are inappropriate.

It is necessary in the fish example to choose the features carefully, and hence achieve a representation (as in figure 3.6) that enabled reasonably successful pattern classification. A central aspect in virtually every pattern recognition problem is that of achieving such a “good” representation, one in which the structural relationships among the components is simply and naturally revealed, and one in which the true (unknown) model of the patterns can be expressed. In some cases patterns should be represented as vectors of real-valued numbers, in others ordered lists of attributes, in yet others descriptions of parts and their relations, and so forth. A representation in which the patterns that lead to the same action are somehow “close” to one another, yet “far” from those that demand a different action is sought for. The extent to which one creates or learns a proper representation and how one quantifies near and far apart will determine the success of the pattern classifier. A number of additional characteristics are desirable for the representation. One might wish to favor a small number of features, which might lead to simpler decision regions and a classifier easier to train. One might also wish to have features that are robust, i.e., relatively insensitive to noise or other errors.

In practical applications it might be required from the classifier to act quickly, or use few electronic components, memory or processing steps.

A central technique, when having insufficient training data, is to incorporate knowledge of the problem domain. Indeed the less the training data the more important is such knowledge, for instance how the patterns themselves were produced. One method that takes this notion to its logical extreme is that of analysis by synthesis, where in the ideal case one has a model of how each pattern is generated. Considering speech recognition, amidst the manifest acoustic variability among the possible "dee"s that might be uttered by different people, one thing they have in common is that they were all produced by lowering the jaw slightly, opening the mouth, placing the tongue tip against the roof of the mouth after a certain delay, and so on. It may be assumed that "all" the acoustic variation is due to the happenstance of whether the talker is male or female, old or young, with different overall pitches, and so forth. At some deep level, such a "physiological" model (or so-called "motor" model) for production of the utterances is appropriate and different from that for "doo" (for example) and indeed all other utterances. If this underlying model of production can be determined from the sound, then the utterance can be classified by how it was produced. That is to say, the production representation may be the "best" representation for classification. The pattern recognition systems should then analyze (and hence classify) the input pattern based on how one would have to synthesize that pattern. The problem is, of course, to recover the generating parameters from the sensed pattern.

Making a recognizer of all types of chairs - standard office chair, contemporary living room chair, beanbag chair, and so forth - based on an image is very difficult. Given the astounding variety in the number of legs, material, shape, and so on, one might despair of ever finding a representation that reveals the unity within the class of chair. Perhaps the only such unifying aspect of chairs is functional: a chair is a stable artifact that supports a human sitter, including back support. Thus one might try to deduce such functional properties from the image, and the property "can support a human sitter" is very indirectly related to the orientation of the larger surfaces, and would need to be answered in the affirmative even for a beanbag chair. Of course, this requires some reasoning about the properties and naturally touches upon computer vision rather than pattern recognition proper.

Without going to such extremes, many real world pattern recognition systems seek to incorporate at least some knowledge about the method of production of the patterns or their functional use in order to insure a good representation, though of course the goal of the representation is classification, not reproduction. For instance, in optical character recognition (OCR) one might confidently assume that handwritten characters are written as a sequence of strokes, and first try to recover a stroke representation from the sensed image, and then deduce the character from the identified strokes.

### *3.4 The Sub-problems of Pattern Classification*

Some of the issues in pattern classification were alluded to, and now a more explicit list of issues is turned to. In practice, these typically require the bulk of the research and development effort. Many are domain or problem specific, and their solution will depend upon the knowledge and insights of the designer. Nevertheless, a few are of sufficient generality, difficulty, and interest that they warrant explicit consideration.

#### *3.4.1 Feature Extraction*

The conceptual boundary between feature extraction and classification proper is somewhat arbitrary: an ideal feature extractor would yield a representation that makes the job of the classifier trivial; conversely, an omnipotent classifier would not need the help of a sophisticated feature extractor. The distinction is forced upon the designer for practical, rather than theoretical reasons. Generally speaking, the task of feature extraction is much more problem and domain dependent than is classification proper, and thus requires knowledge of the domain. A good feature extractor for sorting fish would surely be of little use for identifying fingerprints, or classifying photomicrographs of blood cells. Some of the problems regarding feature extraction are the ability to know which features are most promising, ways to automatically learn which features are best for the classifier, and the number of features that should be used.

### *3.4.2 Noise*

The lighting of the fish may vary, there could be shadows cast by neighboring equipment, the conveyor belt might shake - all reducing the reliability of the feature values actually measured. Noise is defined in very general terms as any property of the sensed pattern due not to the true underlying model but instead to randomness in the world or the sensors. All non-trivial decision and pattern recognition problems involve noise in some form. An important problem is knowing somehow whether the variation in some signal is noise or instead due to complex underlying models of the fish. Another problem is reducing the effect of such noise on the classification process.

### *3.4.3 Overfitting*

In going from figure 3.4 to figure 3.5 in the fish classification problem, a more complex model of sea bass and of salmon was used. That is, the complexity of the classifier was adjusted. While an overly complex model may allow perfect classification of the training samples, it is unlikely to give good classification of novel patterns - a situation known as overfitting. One of the most important areas of research in statistical pattern classification is determining how to adjust the complexity of the model - not so simple that it cannot explain the differences between the categories, yet not so complex as to give poor classification on novel patterns. Principled methods for finding the best intermediate complexity for a classifier are sought for.

### *3.4.4 Model Selection*

A designer might be unsatisfied with the performance of the fish classifier in figures 3.4 and 3.5, and thus jumps to an entirely different class of models, for instance one based on some function of the number and position of the fins, the color of the eyes, the weight, shape of the mouth, and so on. If the process of model selection can be automated, many of the performance problems will be reduced greatly.

### *3.4.5 Prior Knowledge*

In one limited sense, it has already been seen how prior knowledge about the brightness of the different fish categories helped in the design of a classifier by suggesting a promising feature. Incorporating prior knowledge can be far more subtle and difficult. In some applications the knowledge ultimately derives from information about the production of the patterns, as seen in analysis-by-synthesis. In others the knowledge may be about the form of the underlying categories, or specific attributes of the patterns, such as the fact that a face has two eyes, one nose, and so on.

### *3.4.6 Missing Features*

Sometimes, the value of one of the features cannot be determined during classification. The two-feature recognizer never had a single-variable threshold value $x^*$ determined in anticipation of the possible absence of a feature (figure 3.3). The naive method of merely assuming that the value of the missing feature is zero or the average of the values for the training patterns, is provably non-optimal. Likewise occasionally the system faces missing features during the creation or learning in the recognizer. The process of training or using a classifier is then more complicated or impossible.

### *3.4.7 Mereology*

Humans effortlessly read a simple word such as BEATS. But other words that are perfectly good subsets of the full pattern are present, such as BE, BEAT, EAT, AT, and EATS. These words never enter one's mind, unless explicitly brought to one's attention. Conversely, one can read the two unsegmented words in POLOPONY without placing the entire input into a single word category. This is the problem of subsets and supersets - formally part of mereology, the study of part/whole relationships. It is closely related to that of prior knowledge and segmentation. It appears as though the best classifiers try to incorporate as much of the input into the categorization as "makes sense," but not too much.

### *3.4.8 Segmentation*

In the fish example, it was tacitly assumed that the fish were isolated, separate on the conveyor belt. In practice, they would often be abutting or overlapping, and the system would have to determine where one fish ends and the next begins - the individual patterns have to be segmented. If the fish have been already recognized then it would be easier to segment them. Nevertheless, is is difficult to recognize the images before they have been segmented. It seems one needs a way to know when one has switched from one model to another, or to know when one just has background or "no category".

### *3.4.9 Context*

One might be able to use context - input-dependent information other than from the target pattern itself - to improve the recognizer. For instance, it might be known for the fish packing plant that if a sequence of salmon is observed, it is highly likely that the next fish will be a salmon (since it probably comes from a boat that just returned from a fishing area rich in salmon). Thus, if after a long series of salmon the recognizer detects an ambiguous pattern (i.e., one very close to the nominal decision boundary), it may nevertheless be best to categorize it too as a salmon. Such a simple correlation among patterns - the most elementary form of context - might be used to improve recognition despite its simplicity.

#### *3.4.10 Invariances*

In seeking to achieve an optimal representation for a particular pattern classification task, the problem of invariances was confronted. In the fish example, the absolute position on the conveyor belt is irrelevant to the category and thus the representation should also be insensitive to absolute position of the fish. Here a representation that is invariant to the transformation of translation (in either horizontal or vertical directions) is sought for. The “model parameters” describing the orientation of the fish on the conveyor belt are horrendously complicated - due as they are to the sloshing of water, the bumping of neighboring fish, the shape of the fish net, etc. - and thus one gives up hope of ever trying to use them. These parameters are irrelevant to the model parameters that are of interest anyway, i.e., the ones associated with the differences between the fish categories. Here the transformation of concern is a two-dimensional rotation about the camera's line of sight. A more general invariance would be for rotations about an arbitrary line in three dimensions. The image of even such a “simple” object as a coffee cup undergoes radical variation as the cup is rotated to an arbitrary angle - the handle may become hidden, the bottom of the inside volume come into view, the circular lip appear oval or a straight line or even obscured, and so forth. The designer must insure that the pattern recognizer is invariant to such complex changes. A large number of highly complex transformations arise in pattern recognition, and many are domain specific.

### *3.4.11 Evidence Pooling*

In the fish example it is seen how using multiple features could lead to improved recognition. One might imagine that one could do better if one had several component classifiers. If these categorizers agree on a particular pattern, there is no difficulty. But if they disagree, a “super” classifier should pool the evidence from the component recognizers to achieve the best decision. If calling in ten experts for determining if a particular fish is diseased or not, while nine agree that the fish is healthy, one expert does not. It may be that the lone dissenter is the only one familiar with the particular very rare symptoms in the fish, and is in fact correct. It is the job of the “super” categorizer to know when to base a decision on a minority or majority opinion.

### *3.4.12 Costs and Risks*

It should be realized that a classifier rarely exists in a vacuum. Instead, it is generally to be used to recommend actions (put this fish in this bucket, put that fish in that bucket), each action having an associated cost or risk. Conceptually, the simplest such risk is the classification error: what percentage of new patterns is called the wrong category. However, the notion of risk is far more general. The classifier is often designed to recommend actions that minimize some total expected cost or risk. Thus, in some sense, the notion of category itself derives from the cost or task. A designer should incorporate knowledge about such risks and study their effects on the classification decision.

### *3.4.13 Computational Complexity*

Some pattern recognition problems can be solved using highly impractical algorithms. For instance, one might try to hand label all possible 20 x 20 binary pixel images with a category label for optical character recognition, and use table lookup to classify incoming patterns. Although one might achieve error-free recognition, the labeling time and storage requirements would be quite prohibitive since it would require labeling each of $2^{20 \times 20}$ patterns. Thus the computational complexity of different algorithms is of importance, especially for practical applications. In more general terms, one may ask how an algorithm scales as a function of the number of feature dimensions, or the number of patterns or the number of categories. The tradeoff between computational ease and performance must be investigated. In some problems the designer know he/she can design an excellent recognizer, but not within the engineering constraints. Thus, the designer must optimize within such constraints.

### *3.5 Learning and Adaptation*

In the broadest sense, any method that incorporates information from training samples in the design of a classifier employs learning. Nearly all practical or interesting pattern recognition problems are so hard that one cannot guess classification decision ahead of time. Creating classifiers then involves posit some general form of model, or form of the classifier, and using training patterns to learn or estimate the unknown parameters of the model. Learning refers to some form of algorithm for reducing the error on a set of training data. Learning comes in several general forms.

#### *3.5.1 Supervised Learning*

In supervised learning, a teacher provides a category label or cost for each pattern in a training set, and one needs to reduce the sum of the costs for these patterns. The learning algorithm should be powerful enough to learn the solution to a given problem and stable to parameter variations. It should converge in finite time, and scale reasonably with the number of training patterns, the number of input features and with the perplexity of the problem. The learning algorithm should appropriately favor “simple” solutions (as in figure 3.6) rather than complicated ones (as in figure 3.5).

### *3.5.2 Unsupervised Learning*

In unsupervised learning or clustering, there is no explicit teacher, and the system forms clusters or “natural groupings” of the input patterns. “Natural” is always defined explicitly or implicitly in the clustering system itself, and given a particular set of patterns or cost function, different clustering algorithms lead to different clusters. Hence a designer should avoid inappropriate representations.

### *3.5.3 Reinforcement Learning*

In reinforcement learning or learning with a critic, no desired category signal is given; instead, the only teaching feedback is that the tentative category is right or wrong. This is analogous to a critic who merely states that something is right or wrong, but does not say specifically how it is wrong. (Thus only binary feedback is given to the classifier; reinforcement learning also describes the case where a single scalar signal, say some number between 0 and 1, is given by the teacher.) In pattern classification, it is most common that such reinforcement is binary - either the tentative decision is correct or it is not. Naturally, if the problem involves just two categories and equal costs for errors, then learning with a critic is equivalent to standard supervised learning.

## *3.6 Conclusions*

The number, complexity and magnitude of these sub-problems are overwhelming. Further, these sub-problems are rarely addressed in isolation and they are invariably interrelated. Thus for instance in seeking to reduce the complexity of a classifier, the designer might affect its ability to deal with invariance. It should be pointed out, though, that the good news is at least three-fold: 1) there is an “existence proof” that many of these problems can indeed be solved - as demonstrated by humans and other biological systems, 2) mathematical theories solving some of these problems have in fact been discovered, and finally 3) there remain many fascinating unsolved problems providing opportunities for progress.

*Chapter 4*

# Committee Machines

In chapter two, the potential of combined systems to obtain good solutions for the problem of face recognition is presented. The difficulties facing any classification system are presented in chapter three. In combined systems, the results from different classifiers are combined to improve the overall performance. These classifiers along with the combination mechanism construct what is called a committee machine. In this chapter, the basic ideas behind committee machines are presented. Section 4.1 is an introduction. In section 4.2, some of the most important architectures and algorithms for committee machines are described. In section 4.3, the reasons for using committee machines are discussed. Some open problems that need to be solved are presented in section 4.4. Finally, the conclusions of this chapter are presented in section 4.5.

## *4.1 Introduction*

During the past decade, the method of committee machines was firmly established as a practical and effective solution for difficult pattern recognition tasks. As stated in [41], the idea appeared under many names: combined classifiers, multiple classifier systems, hybrid methods, decision combination, multiple experts, mixture of experts, classifier ensembles, cooperative agents, opinion pool, sensor fusion, and more. In committee machines, an ensemble of estimators is generated by means of a learning process and the prediction of the committee for a new input is generated in form of a combination of the predictions of the individual committee members. Three reasons are given in [41] for the usefulness of committee machines. First, the committee might exhibit a test set performance unobtainable by an individual committee member on its own. The reason is that the errors of the individual committee members cancel out to some degree when their predictions are combined. The surprising discovery of this line of research is that even if the committee members were trained on disturbed versions of the same data set, the predictions of the individual committee members might be sufficiently different such that this averaging process takes place and is beneficial. A second reason for using committee machines is modularity. It is sometimes beneficial if a mapping from input to target is not approximated by one estimator but by several estimators, where each estimator can focus on a particular region in input space. The prediction of the committee is obtained by a locally weighted combination of the predictions of the committee members. It could be shown that in some applications the individual members self-organize in a way such that the prediction task is divided into meaningful modules. The most important representatives of this line of research are the mixture of experts approach and its variants. A third reason for using committee machines is a reduction in computational complexity. Instead of training one estimator using all training data it is computationally more efficient for some type of estimators to partition the data set into several data sets, train different estimators on the individual data sets and then combine the predictions of the individual estimators. Typical examples of estimators for which this procedure is beneficial are Gaussian process regression, kriging, regularization neural networks, smoothing splines, and the support vector machine, since for those systems, training time increases drastically with increasing training data set size. By using a committee machine approach, the computational complexity increases only linearly with the size of the training data set.

### *4.2 Constructing Committee Machines*

As stated in [42] algorithms used to construct committee machines could be divided into two broad categories: generative and non-generative algorithms. Generative algorithms generate sets of base learners acting on the base learning algorithm or on the structure of the data set and try to actively improve diversity and accuracy of the base learners. On the other hand, non-generative algorithms confine themselves to combine a set of given possibly well-designed base learners; they do not actively generate new base learners but try to combine in a suitable way a set of existing base classifiers.

### *4.2.1 Generative Algorithms*

Generative ensemble methods try to improve the overall accuracy of the ensemble by directly boosting the accuracy and the diversity of the base learners [42]. They can modify the structure and the characteristics of the available input data, as in resampling methods or in feature selection methods, they can manipulate the aggregation of the classes (Output Coding methods), can select base learners specialized for a specific input region (mixture of experts methods), can select a proper set of base learners evaluating the performance and the characteristics of the component base learners (test-and-select methods), or can randomly modify the base learning algorithm (randomized methods). Referring to [42], the following is a review of generative algorithms.

#### *4.2.1.1 Resampling Methods*

Resampling techniques can be used to generate different hypotheses. For instance, bootstrapping techniques may be used to generate different training sets and a learning algorithm can be applied to the obtained subsets of data in order to produce multiple hypotheses. These techniques are effective especially with unstable learning algorithms, which are algorithms very sensitive to small changes in the training data, such as neural-networks and decision trees. In bagging, the ensemble is formed by making bootstrap replicates of the training sets, and then multiple generated hypotheses are used to get an aggregated predictor. The aggregation can be performed by averaging the outputs in regression or by majority or weighted voting in classification problems. While in bagging the samples are drawn with replacement using a uniform probability distribution, in boosting methods the learning algorithm is called at each iteration using a different distribution or weighting over the training examples. This technique places the highest weight on the examples most often misclassified by the previous base learner: in this way, the base learner focuses its attention on the hardest examples. Then the boosting algorithm combines the base rules taking a weighted majority vote of the base rules. It was shown that the training error exponentially drops down with the number of iterations. Experimental work showed that bagging is effective with noisy data, while boosting, concentrating its efforts on noisy data, seems to be very sensitive to noise.

#### *4.2.1.2 Feature Selection Methods*

This approach consists in reducing the number of input features of the base learners, a simple method to fight the effects of the classical curse of dimensionality problem. For instance, in the Random Subspace Method, a subset of features is randomly selected and assigned to an arbitrary learning algorithm. This way, one obtains a random subspace of the original feature space, and constructs classifiers inside this reduced subspace. The aggregation is usually performed using weighted voting on the basis of the base classifiers' accuracy. It has been shown that this method is effective for classifiers having a decreasing learning curve constructed on small and critical training sample sizes.

#### *4.2.1.3 Mixtures of Experts Methods*

The recombination of the base learners can be governed by a supervisor-learning machine, which selects the most appropriate element of the ensemble based on the available input data. This idea led to the mixture of experts methods, where a gating network performs the division of the input space and small neural networks perform the effective calculation at each assigned region separately. An extension of this approach is the hierarchical mixture of experts method, where the outputs of the different experts are non-linearly combined by different supervisor gating networks hierarchically organized.

#### *4.2.1.4 Output Coding Decomposition Methods*

Output Coding (OC) methods decompose a multi-class classification problem in a set of two-class sub-problems, and then recompose the original problem combining them to achieve the class label. An equivalent way of thinking about these methods consists in encoding each class as a bit string (named codeword), and in training a different two-class base learner (dichotomizer) in order to separately learn each codeword bit. When the dichotomizers are applied to classify new points, a suitable measure of similarity between the codeword computed by the ensemble and the codeword classes is used to predict the class. Error Correcting Output Coding is the most studied OC method, and has been successfully applied to several classification problems. This decomposition method tries to improve the error correcting capabilities of the codes generated by the decomposition through the maximization of the minimum distance between each couple of codewords. This goal is achieved by means of the redundancy of the coding scheme.

#### *4.2.1.5 Test and Select Methods*

The test and select methodology relies on the idea of selection in ensemble creation. The simplest approach is a greedy one, where a new learner is added to the ensemble only if the resulting squared error is reduced, but in principle, any optimization technique can be used to select the "best" component of the ensemble, including genetic algorithms. It should be noted that the time complexity of the selection of optimal subsets of classifiers is exponential with respect to the number of base learners used. From this point of view heuristic rules, as the "choose the best" or the "choose the best in the class", using classifiers of different types strongly reduce the computational complexity of the selected phase, as the evaluation of different classifier subsets is not required. Moreover, test and select methods implicitly include a "production stage", by which a set of classifiers must be generated. Another interesting approach uses clustering methods and a measure of diversity to generate sets of diverse classifiers combined by majority voting, selecting the ensemble with the highest performance.

#### *4.2.1.6 Randomized Ensemble Methods*

Injecting randomness into the learning algorithm is another general method to generate ensembles of learning machines. For instance, if the weights in the back-propagation algorithm are initialize with random values, different learning machines will be obtained that can be combined into an ensemble. Several experimental results showed that randomized learning algorithms used to generate base elements of ensembles improve the performances of single non-randomized classifiers.

### *4.2.2 Non- Generative Algorithms*

As stated in [42], this large group of ensemble methods embraces a large set of different approaches to combine learning machines. They share the very general common property of using a predetermined set of learning machines previously trained with suitable algorithms. The base learners are then put together by a combiner module that may vary depending on its adaptivity to the input patterns and on the requirement of the output of the individual learning machines. The type of combination may depend on the type of output. If only labels are available or if continuous outputs are hardened, then majority voting, that is the class most represented among the base classifiers, is used [43]. This approach can be refined, assuming mutual independence between classifiers, using a Bayesian decision rule that selects the class with the highest posterior probability computed through the estimated class conditional probabilities and the Bayes formula ([44] and [45]). To overcome the problem of the independence assumption (that is unrealistic in most cases), the Behavior-Knowledge Space (BKS) method [46] considers each possible combination of class labels, filling a look-up table using the available data set, but this technique requires a huge volume of training data. Where the classifier outputs are interpreted as the support for the classes, fuzzy aggregation methods can be applied, such as simple connectives between fuzzy sets or the fuzzy integral ([47], [48], [49], and [50]). Statistical methods and similarity measures to estimate classifier correlation have also been used to evaluate expert system combination for a proper design of multi-expert systems [51]. The base learners can also be aggregated using simple operators as Minimum, Maximum, Average and Product and Ordered Weight Averaging. Another general approach consists in explicitly training combining rules, using second-level learning machines on top of the set of the base learners [52]. This stacked structure makes use of the outputs of the base learners as features in the intermediate space: the outputs are fed into a second-level machine to perform a trained combination of the base learners. In [53], six classifier fusion methods are studied theoretically: minimum, maximum, average, median, majority vote, and the oracle. It was found that if the classifiers' error distribution were uniform, the six fusion methods would decrease the overall error rate by different ratios. This contradicts the common literature claim that combination methods are less important than the diversity of the team.

### *4.3 Why Committee Machines Work*

Researchers give many explanations for this matter. For example in [54] the author states that uncorrelated errors made by the individual classifiers can be removed by voting. There are at least three reasons why good ensembles can be constructed (according to [54]) and why it may be difficult or impossible to find a single classifier that performs as well as the ensemble. To understand these reasons, one must consider the nature of machine learning algorithms. Machine learning algorithms work by searching a space of possible hypotheses *H* for the most accurate hypothesis (that is, the hypothesis that best approximates the unknown function *f*). Two important aspects of the hypothesis space *H* are its size and whether it contains good approximations to *f*. If the hypothesis space is large, then a large amount of training data is needed to constrain the search for good approximations. Each training example rules out (or makes less plausible) all those hypotheses in *H* that misclassify it. In a two-class problem, ideally each training example can eliminate half of the hypotheses in *H*, so O(log(*H*)) examples are required to select a unique classifier from *H*. The first "cause" of the need for ensembles is that the training data may not provide sufficient information for choosing a single best classifier from *H*. Most of the learning algorithms consider very large hypothesis spaces, so even after eliminating hypotheses that misclassify training examples, there are many hypotheses remaining. All of these hypotheses appear equally accurate with respect to the available training data. One may have reasons for preferring some of these hypotheses to others (e.g., preferring simpler hypotheses or hypotheses with higher prior probability), but nonetheless, there are typically many plausible hypotheses. From this collection of surviving hypothesis in *H*, an ensemble of classifiers could be easily constructed and combined using the methods described above. A second "cause" of the need for ensembles is that the learning algorithms may not be able to solve the difficult search problems that are posed. For example, the problem of finding the smallest decision tree that is consistent with a set of training examples is NP-hard. Hence, practical decision tree algorithms employ search heuristics to guide a greedy search for small decision trees. Similarly, finding the weights for the smallest possible neural network consistent with the training examples is also NP-hard. Neural network algorithms therefore employ local search methods (such as gradient descent) to find locally optimal weights for the network. A consequence of these imperfect search algorithms is that even if the combination of the training examples and prior knowledge (e.g., preferences for simple hypotheses, Bayesian priors) determines a unique best hypothesis, one may not be able to find it. Instead, one will typically find a hypothesis that is somewhat more complex (or has somewhat lower posterior probability). If the search algorithms are run with a slightly different training sample or injected noise (or any of the other techniques described above), a different (sub-optimal) hypothesis will be found. Ensembles can be seen therefore as a way of compensating for imperfect search algorithms. A third "cause" of the need for ensembles is that the hypothesis space *H* may not contain the true function *f*. Instead, *H* may include several equally good approximations to *f*. By taking weighted combinations of these approximations, one might be able to represent classifiers that lie outside of *H*.

### *4.4 Open Problems in Committee Machines*

As stated in [54], there are still many questions about the best way to construct ensembles as well as issues about how best to understand the decisions made by ensembles. In principle, there can be no single best ensemble method, just as there can be no single best learning algorithm. However, some methods may be uniformly better than others. In addition, some methods may be better than others in certain situations. There have been very few systematic studies of methods for constructing ensembles of neural networks, rule-learning systems, and other types of classifiers. Much work remains in this area. While ensembles provide very accurate classifiers, there are problems that may limit their practical application. One problem is that ensembles can require large amounts of memory to store and large amounts of computation to apply. An important line of research, therefore, is to find ways of converting these ensembles into less redundant representations, perhaps by deleting highly correlated members of the ensemble or by representational transformations. A second difficulty with ensemble classifiers is that an ensemble provides little insight into how it makes its decisions. A single decision tree can often be interpreted by human users, but an ensemble of 200 voted decision trees is much more difficult to understand. It would be helpful if methods can be found for obtaining explanations (at least locally) from ensembles.

### *4.5 Conclusions*

From the discussion in this chapter, the following could be concluded:

- The basic idea behind committee machines is not a new one, it is seen everywhere in our daily lives. Humans have been using committees throughout human history (parliaments, boards of directors, principle of “Shura” in Islam, …etc.).
- Committee machines techniques are capable of obtaining better solutions than individual classifiers in many classification problems.
- There is no universal agreement on how component classifiers should be selected, trained or combined to produce a successful committee machine for a certain application.

Many open problems still need to be solved. Nevertheless, many applications have already appeared.

*Chapter 5*

# The Proposed Algorithm

In this chapter, a through description of the proposed algorithm is presented. Section 5.1 generally describes the system and its operation during training and classification. Section 5.2 introduces LVQ neural networks. Section 5.3 describes methods to enhance the system performance. Section 5.4 describes the whole algorithm in detail. Section 5.5 illustrates the relation of the proposed system to other combined systems. Finally, section 5.6 presents the conclusions of this chapter.

## *5.1 The Proposed Solution*

The solution is based on the idea of combining the decisions of multiple classifiers, each of which is trained on a bootstrap (subset) of the training data. In order for the combination to be effective, the classifiers should have the following characteristics [24]:

- The classifiers should be efficient during training and classification in both time and storage requirements.
- The classifiers should be independent in the errors they make during classification.
- Each classifier should be accurate in classifying the patterns on which it was trained.

The first condition is necessary to make the resulting system feasible. The second condition means that the classifiers should be trained to view different characteristics of the input space so that combining their decision will be meaningful. If the decisions of classifiers that make the same mistakes are combined, the combination is meaningless. This condition is sometimes called classifiers' diversity. The third condition means that each classifier performs better than random guessing. It does not imply a certain recognition rate for the individual classifier. If the classifiers satisfy the second and third conditions, the committee should perform better than any individual classifier.

### *5.1.1 Training Phase*

The training set contains (n) training images for each one of the (M) classes (persons) to be classified. This results in a total of (n M) input patterns. From the (M) classes, (N) bootstraps (subclasses) are extracted. Each bootstrap contains (m) classes, where (1 < m < M). The (N) bootstraps are intended to have the following properties:

1) Each class of the (M) classes must be included at least in one bootstrap.
2) The classes present in each bootstrap are all different from each other.
3) The number of bootstraps including a certain class must be the same for all classes.
4) No two bootstraps are allowed to contain exactly the same classes.

The purpose of these rules is to ensure that each class has a fair chance of training as any other class. The other alternative, which is uniform random selection, will not guarantee fairness unless applied to a large number of classifiers, which is not always the case. Figure 5.1 represents the system during training.

As an example, 8 classes (numerically labeled for simplicity) are given: {0, 1, 2, 3, 4, 5, 6, 7}. Four bootstraps are to be defined on these classes: B1, B2, B3, and B4. The following are two acceptable configurations:

1) B1 = {0, 1, 2, 3}, B2 = {2, 3, 4, 5}, B3 = {4, 5, 6, 7}, B4 = {6, 7, 0, 1}
2) B1 = {0, 1}, B2 = {2, 3}, B3 = {4, 5}, B4 = {6, 7}

Whereas the following are four illegal configurations:

1) B1 = {0, 1, 2, 3}, B2 = {1, 2, 3, 4}, B3 = {2, 3, 4, 5}, B4 = {3, 4, 5, 6} (Class 7 not present in any bootstrap)

2) B1 = {0, 1, 2, 2}, B2 = {2, 3, 4, 5}, B3 = {4, 5, 6, 7}, B4 = {6, 7, 0, 1} (Repeated class in B1)

3) B1 = {0, 1, 2, 3}, B2 = {3, 4, 5, 6}, B3 = {6, 7, 0, 1}, B4 = {1, 2, 3, 4} (Class 1 is present in 3 bootstraps while class 2 is present in 2 bootstraps)

4) B1 = {0, 1, 2, 3}, B2 = {3, 2, 1, 0}, B3 = {6, 7, 0, 1}, B4 = {1, 2, 3, 4} (Bootstraps B1 and B2 are the same)

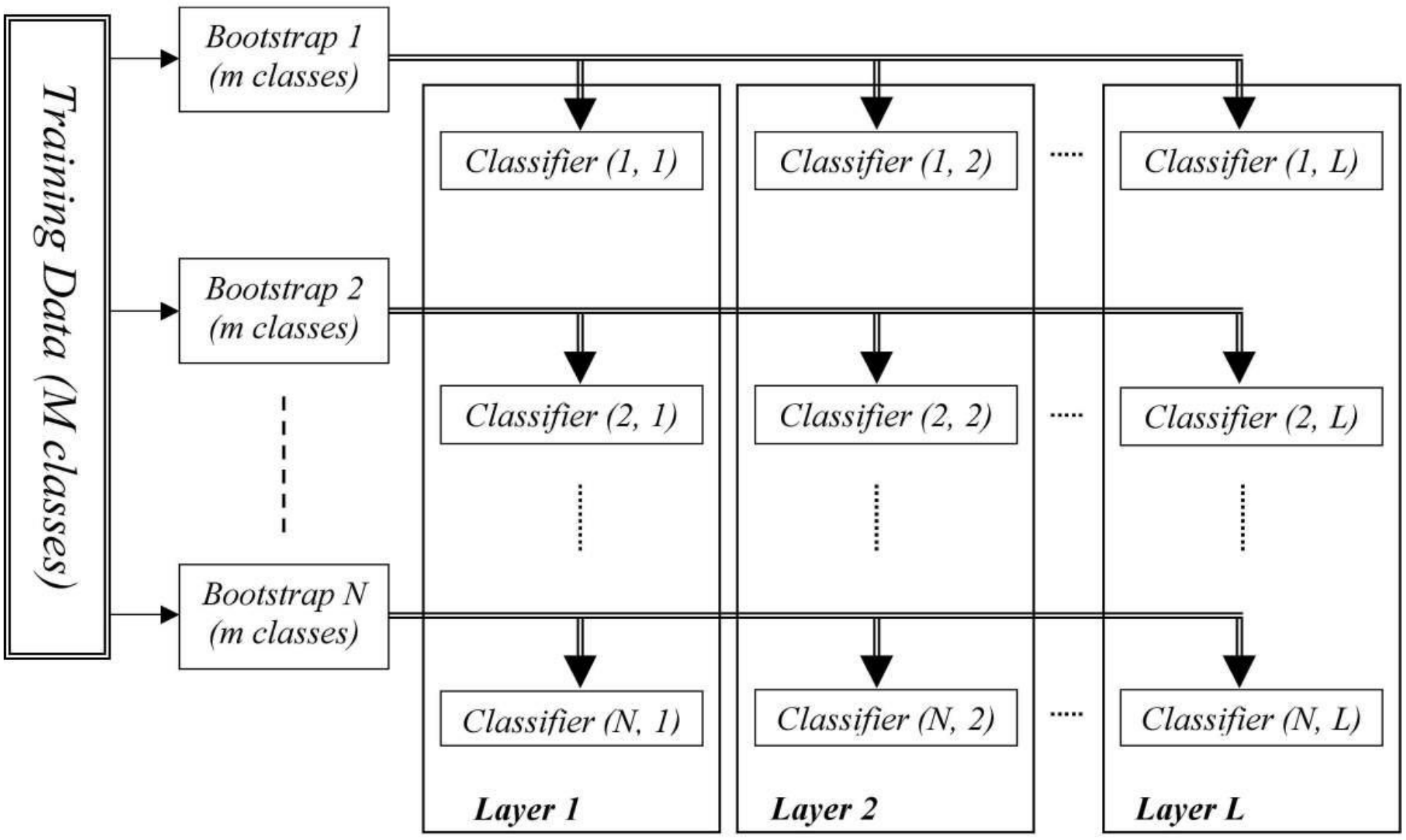

Figure 5.1. The proposed system during training

It is clear that any two bootstraps may or may not contain overlapping classes. An overlap parameter (V) is defined as the maximum number of identical classes present in any two bootstraps for the selected configuration. Hence $0 \leq V < m$ (the 4th condition prevents V from being equal to m). For example in the following configuration, V equals 2:

B1 = {0, 1, 2, 3}, B2 = {2, 3, 4, 5}, B3 = {4, 5, 6, 7}, B4 = {6, 7, 0, 1}

While in this one, V is zero:

B1 = {0, 1}, B2 = {2, 3}, B3 = {4, 5}, B4 = {6, 7}

After selecting the bootstraps, the K classifiers that construct the committee are trained as follows:

- L classifiers are trained on each bootstrap hence constructing L layers of classifiers.
- For any two classifiers trained on the same bootstrap (they will naturally be in different layers), they are trained with different initial conditions and parameters to introduce classifiers' diversity.
- Each classifier is trained on its bootstrap such that its recognition rate is better than random guessing.

Having the values of M, m, L, and V, the following important quantities can be deduced:

- The total number of bootstraps: $N = M / (m - V)$ …(5.1)
- The total number of classifiers: $K = L N = L M / (m - V)$ …(5.2)
- The total number of classifiers trained on any single class:

$$R = L m / (m - V) \quad \text{…(5.3)}$$

Since L, M, N, m, V, K and R are all integers the following must hold: The quantities M and m must be divisible by (m - V). Thus, not any configuration is possible. For example for (M = 6), only the seven bootstrap configurations shown in table 5.1 are possible.

The total number of patterns in the training set is (n M). Each classifier is trained on (n m) patterns. Assuming a classifier correctly recognizes a total of (r) patterns from the (n m) patterns it was trained on ($r \leq n\, m$), the actual recognition rate of any classifier is thus equal to:

$$E = r / (n M) \quad \text{…(5.4)}$$

The value of E must be better than the recognition rate of random guessing. Since any classifier will be used to classify M classes, random guessing will be correct in classifying (1 / M) of the patterns (assuming the input patterns have a uniform distribution for their probability of occurrence) and hence:

$$E > 1 / M \text{ or equivalently } r > n \qquad \ldots(5.5)$$

Table 5.1. Possible configurations for M = 6 classes

| | M | m | V | m – V | N | K | R | Example |
|---|---|---|---|---|---|---|---|---|
| 1) | 6 | 2 | 1 | 1 | 6 | 6 L | 2 L | B1 = {0, 1}, B2 = {1, 2}, B3 = {2, 3}, B4 = {3, 4}, B5 = {4, 5}, B6 = {5, 0} |
| 2) | 6 | 3 | 2 | 1 | 6 | 6 L | 3 L | B1 = {0, 1, 2}, B2 = {1, 2, 3}, B3 = {2, 3, 4}, B4 = {3, 4, 5}, B5 = {4, 5, 0}, B6 = {5, 0, 1} |
| 3) | 6 | 4 | 3 | 1 | 6 | 6 L | 4 L | B1 = {0, 1, 2, 3}, B2 = {1, 2, 3, 4}, B3 = {2, 3, 4, 5}, B4 = {3, 4, 5, 0}, B5 = {4, 5, 0, 1}, B6 = {5, 0, 1, 2} |
| 4) | 6 | 5 | 4 | 1 | 6 | 6 L | 5 L | B1 = {0, 1, 2, 3, 4}, B2 = {1, 2, 3, 4, 5}, B3 = {2, 3, 4, 5, 0}, B4 = {3, 4, 5, 0, 1}, B5 = {4, 5, 0, 1, 2}, B6 = {5, 0, 1, 2, 3} |
| 5) | 6 | 2 | 0 | 2 | 3 | 3 L | L | B1 = {0, 1}, B2 = {2, 3}, B3 = {4, 5} |
| 6) | 6 | 4 | 2 | 2 | 3 | 3 L | 2 L | B1 = {0, 1, 2, 3}, B2 = {2, 3, 4, 5}, B3 = {4, 5, 0, 1} |
| 7) | 6 | 3 | 0 | 3 | 2 | 2 L | L | B1 = {0, 1, 2}, B2 = {3, 4, 5} |

Constructing the bootstraps can be accomplished as follows: Assuming the M classes are numerically labeled as {0, 1, …, M - 1} and the bootstraps are labeled: B1, B2, …, BN. A parameter (h) is called the shift parameter and is defined such that $0 < h \leq m < M$. The bootstraps are constructed as follows:

$B1 = \{b_1(0), b_1(1), b_1(2), \ldots, b_1(m - 1)\}$,

$B2 = \{b_2(0), b_2(1), b_2(2), \ldots, b_2(m - 1)\}$,

.

.

.

$BN = \{b_N(0), b_N(1), b_N(2), \ldots, b_N(m - 1)\}$ …(5.6)

Where:

$b_1(0) = 0, b_k(t + 1) = rem[b_k(t) + 1, M], t = 0, 1, \ldots, (m - 1)$ …(5.7)

$b_{k+1}(0) = rem[(h + b_k (0)), M] = rem[(k - 1) h, M],$

$k = 1, 2, \ldots, (N - 1);$ …(5.8)

And rem[x, y] is the remainder of dividing the integer x by the integer y.

For example let M = 8, m = 5 and h = 3. Then 8 bootstraps are obtained:

B1 = {0, 1, 2, 3, 4, 5}

B2 = {3, 4, 5, 6, 7, 0}

B3 = {6, 7, 0, 1, 2, 3}

B4 = {1, 2, 3, 4, 5, 6}

B5 = {4, 5, 6, 7, 0, 1}

B6 = {7, 0, 1, 2, 3, 4}

B7 = {2, 3, 4, 5, 6, 7}

B8 = {5, 6, 7, 0, 1, 2}

This configuration can be simply rearranged to become:

B1 = {0, 1, 2, 3, 4, 5}

B2 = {1, 2, 3, 4, 5, 6}

B3 = {2, 3, 4, 5, 6, 7}

B4 = {3, 4, 5, 6, 7, 0}

B5 = {4, 5, 6, 7, 0, 1}

B6 = {5, 6, 7, 0, 1, 2}

B7 = {6, 7, 0, 1, 2, 3}

B8 = {7, 0, 1, 2, 3, 4}

The last configuration can be obtained by making h = 1 rather than h = 3. This implies the fact that different values of h may produce similar bootstrap configurations. The reason for this will be given shortly. From the above method for constructing the bootstraps it can easily be seen that

$$V = m - h \text{ or equivalently } h = m - V \qquad \ldots(5.9)$$

If one more bootstrap is added using the same shift (h = 3 before rearranging or h = 1 after rearranging) the first bootstrap is regenerated again (B9 = B1) which is forbidden and so the bootstrap generation process is stopped at B8. It is clear that this configuration naturally satisfies the 2nd condition because (m < M). In order to satisfy the 1st, 3rd and 4th conditions, N is selected such that:

$$N = M / \gcd[M, h] \qquad \ldots(5.10)$$

where gcd[x, y] is the greatest common divisor of the integers x and y. The number of steps required to make $b_k = 0$ again (where k > 1) is equal to M / gcd[M, h]. This means that if N > M / gcd[M, h] the sequence {0, 1, …, m - 1} will repeat itself for k = 1 + M / gcd[M, h], 1 + 2 M / gcd[M, h], …, etc. On the other hand, if N < M / gcd[M, h] some classes may not be present in any bootstrap. Hence N = M / gcd[M, h] is the only allowable value for N. The total number of classifiers trained on any single class is equal to:

$$R = L\, m\, N / M = L\, m / \gcd[M, h] \qquad \ldots(5.11)$$

Hence the 3rd condition and the fact that L m / gcd[M, h] must be an integer can only be satisfied if rem(m, gcd[M, h]) = 0. Since (m) is divisible by gcd[M, h] then by letting s = gcd[M, h] and from the properties of the gcd:

$$s = \gcd[M, h] = \gcd[M, m, h] \leq h \qquad \ldots(5.12)$$

Replacing gcd[M, h] by (s) the final relations are obtained:

$$N = M / s,\ K = L\, M / s,\ R = L\, m / s. \text{ and } s < h \qquad \ldots(5.13)$$

It is noted that (s) can be viewed as a shift parameter just like (h). Hence (V = m – s) is a valid relation. The meaning of this is that taking (h) as the shift parameter may produce similar configurations for dissimilar combinations of (M, m, h). However, taking (s) as the shift parameter will always produce dissimilar configurations for dissimilar combination of (M, m, s). This can be seen from the previous example by comparing the combination resulting from (M = 8, m = 5, h = 3) with the combination resulting from (M = 8, m = 5, h = 1). In both cases:

$$s = gcd[8, 5, 3] = gcd[8, 5, 1] = 1 \qquad \ldots(5.14)$$

### *5.1.2 Testing / Classification Phase*

During the testing phase (and naturally during the normal operation of the system) some input images that need to be classified are given. Each image is presented to all of the K classifiers and hence K decisions (one from each classifier) are to be combined. The combination is achieved through plurality voting and the final decision is simply the class mostly voted for by the classifiers.

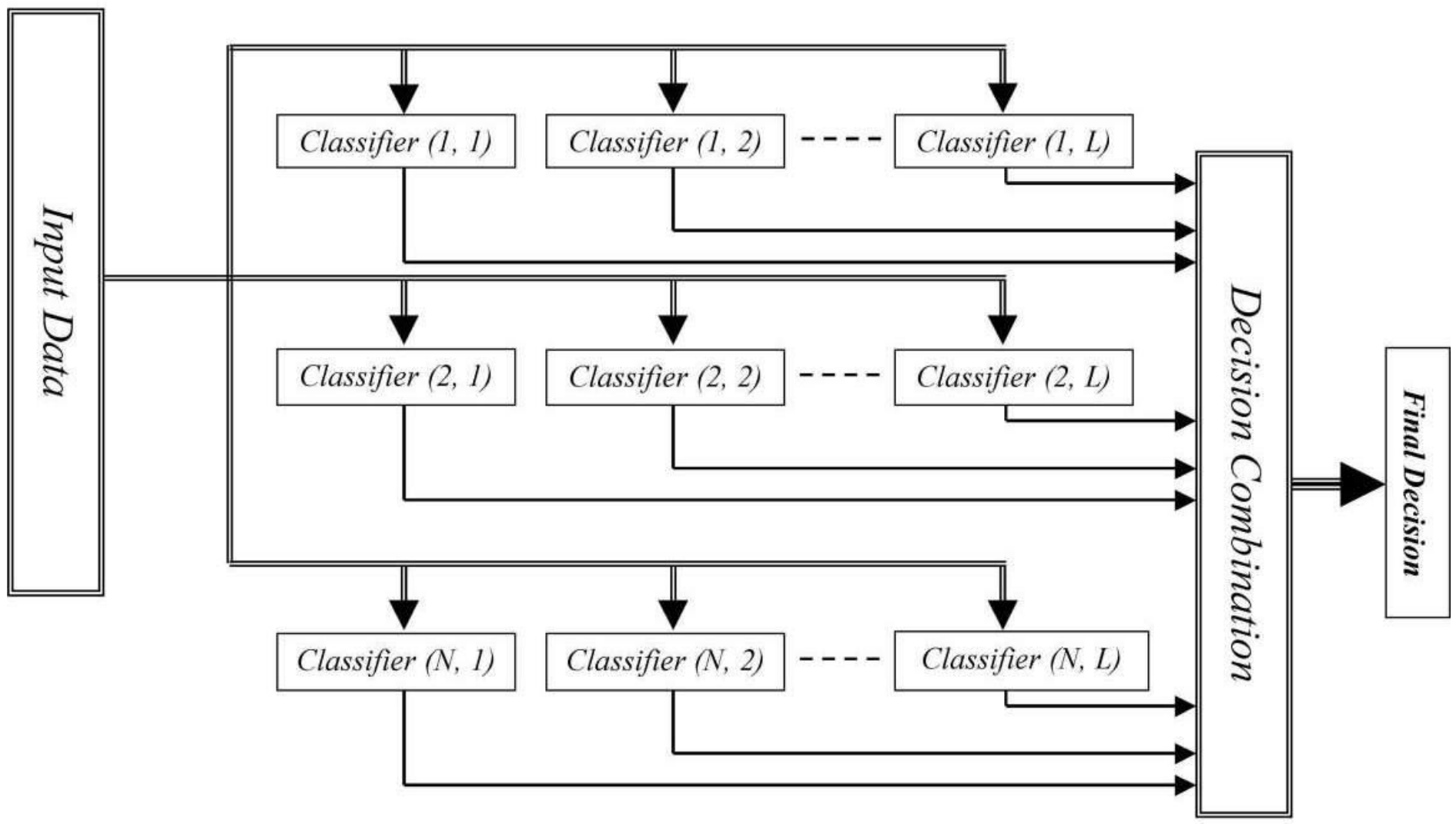

Figure 5.2. The proposed system during classification

For example having 12 classifiers that produced the following votes: {0, 1, 2, 3, 4, 0, 0, 2, 3, 4, 5, 5} then the final classification is class 0 because it got the highest number of votes. The system described above can be considered as a general scheme. The actual type of the underlying classifiers is not the primary concern of the system. So, virtually any classifier can be used. Here LVQ classifiers are used for their simplicity and efficiency during both training and classification.

### *5.2 Learning Vector Quantization (LVQ)*

LVQ is short for Learning Vector Quantization, it is a supervised learning artificial neural network based on competition widely used in pattern classification problems. As stated in [55] it can be described as:

"A pattern classification method in which each output unit represents a particular class or category. (Several output units should be used for each class.) The weight vector for an output unit is often referred to as a reference (or codebook) vector for the class that the unit represents. During training, the output units are positioned (by adjusting their weights through supervised training) to approximate the decision surfaces of the theoretical Bayes classifier. It is assumed that a set of training of reference vectors with known classifications is provided, along with an initial distribution of reference vectors (each of which represents a known classification). After training, an LVQ net classifies an input vector by assigning it to the same class as the output unit that has its weight vector (reference vector) closest to the input vector."

The architecture and training algorithm of LVQ net as described in [55] are shown below.

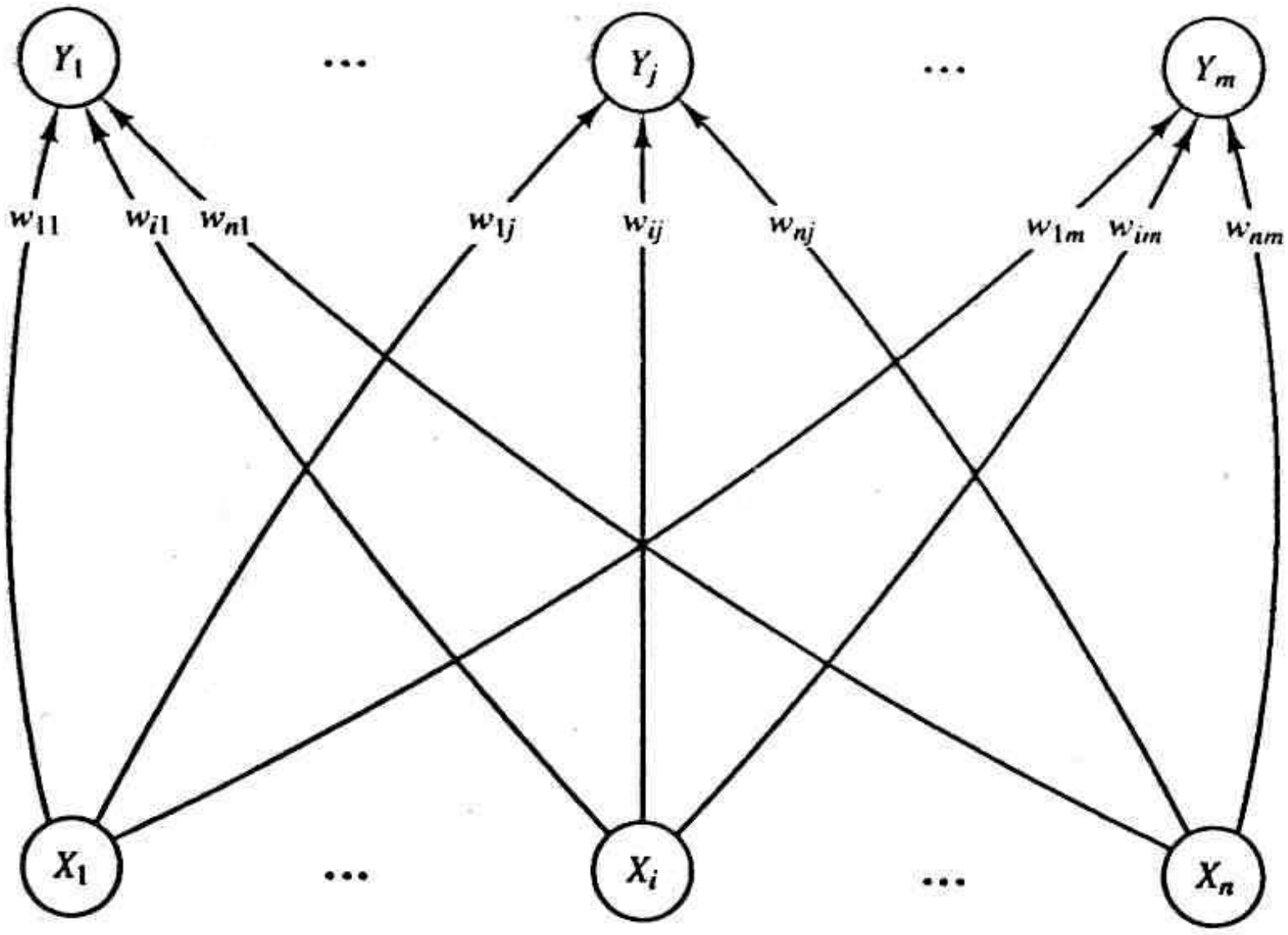

Figure 5.3. LVQ neural network

Algorithm:

| | |
|---|---|
| $a$ | : learning rate, $0 < a < 1$ |
| x | : training vector ($x_1, x_2, \ldots, x_n$). |
| T | : correct category or class for training vector. |
| $w_j$ | : Weight vector for jth output unit ($w_{1j}, w_{2j}, \ldots, w_{nj}$). |
| $C_j$ | : category or class represented by jth output unit. |
| $d(x, w_j)$ | : Distance metric between input vector and weight vector for jth output unit. |

Step 1: Initialize reference vectors.

Step 2: While stopping condition is false, do steps 3 to 7

Step 3: For each training input vector x, do steps 4 to 5

Step 4: Find J so that $d(x, w_J)$ is minimum. (weight $w_J$ is fired by the input pattern)

Step 5: Update $w_J$ as follows:

if $T = C_J$ then

$$w_J(new) = w_J(old) + a\,[x - w_J(old)]$$

else

$$w_J(new) = w_J(old) - a\,[x - w_J(old)]$$

Step 6: Reduce learning rate *a*

Step 7: Test stopping condition (may be a fixed number of iterations or a desired error rate)

As an example, assuming 60 2-dimensional training inputs are given, each point is on the form (x1, x2). The points represent 4 classes as shown in figure 5.4.a. It can be seen from figure 5.4.a that class 1 (triangles) is represented by 20 inputs, class 2 (circles) by 20 other inputs, class 3 (stars) by 10 other inputs and class 4 (diamonds) by the remaining 10 inputs. The first step in the algorithm is to initialize the weights of the LVQ network. Six weight vectors are randomly placed initially as shown in figure 5.4.b. The weights are assigned classes such that they represent the same ratios of the inputs (2 weights for each of the classes 1 and 2, and 1 weight for each of the classes 3 and 4). The network is trained with a learning rate of 0.01 and a number of training epochs (iterations) of 100. In figure 5.4.c to 5.4.h, the 6 weights are shown as they move during training to their final locations. Figure 5.4.i shows the decision boundaries that define the decision regions for the weight vectors. During classification, a 2-D input is compared to all the 6 weight vectors and the input is classified as the class assigned to the weight vector that is nearest to the input. The decision region of any weight vector is the set of points that are nearer to that weight vector than to any of the other weight vectors. From this example, some important properties of LVQ networks can be sensed. The basic idea behind the LVQ algorithm is to train the network to encode the input space using a small number of weight vectors. The weight vectors divide the input space into a similar number of decision regions. Each region represents a certain class as predicted by the LVQ classifier. The number of weights should be sufficient to encode the training inputs. This is not always guaranteed in practical applications because the number of dimensions of the input vectors is usually very high (hundreds or even thousands of dimensions). Hence, the data cannot be visualized in advance to know how many weight vectors are needed. The learning rate and number of training epochs are very important parameters for the network. If the network in the example is trained with a very high learning rate (say near unity), the weight vectors may jump widely and eventually drift away from their correct final positions. On the other hand, if it is too small then the number of training epochs must be increased to the extent that it might be impractical to train the network in reasonable time. It is clear that the performance of LVQ networks is usually very sensitive to these parameters and in order to reach a very high classification rate (near 100%) using a single LVQ network it is required to search for the optimal parameters which is a very long and difficult road to walk. Hence, it is usually much easier to train an LVQ network to correctly classify 75% of the inputs than it is to train it to correctly classify 99% of the same inputs.

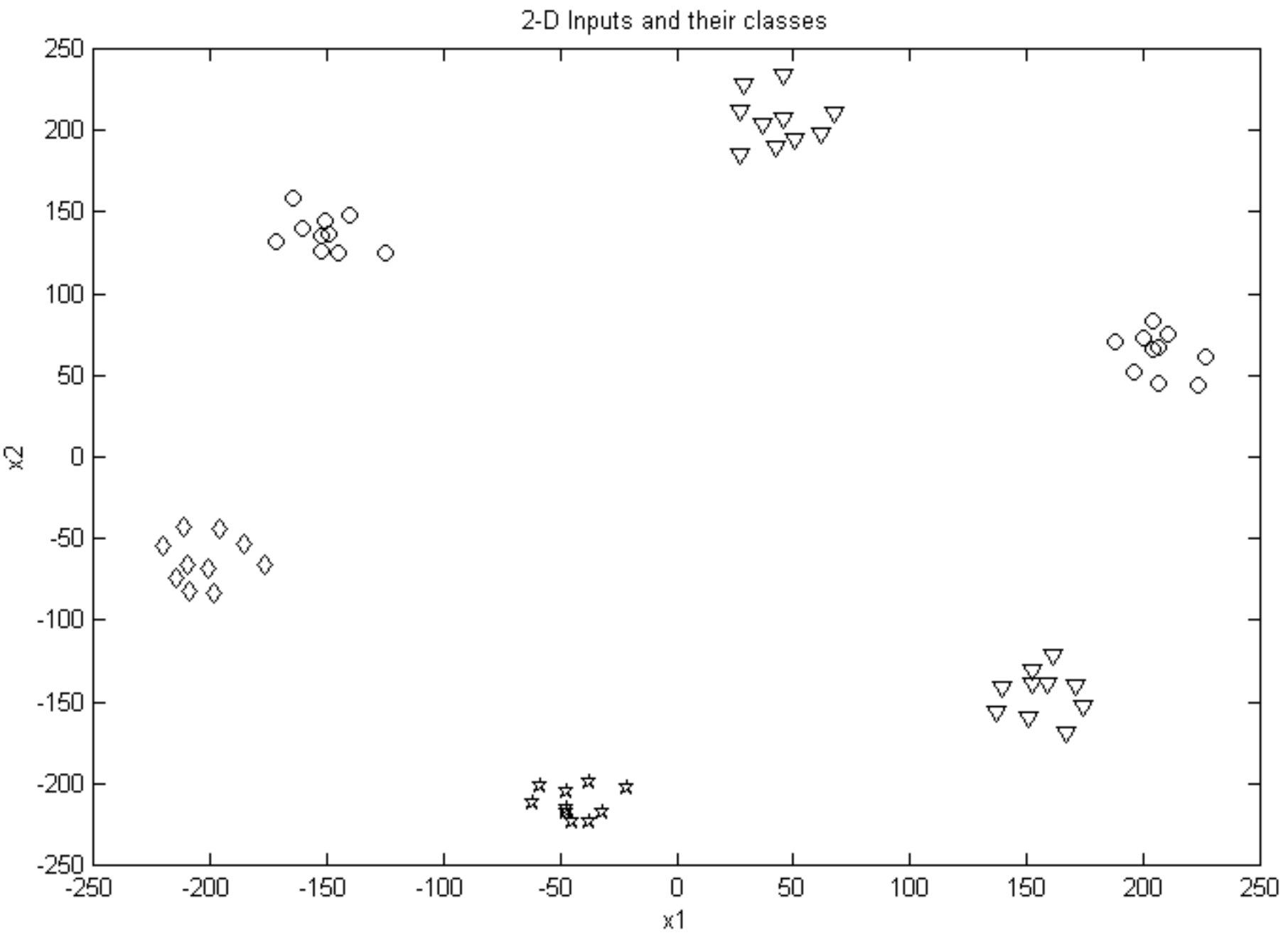

Figure 5.4.a. Training inputs and their corresponding classes

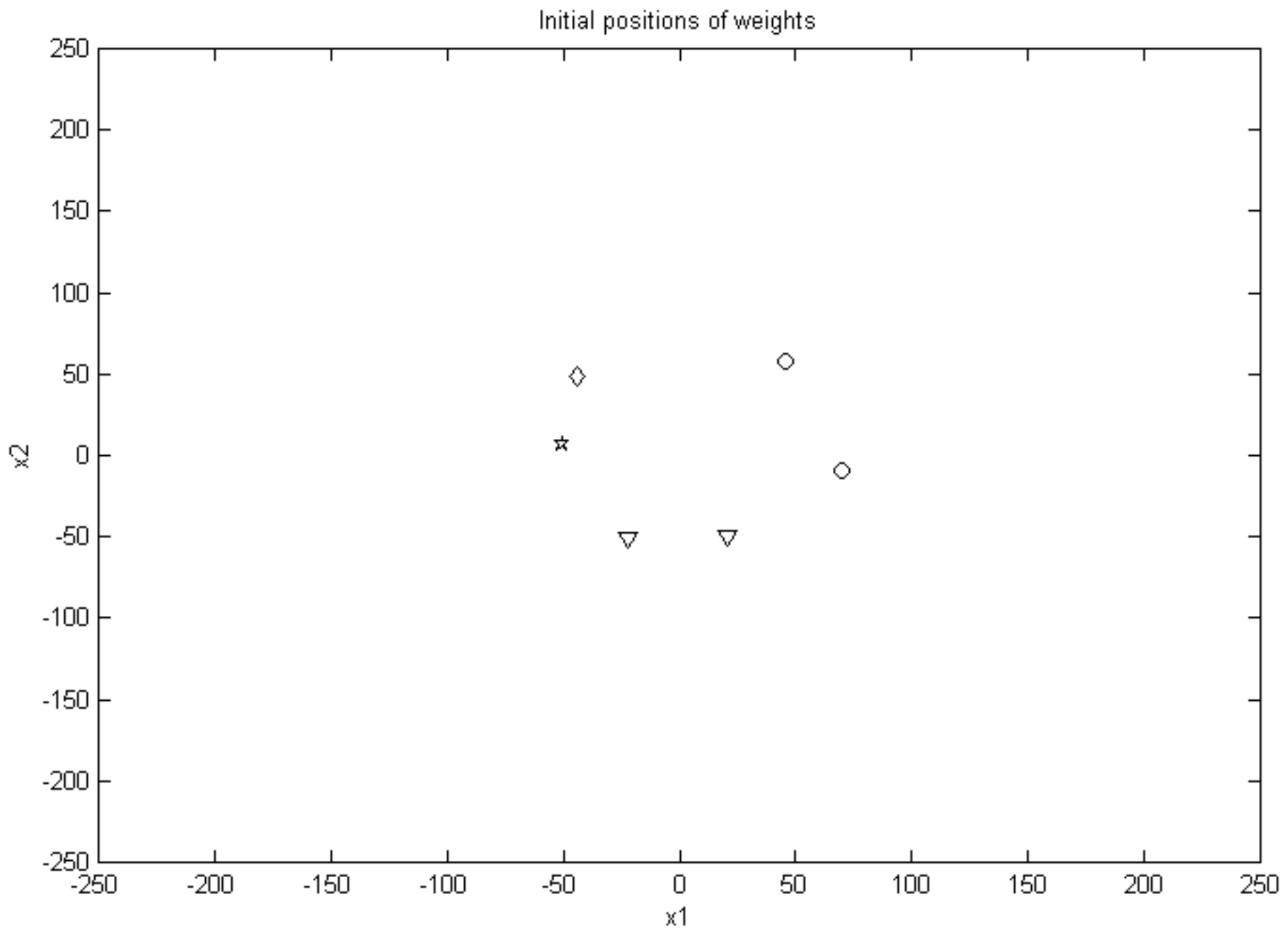

Figure 5.4.b. Initial weight positions for the LVQ network

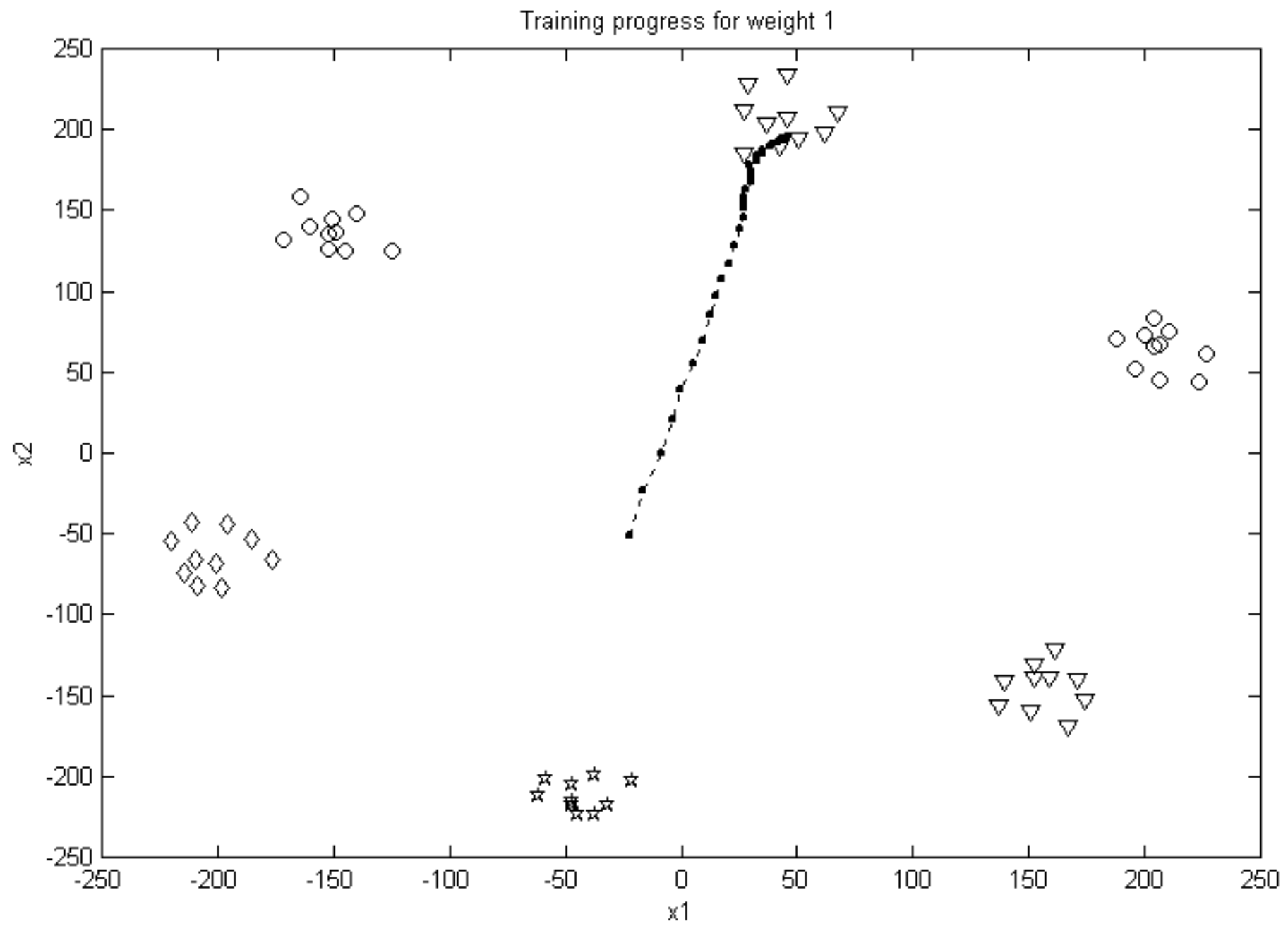

Figure 5.4.c. Training progress for first LVQ weight

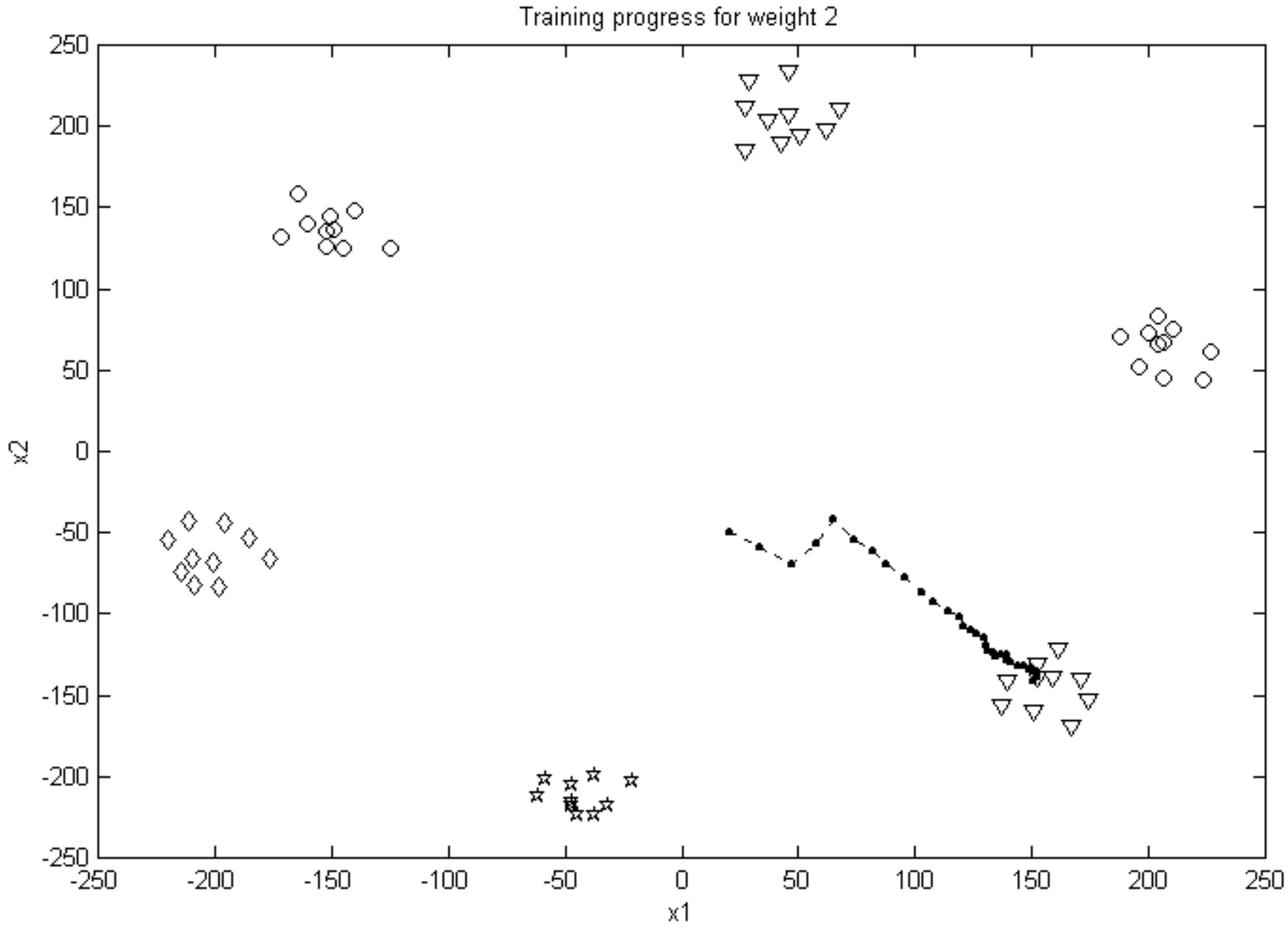

Figure 5.4.d. Training progress for second LVQ weight

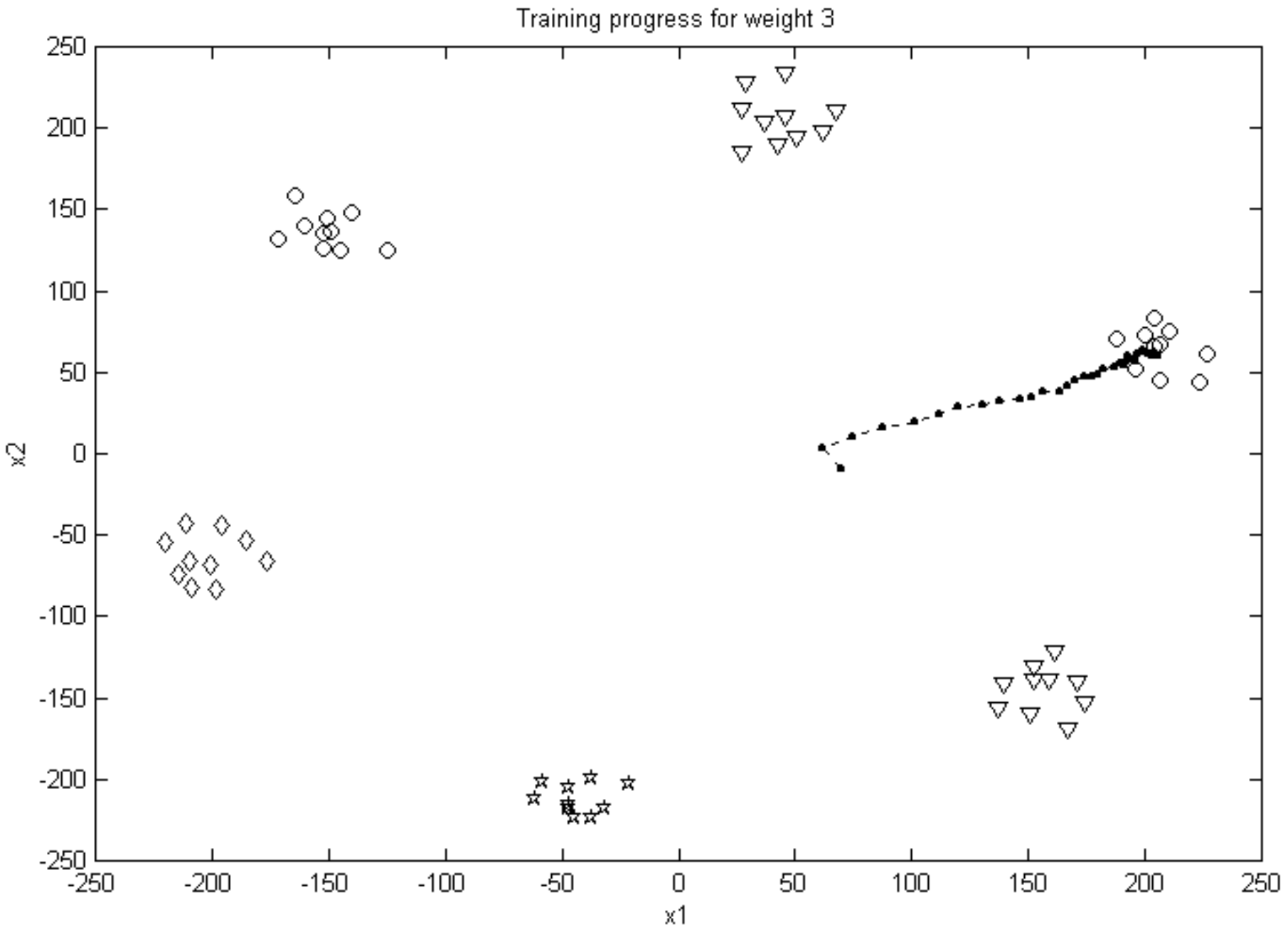

Figure 5.4.e. Training progress for third LVQ weight

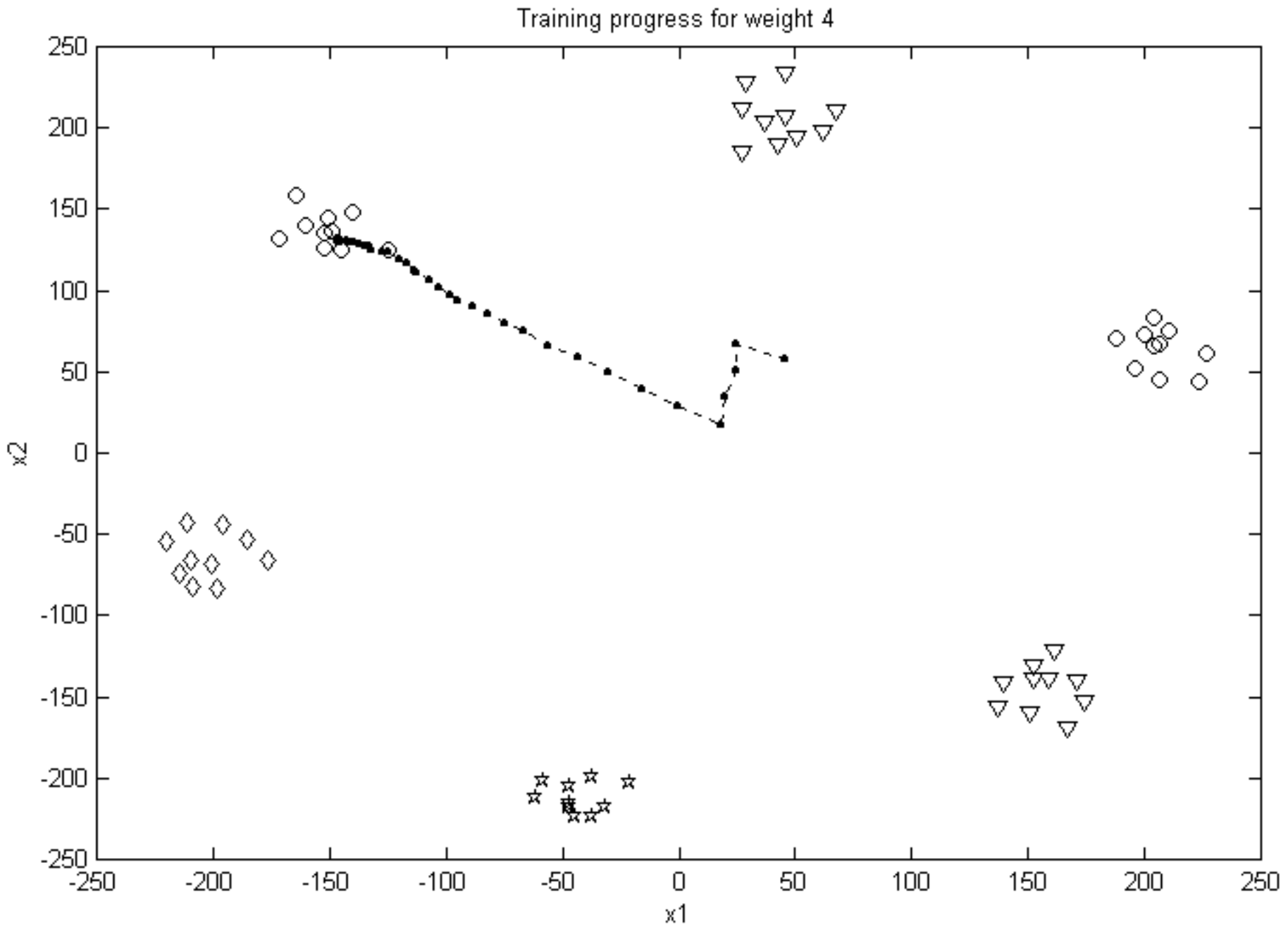

Figure 5.4.f. Training progress for fourth LVQ weight

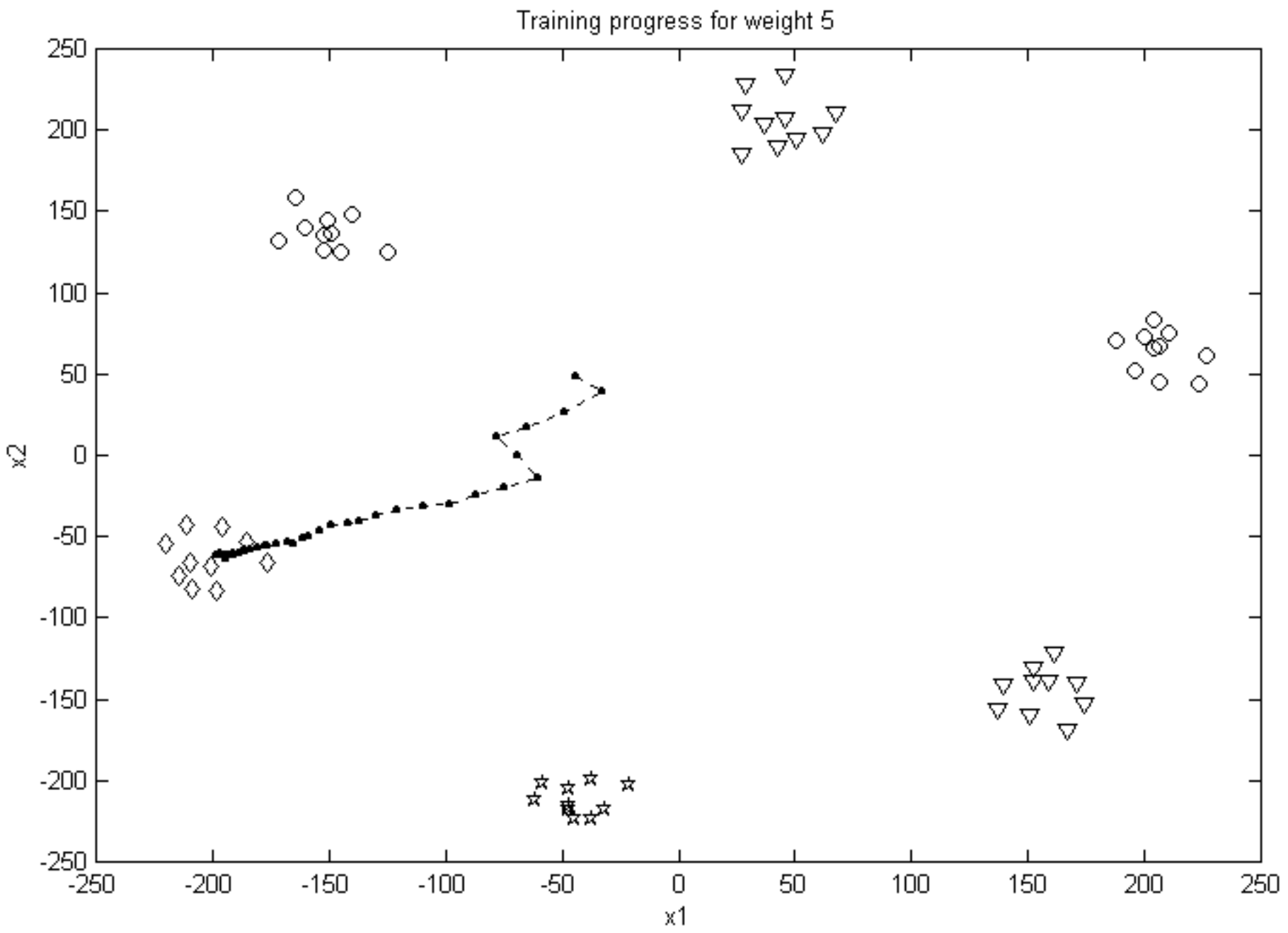

Figure 5.4.g. Training progress for fifth LVQ weight

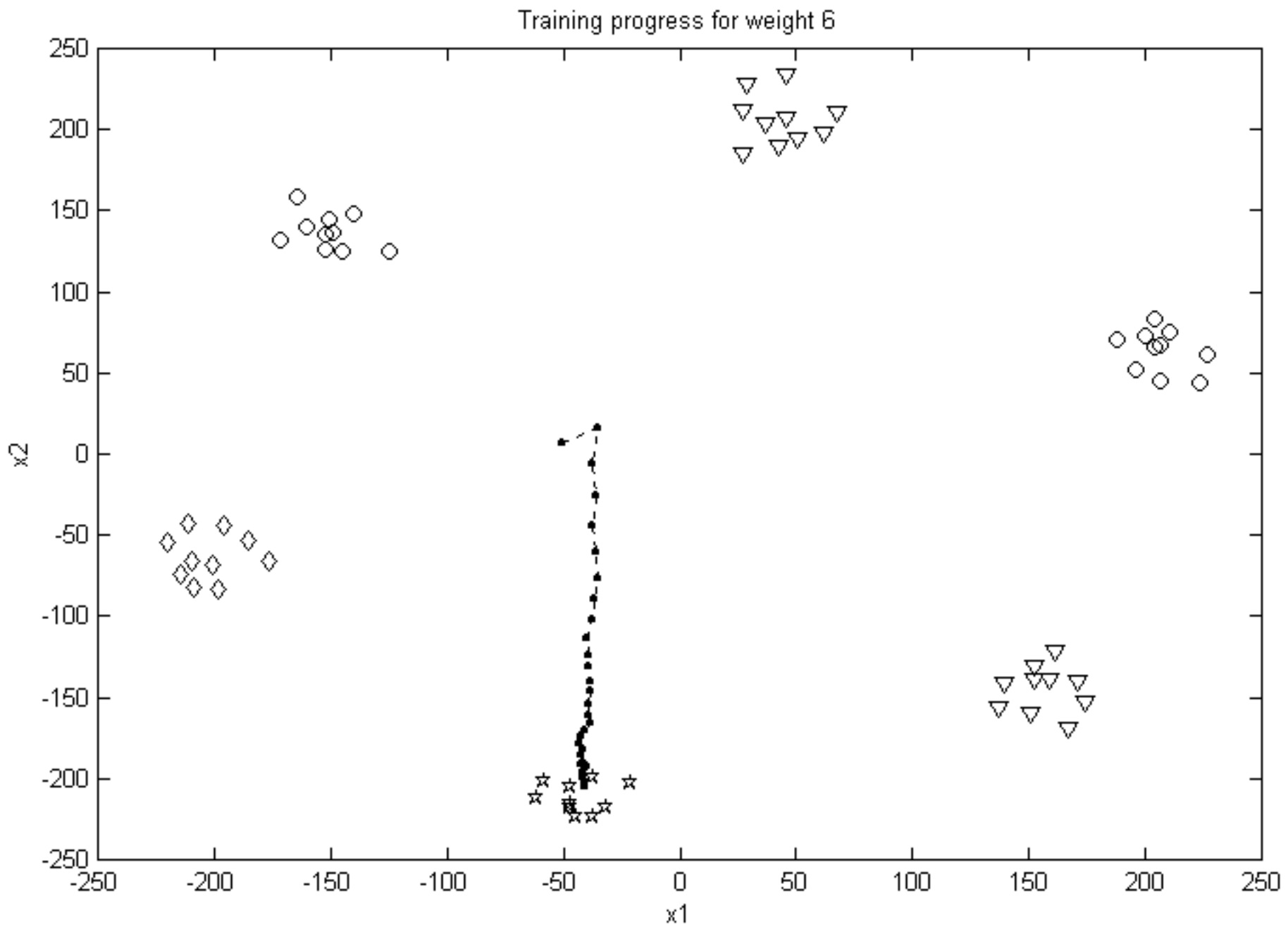

Figure 5.4.h. Training progress for sixth LVQ weight

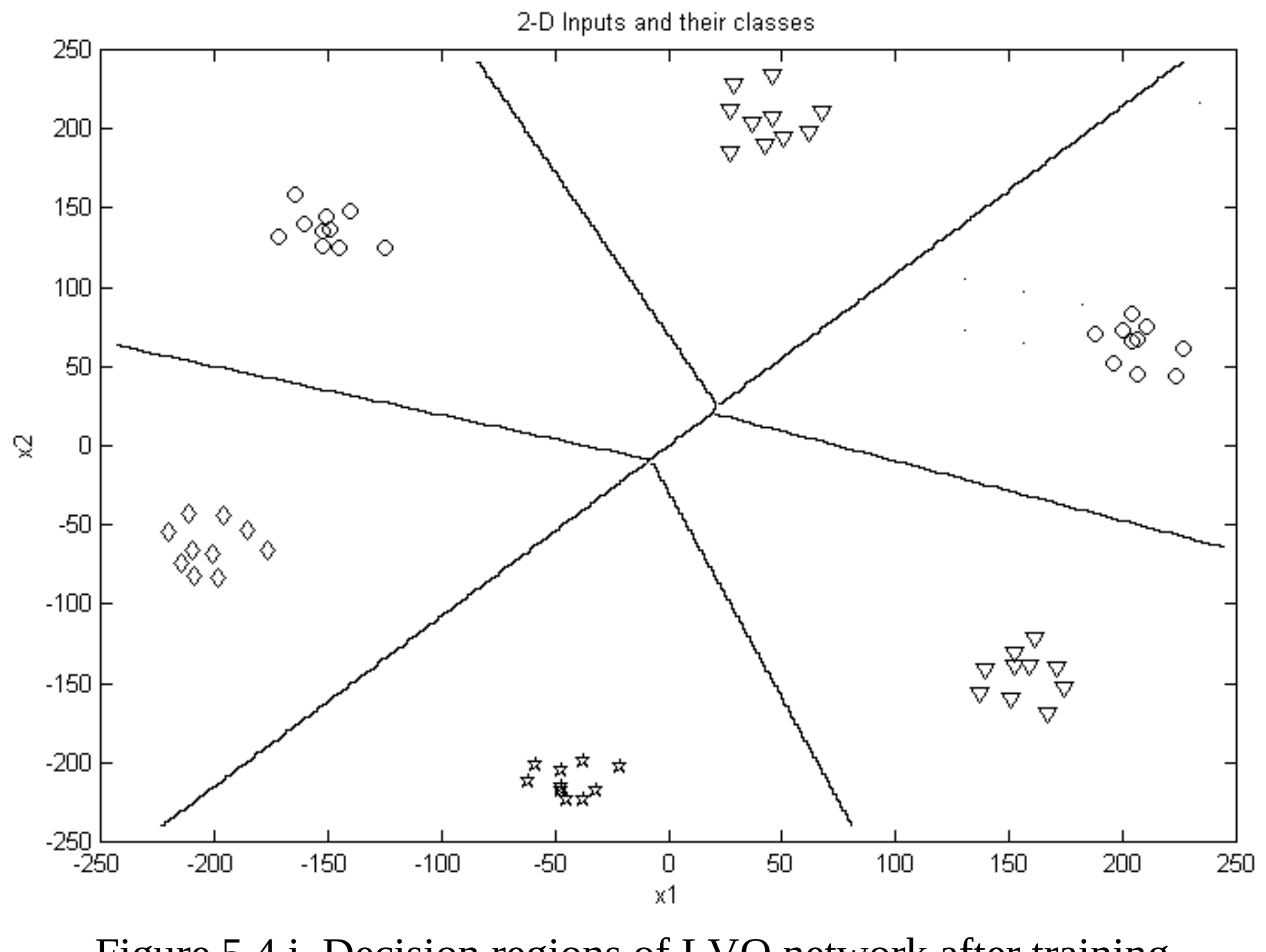

Figure 5.4.i. Decision regions of LVQ network after training

To illustrate the effect of training on the weights of the LVQ when used as a face recognizer, figure 5.5 shows the progress of one weight vector during 19 epochs of training. Since the inputs to the LVQ networks in this system are actually images of faces reshaped on the form of vectors, and since the LVQ weights encode the locations of input clusters then the weights can be displayed as images after reshaping them back into suitable dimensions.

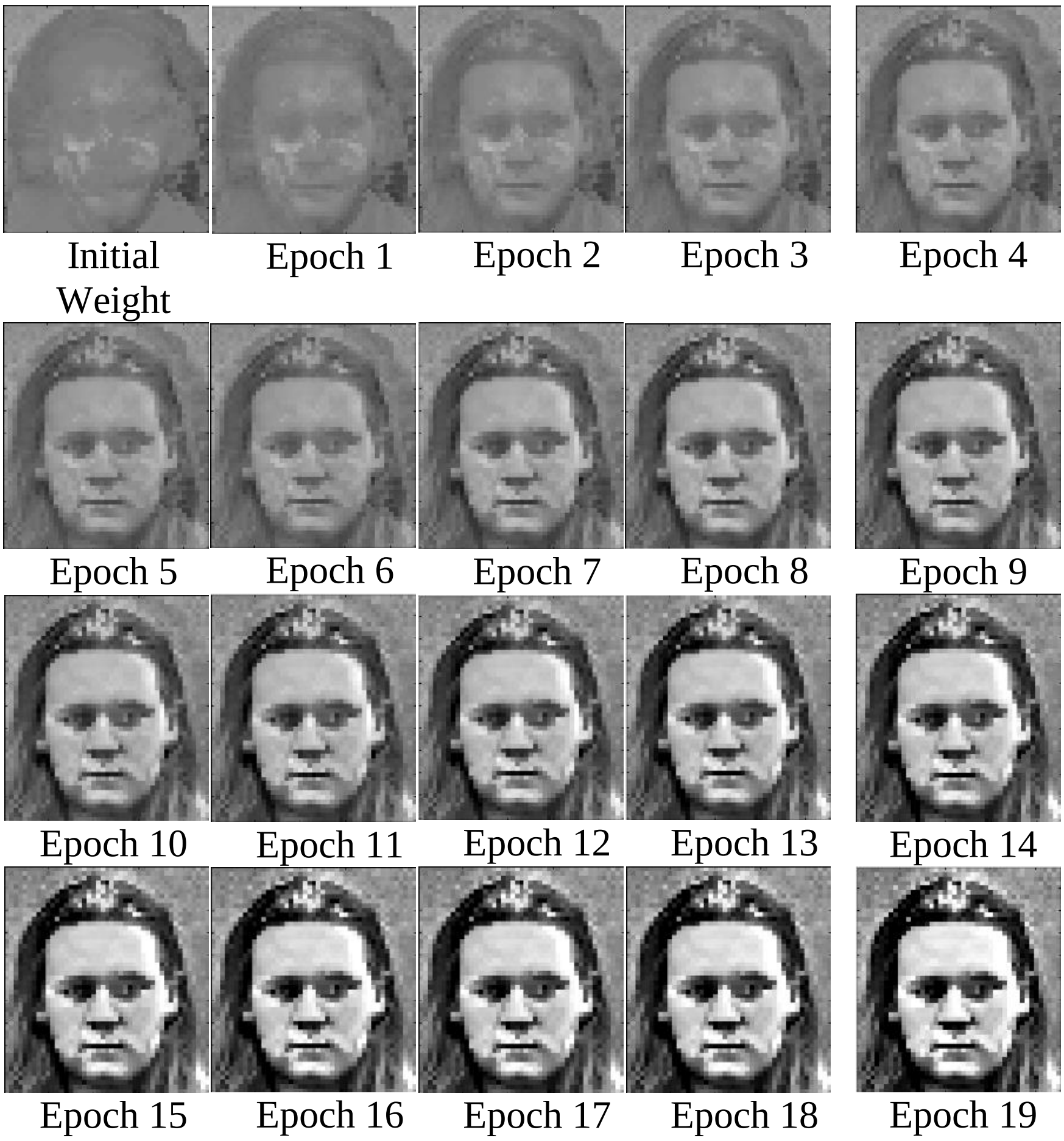

Figure 5.5. Training progress of an LVQ weight vector trained on input face images

Since combining classifiers increase the overall classification rate, LVQ classifiers can be used as the component classifiers of the committee while combining their decisions using plurality voting. This LVQ based structure shall be called a {System A} committee. The LVQ classifiers are trained such that a classifier joins the committee only if it correctly classifies more than (n) patterns. If the average recognition rate of the classifiers E = r' / (n M) is too small (where r' is the average number of patterns correctly classified per classifier) the increase in classification rate provided by the combination of the classifiers may not be sufficient to reach an acceptable overall classification rate. How to solve this problem is the next step that will be studied in the following section.

### *5.3 Enhancing System Performance*

In order for the system to perform as a reliable face recognition system, it must have a very high classification rate. This is necessary even if the number of classes (persons) is large. If a single very large LVQ classifier is used, it will be very hard to train it to classify accurately. On the other hand if several small LVQ classifiers are combined using the {System A} committee, their overall recognition rate will be increased but not enough to reach the desired high accuracy (by large and small it is meant the number of weights of the LVQ network). Another important characteristic of the system is the time and storage-space required to train and use the system. This section describes how to make LVQ classifiers' combination more accurate and less space and time demanding using two main methods: a) classifier pruning and b) using a controlled LVQ front-end classifier.

#### *5.3.1 Classifier Pruning*

As shown earlier, the number of weights in an LVQ network must be sufficient to encode the training input vectors clusters distribution in input space. In this system, the number of weights is selected randomly and thus might be more than enough in many cases. This means that for a certain LVQ classifier there might be some weights that are not actively used in classification because they did not go far from their initial positions during training. These weights are thus useless and should be eliminated to reduce both classification time and storage requirements. Given an LVQ classifier with a training set consisting of (k) gray scale images each having (p x q) points. The first step to prune this classifier is to obtain the standard deviation of the intensities for each image. This is done by assuming that the intensity at each point is a value for a random variable. The standard deviation of that random variable needs to be estimated from (p x q) samples. An image containing a human face should have relatively large variations in its intensities. Next, the smallest standard deviation among all images is calculated and divided by 2 as a threshold value to be used later. For the classifier weights it has been seen that each weight encodes a cluster of training inputs and hence the active weights will have the general intensity distribution as the training images. Any weight with a standard deviation less than the calculated threshold is discarded. This reduces the number of weights in many classifiers by up to 70% or more (as will be seen in the next chapter) hence reducing classification time and storage requirements effectively. To illustrate the idea, figure 5.6.a represents an actual training image with an illumination standard deviation of 74.1195. Figure 5.6.b is a weight trained on the class of that training image. Its standard deviation is 94.732. On the other hand, figure 5.6.c is a weight not trained sufficiently on any class and its standard deviation is very low. This means that this weight can be removed from the network to speed up classification and reduce the required storage. The main ideas behind the {System A} committee and LVQ pruning are due to the fruitful discussions with Prof. Dr. Ahmad S. Tolba who applied similar techniques to produce a solution for the problem of gender classification based on facial images.

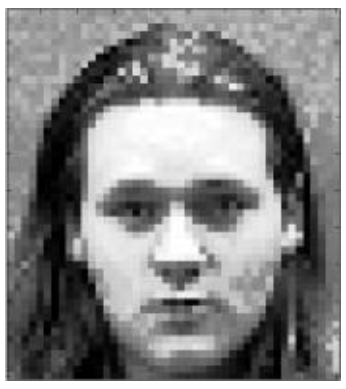

a
Standard Dev. =
74.1195

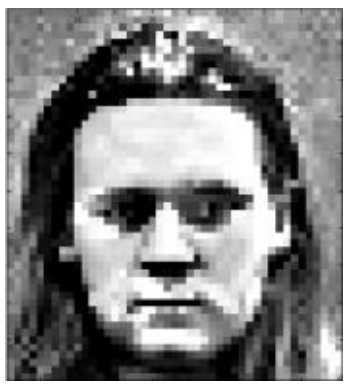

b
Standard Dev. =
94.732

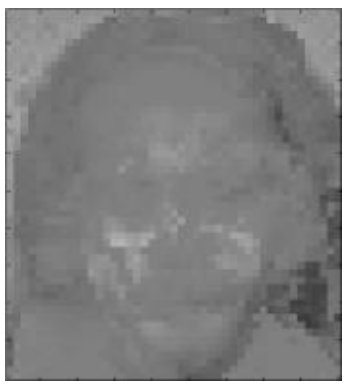

c
Standard Dev. =
14.7459

Figure 5.6. Standard deviation values for a face image (a), a trained LVQ weight (b), and an untrained LVQ weight (c)

***5.3.2 Controlled LVQ (CLVQ)***

A modification to the basic training and classification algorithms of the LVQ network proved to be very effective in enhancing the performance of the system. A controlled LVQ network is trained like an ordinary LVQ network with one difference: in a normal LVQ network, each training input is compared to all the weights of the network to select the winning weight. However, in the CLVQ each input is compared to some of the weights. Selection of the weights that are to be compared with a certain training input 'x' is accomplished through another set of controlling inputs. These controlling inputs represent prior knowledge about that training input 'x'. More specifically they represent the classes to which 'x' is very near. This includes the actual class of 'x' and some classes that another prior classifier might be 'confused' about to which of them ‘x’ actually belongs. This means that CLVQ can act as a second stage after a classification system produces its results to enhance the results. This is accomplished by training CLVQ on classes about which the prior classification system is confused. If the training input 'x' is very clear to the prior classification system (its actual class is predicted accurately with a very high vote compared to other classes), the CLVQ is not trained on 'x'. After training the CLVQ, pruning can be applied to eliminate the CLVQ weights not involved in its training. Hence the CLVQ classifier acts as a Front End Classifier (FEC) to enhance the recognition rate.

This method requires more weights to be trained on the training inputs. This second training stage can, however, be completely eliminated. All of the base classifiers are LVQ and so for the FEC. Hence, some of (or even all of) the weights of the base LVQ classifiers (already trained on the inputs) can be used as the weights for the FEC LVQ network. By doing so, training another very large classifier is avoided. In addition, the storage space required to store the weights of the FEC is reduced to nothing. Two choices are present during classifying an input: either the input is compared with all the CLVQ weights or the input is compared with some of them depending on the results obtained from a previous classifier (which is a {System A} committee). The CLVQ FEC replaces plurality voting decision combination block in the system shown in figure 5.2. This system shall be called a {System B} committee.

To make this FEC technique clearer, what shall be called a 'confusion set' ($S_C$) is defined. The confusion set ($S_C$) for a class (C) is the set of all classes highly voted for by the committee when the training inputs belonging to class (C) are presented to the committee. How high a vote is will be defined as follows: Assuming after training the committee that the average recognition rate per classifier is (t) and that there are (R) classifiers trained on any class (C). Any pattern belonging to (C) will probably have more than (t R) correct votes. On that basis any class having more than, say, (0.75 t R) votes is added to the confusion class ($S_C$). For example having a committee consisting of 64 classifiers each trained on (M = 16) classes with (R = 8) classifiers trained on any single class and (t = 66%). Assuming a training input belonging to class 2 gets the following votes: seven votes for class 0, six votes for class 1, eight votes for classes 2 and 11, five votes for classes 4, 6, 8, 13, 14, and 15 and one vote for classes 3, 5, 7, 9, and 12. Then classes 0, 1, 2, 11, 4, 6, 8, 13, 14, and 15 are added to the confusion set ($S_2$) because all of them got votes > 3.96 (since 0.75 t R = 3.96). The confusion set ($S_C$) must at least include the class (C) itself. If $S_C = \{C\}$ then no confusion is present between (C) and any other class from the committee's point of view. Hence, the CLVQ is not to be trained on inputs belonging to class (C). Another set, called a 'suspicion set' ($T_C$) for a class (C), is derived from the confusion sets. The set ($T_C$) is the set of all classes such that (C) belongs to the confusion sets of these classes. $T_C$ describes the probable classes that the input might actually belong to if the predicted class is (C). It is clear that the suspicion sets for all classes can be deduced having the confusion sets for all classes and vise versa. It is also clear that the suspicion set must have at least one class. The confusion sets are used primarily in the system to deduce the suspicion sets. The suspicion sets can be used during classification to select the weights to be compared with the input vector that needs to be classified. This is done by supplying the CLVQ FEC with the union of all sets $T_C$ as a controlling input where C is any class that got a high vote by the committee.

### *5.4 The Proposed Algorithm*

Instead of presenting a through mathematical formulation of the complete algorithm, a general explanation for the algorithm is given. The reason for choosing that style is that the algorithm is full of details that might be confusing if put in a rigid mathematical formulation and may affect the insight into the ideas behind the algorithm. The system operates in two phases: training, normally done once, and classification. The general steps are thus as follows:

*Training phase:*

Part 1: Construct the {System A} committee

Step 1.1: Construct (N) bootstraps from the training data.

Step 1.2: For each bootstrap train (L) LVQ classifiers on the bootstrap.

Step 1.3: For each trained LVQ classifier prune the classifier to eliminate unwanted weights.

Part 2: Complete the {System B} committee by constructing the FEC

Step 2.1: Construct the FEC LVQ classifier by regrouping the {System A} weights already trained in Part 1

Step 2.2: Construct the confusion sets for all classes based on the votes of the {System A} committee

Step 2.3: Construct the suspicion sets for all classes based on the confusion sets of step 2.2

*Classification phase:*

Part 3: Introduce the input pattern to the component classifiers

Step 3.1: For each classifier, introduce the input pattern and obtain its classification

Step 3.2: Obtain the predicted classes and their corresponding votes from the outputs of step 3.1

Step 3.3: Construct a set of classes that the {System A} committee gave high votes for

Part 4: Obtain the final classification using the FEC LVQ

Step 4.1: Construct the union of suspicion sets based on the suspicion sets calculated in step 2.3 and the set of classes in step 3.3

Step 4.2: From the regrouped weights of step 2.1, select the weights corresponding to the classes belonging to the union of the suspicion sets of step 4.1

Step 4.3: Introduce the input pattern to the FEC LVQ and obtain its final classification as the final decision

### *5.4.1 Explaining Part 1*

The purpose of the training phase is to construct a system capable of collecting enough information to be used during classification. The training phase starts with a large set of static images belonging to many classes. In step 1.1, (N) bootstraps are constructed from (M) classes. Each bootstrap contains the training images from (m) different classes. This is done using the technique described in section 5.1.1. Each bootstrap is considered as an independent training set. Next, (L) LVQ classifiers are trained on that training set. Each one of the (L) classifiers is trained using the LVQ training algorithm given in section 5.2. Three parameters are varied randomly among the (L) classifiers: Initial number of weights, maximum number of training epochs and the learning rate. After training all the classifiers a total of (K) LVQ classifiers (K = L N) are obtained. Most of these classifiers contain weights that have not moved much from their initial position. In step 1.3, each classifier is pruned as described in section 5.3.1 to reduce the time and space requirements of the system during the following steps. After pruning, The LVQ classifiers are tested for two conditions. Assuming an LVQ is trained on (m) classes and (n) images per class then there must exist at least one weight after pruning that corresponds to each class of the (m) classes. The other condition is that the classifier must at least recognize more than (n) patterns correctly to be better than random guessing as described in section 5.1.1. If the LVQ does not fulfill these two conditions, it is discarded and another LVQ is trained instead. The purpose of such conditions is to ensure fairness of training for all classes with acceptable accuracy. At the end of part 1, a set of (K) LVQ classifiers is obtained. The classifiers of this set are trained on all of the (M) classes as a whole with a reduced set of weights that encode the distribution of training inputs in the input space. This is the {System A} committee that will be completed in part 2 to become a {System B} committee.

### *5.4.2 Explaining Part 2*

In part 2, the system completes the final stage in information gathering to be ready for classification. The first step is to regroup the weights of the already trained LVQ classifiers into one large FEC LVQ network. Each class of the (M) classes has (R) LVQ classifiers trained on its images. Hence, the FEC will contain at least (R) weights per class. This is because each LVQ classifier contains at least one weight to represent each of the classes on which it was trained. In step 2.2, a confusion set is constructed for each class as described in section 5.3.2. As an example to illustrate the following steps, assuming 8 classes are given {0, 1, 2, 3, 4, 5, 6, 7} and assuming their confusion sets are as follows:

$S_0$ = {0}

$S_1$ = {0, 1, 4}

$S_2$ = {2, 6}

$S_3$ = {3, 4, 7}

$S_4$ = {0, 1, 4, 6}

$S_5$ = {5}

$S_6$ = {4, 6}

$S_7$ = {2, 3, 7}

It is noted that the set $S_C$ must contain the class C among its members. Step 2.3 extracts the suspicion sets for all the classes. From the above example the suspicion sets are:

$T_0$ = {0, 1, 4} (i.e. class 0 appears in $S_0$, $S_1$, and $S_4$)

$T_1$ = {1, 4}

$T_2 = \{2, 7\}$

$T_3 = \{3, 7\}$

$T_4 = \{1, 3, 4, 6\}$

$T_5 = \{5\}$

$T_6 = \{2, 4, 6\}$

$T_7 = \{3, 7\}$

It is clear that the set $T_C$ must contain the class C among its members. The purpose of the sets $T_C$ will be explained in the following section.

### *5.4.3 Explaining Parts 3 and 4*

In part 3, the system is ready to receive one input image to be classified. In the first step, the input is presented to all of the LVQ component classifiers to obtain (K) classifications (votes). These votes are used to construct a set of class labels in a similar way to the construction of the confusion set as describes in section 5.3.2. For example, assuming that this set is S = {0, 1, 5} for a certain input, the opinion of the {System A} committee about this input is that it belongs to one of the classes in S as calculated in step 3.3. This opinion may or may not be correct. From the information gathered during training, a pattern is sometimes classified as belonging to class 0 when it actually belongs to one of the classes {0, 1, 4} as can be seen from the set $T_0$ calculated above. Also a pattern is sometimes classified as belonging to class 1 when it actually belongs to one of the classes {1, 4} as can be seen from $T_1$. Finally, a pattern is classified as belonging to class 5 only when it actually belongs to class 5 as can be seen from $T_5$. The possible classes that the input pattern belongs to can be obtained by uniting these three suspicion sets $T_0$, $T_1$, and $T_5$ into the set T = {0, 1, 4, 5}. The set (T) is constructed in step 4.1. The weights that encode the classes belonging to (T) are used as an LVQ classifier and the input is finally presented to that constructed classifier to obtain the final classification of the system.

### *5.5 Relation to Other Combined Systems*

In chapter 4, generative and non-generative algorithms for constructing committee machines are presented. The proposed system presented in this chapter contains features found in both generative and non-generative algorithms for constructing committee machines. The focus of generative algorithms is on the production of diverse base classifiers. This is seen in the bootstrapping technique in part 1 of the algorithm. On the other hand, non-generative algorithms focus on using a powerful combination method to produce an enhanced classification decision from the base classifiers. This can be found in the design of the FEC classifier in part 2 of the algorithm.

#### *5.5.1 Relation to Generative Algorithms*

The base classifiers are trained using a technique similar to resampling methods presented in section 4.2.1.1. It differs from bagging in the random nature of selection used in bagging that may result in unfair training chance for some classes or inputs. It differs from boosting in the independence of each classifier from other previously trained classifiers. This is in direct contrast with boosting in which the training of each new classifier depends on the errors made by previously trained classifiers. The other difference from typical resampling methods is in the decision combination technique that is very simple in typical resampling methods (voting for example). The system avoids using feature extraction completely. In a typical face recognition system, a feature extractor is a mapping from the input space to the feature space that associates each input image with a feature vector. The classifier is then a mapping from the feature space to the label space that associates each feature vector with a class label. If the high performance feature extractor is used, the job of the classifier is made easy and a very simple classification technique may be applied. If a low performance feature extractor is used, a powerful classifier is required to accomplish the task. In this system, the algorithm uses a very simple feature extractor: the identity mapping. Hence, the whole classification task is performed by the classifier. This simplifies the design process of the system and reduces the number of training parameters to be selected. The system also bears some resemblance to the mixture of experts methods (section 4.2.1.3). Instead of a gating network that selects the suitable classifiers from the committee, the FEC selects the suitable LVQ weights based on the information obtained from the base classifiers. A similarity with test and select methods is also present. For a new LVQ classifier to be added to the committee two conditions must hold true: The recognition rate of the LVQ must be high enough and it must contain at least one trained weight for each one of the classes it was trained on. As in test and select methods not any classifier is added but only the ones fulfilling certain conditions. Finally the random selection of LVQ training parameters (like number of hidden units, learning rate, and training epochs) is similar to the randomized ensemble methods.

### *5.5.2 Relation to Non-Generative Algorithms*

Although the many similarities between this system and many generative approaches, the system has little in common with the non-generative approaches presented in section 4.2.2. The nearest non-generative approach to this system is the stacked approach where the outputs of the base classifiers are inputs to a next stage classifier. The proposed system takes one step ahead by using such outputs together with the input image as inputs to a second stage classifier (the FEC) in the way described in parts 2, 3 and 4 of the {System B} algorithm. This results in many desirable effects like high stability and high accuracy for the combined system as will be seen in the next chapter. Nevertheless, The (System B) committee focuses on enhancing the combination strategy as in all non-generative algorithms.

From the above discussion of similarities, it appears that the {System B} committee can be considered a general scheme for combining classifiers that possesses many desirable features from other widely used combination algorithms.

### *5.6 Conclusions*

In this chapter, a combined system that is capable of performing face recognition is presented. This system require very few parameters to be selected. Feature extraction is completely avoided in this system. There are many similarities between this system and many generative approaches for constructing committee machines. The combination method used in this system is different from the commonly used methods in combined classifiers systems. The system can be implemented using parallel processing techniques. This is because both training and testing require little interaction between the  base classifiers.

*Chapter 6*

# Experiments and Results

In this chapter, the performance and characteristics of the suggested face recognition system is presented through a set of experiments. Section 6.1 describes the preparations made for the experiments. Section 6.2 describes the experiments conducted on the face databases using the proposed system. Section 6.3 is a discussion of the results and finally the conclusions are presented in section 6.4.

## *6.1 Preparations*

Before presenting the experiments, the software packages used to implement the proposed algorithm are presented in section 6.1.1. Following that, a description of the used face databases is given in section 6.1.2. Next, a description of preprocessing performed on the individual images of the databases is provided in section 6.1.3.

### *6.1.1 Software Packages used for Implementation*

The proposed algorithm is implemented using Matlab 5.3 and VisualBasic 6.0. Matlab is used for performing the actual calculations while VisualBasic is used for interfacing with the program user. The data exchange between VisualBasic and Matlab is through the ActiveX capabilities of both software packages. The neural network toolbox in Matlab is used to implement the base LVQ classifiers. The FEC algorithm is manually programmed using Matlab. The image processing toolbox of Matlab is used to implement the image preprocessing steps and the image file read operations.

### *6.1.2 Face Database Description*

The Essex face database is the main database used in the experiments [56]. It consists of static images of 395 individuals with 20 images per individual. It contains images of male and female subjects of various racial origins. The images are mainly of first year undergraduate students, so the majority of individuals are between 18-20 years old but some older individuals are also present. The image format is 24-bit color JPEG taken using an S-VHS camcorder under an artificial mixture of tungsten and fluorescent overhead lighting. The database is divided into 4 sets of images. The four sets contain 20 images per individual taken by a fixed camera as a single sequence. The first set is called 'faces94' where the subjects sit at fixed distance from the camera and are asked to speak, whilst a sequence of images is taken. The speech is used to introduce facial expression variation. This set contains 180 by 200 pixels images of 153 individuals with a plain green background. The head turn, tilt and slant are almost constant with minor changes in the position of the face in the images. No lighting variations are present. Considerable expression changes are present with no individual hairstyle variation as the images were taken in a single session. The second set is called 'faces95'. During the sequence, the subject takes one step forward towards the camera. This movement is used to introduce significant head (scale) variations between images of the same individual. There is about 0.5 seconds between successive frames in the sequence. The set contains 180 by 200 images of 72 individuals. The background consists of a red curtain. Background variation is caused by shadows as subject moves forward. The set contains large head scale variation and minor head turn, tilt and slant variations with some translation in the position of the face in the images and some expression variation. As a subject moves forward, significant lighting changes occur on faces due to the artificial lighting arrangement. The third set is called 'faces96'. During the sequence, the subject takes one step forward towards the camera. This movement is used to introduce significant head variations between images of the same individual. There is about 0.5 seconds between successive frames in the sequence. 196 by 196 pixels images of 152 individuals were taken. The background is complex (glossy posters) with a large variation in head scale. The images contain minor variation in head turn, tilt and slant with some translation in the position of the face in the images and some facial expression variation. As a subject moves forward, significant lighting changes occur due to the artificial lighting arrangement. The last set is called 'grimace'. During the sequence, the subject moves his/her head and makes grimaces, which get more extreme towards the end of the sequence. Otherwise, the setup is similar to 'faces95'. There is about 0.5 seconds between successive frames in the sequence. 180 by 200 pixels images of 18 individuals are taken. The background is plain with small head scale variation and considerable variation in head turn, tilt and slant with some translation in the position of the face in the images. Very little image lighting variation is present with major expression Variation. There is no hairstyle variation as the images were taken in a single session.

Another very important database is the ORL database [39], currently maintained by AT&T Laboratories Cambridge. This database contains a set of faces taken between April 1992 and April 1994 at the Olivetti Research Laboratory in Cambridge, UK. There are 40 distinct subjects with 10 different images for each. The size of each image is 92 by 112, 8-bit gray levels. For some of the subjects, the images were taken at different times, varying lighting slightly, facial expressions (open/closed eyes, smiling/non-smiling) and facial details (glasses/no-glasses). All the images are taken against a dark homogeneous background and the subjects are in up right, frontal position (with tolerance for some side movement).

### *6.1.3 Data Preparations and Preprocessing*

Each person (class) has (n) images in the database. For each class the (n) images are divided into (n / 2) images for training and (n / 2) images for testing. Before an image is presented to the system, it is preprocessed as follows:

1- A colored image is converted to grayscale by averaging its RGB color components.

2- Histogram equalization is applied to the image to reduce lighting variation effects.

3- The image is down-sampled to a suitable size. A size around 50 by 50 pixels is suitable to reduce input vector dimensions while preserving most of the image details.

4- The image rows are concatenated to produce a single input vector (of more than 2000 dimensions in most cases).

Figure 6.1 shows the training and testing images of the first class of each of the four sets of the Essex database. Figure 6.2 shows the preprocessing applied to a single image from the sequence. Figure 6.3 shows a sample class from the ORL database.

## *6.2 Experiments and Results*

In this section, the experiments conducted on the proposed system are described along with their results. The experiments were not meant to be a through experimental study of the system. They were just meant to be indicators for the features, disadvantages and possible applications of the system. A through experimental study would require hundreds or even thousands of experiments on many face databases. This is not a possibility because of the lack of time and data required for covering such a study. The first set of experiments in section 6.2.1 aims at highlighting the stability of the system. The second set in section 6.2.2 illustrates the usefulness of using a {System B} committee compared to a {System A} committee. The third set in section 6.2.4, consisting of a single experiment, illustrates one important disadvantage of the system. The final experiment in section 6.2.5 illustrates the {System B} capabilities in classification of a large dataset consisting of 392 classes. All recognition rates in the following sections are calculated for the testing patterns (not the training patterns) unless stated otherwise.

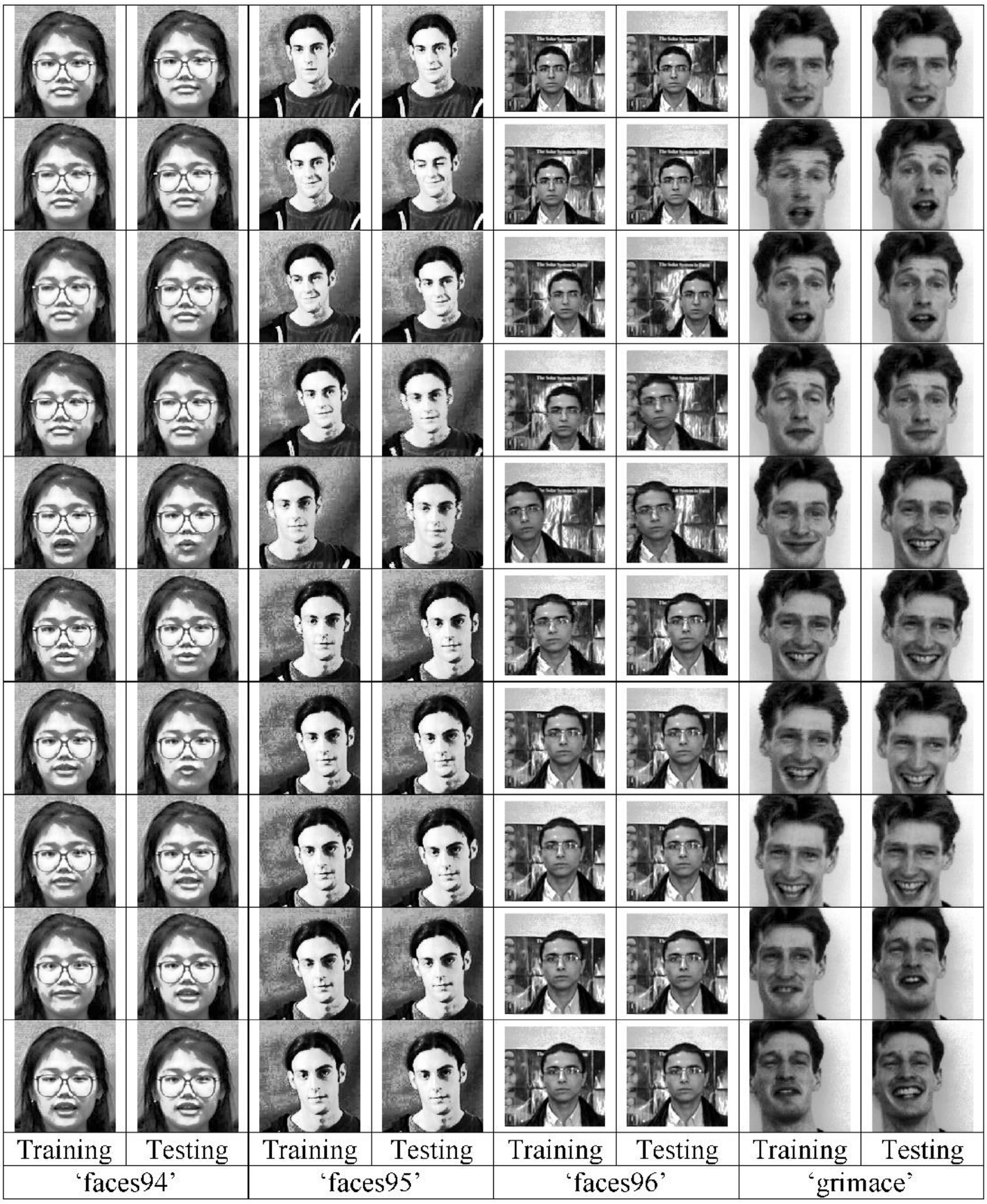

Figure 6.1. Preprocessed image samples from the four sets of the Essex face database

| | |
|---|---|
| 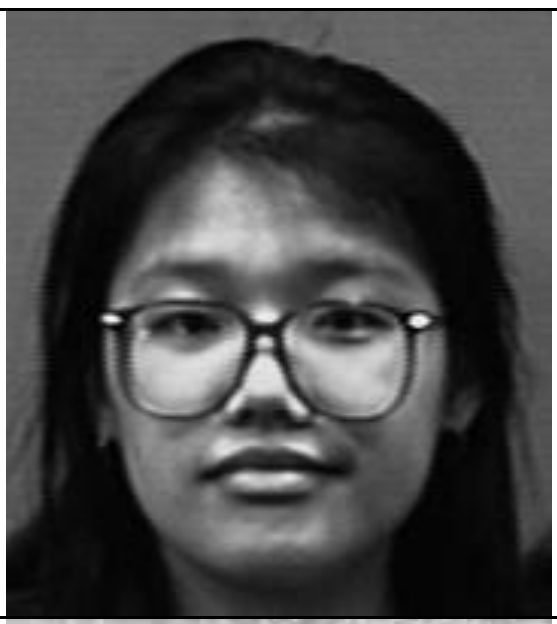 | Gray-scale image |
| 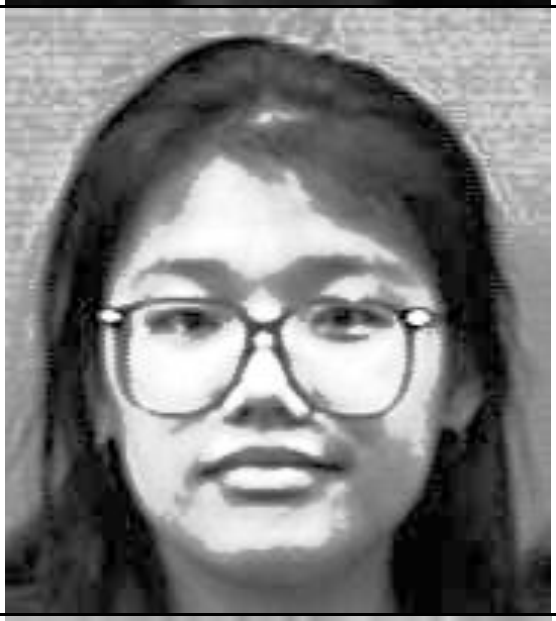 | Histogram equalization |
| 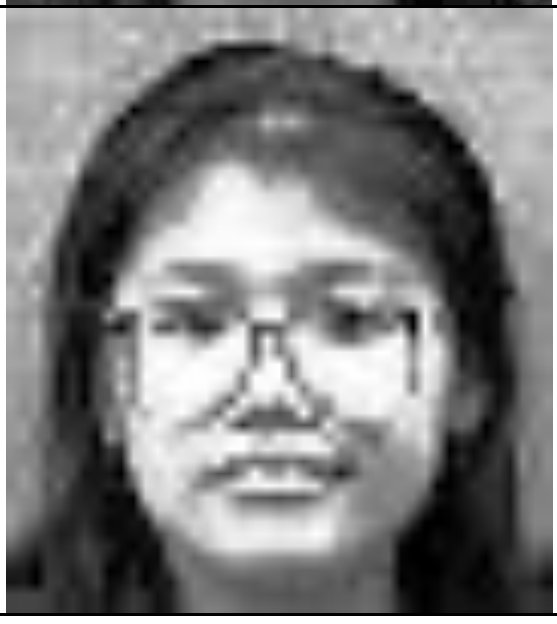 | Image size reduction via down-sampling |

Figure 6.2. Preprocessing Stages

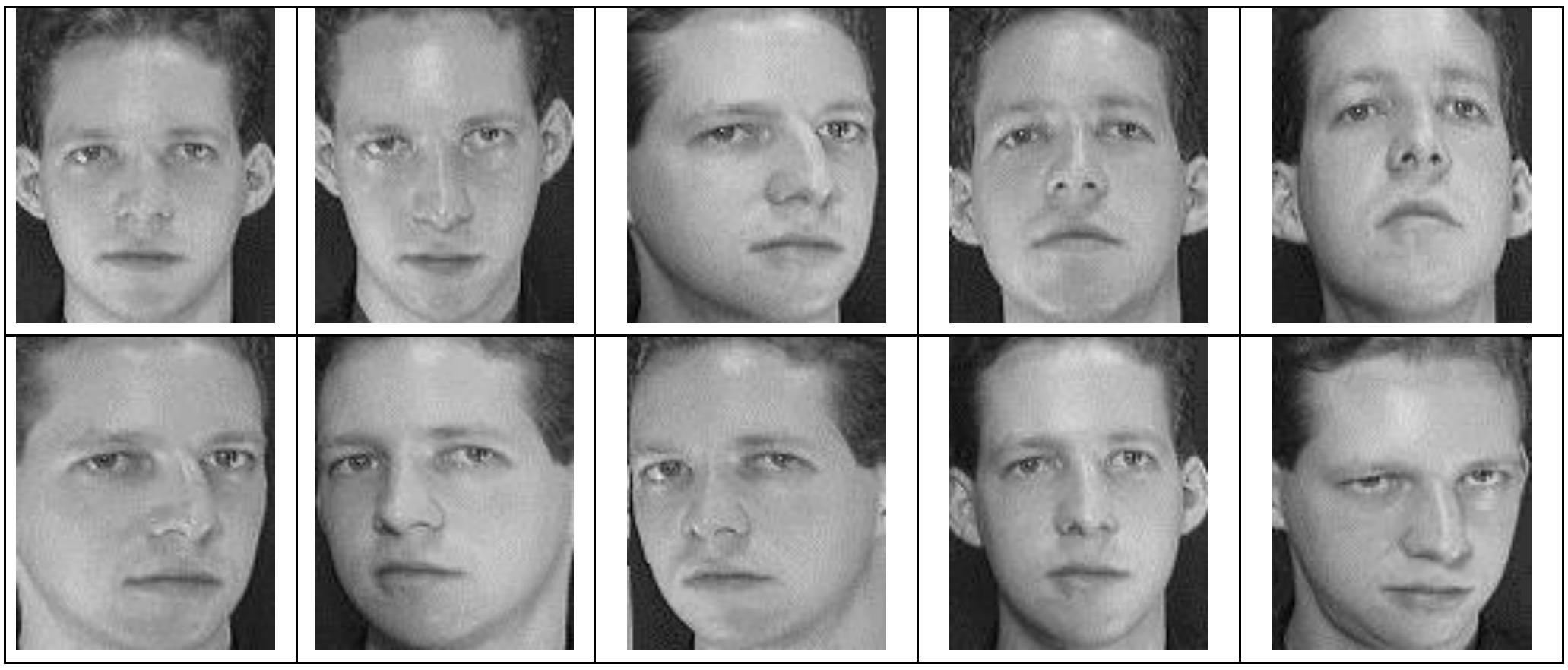

Figure 6.3. Image samples from the ORL face database

### *6.2.1 Stability*

Stability of a classifier is very important for practical applications. A classifier is said to be stable if its recognition performance on some dataset is not too sensitive to its initial conditions and parameters. The performance of a stable classifier should not be too sensitive to parameters such as number of training iterations, the order of presenting training inputs, exact details of its underlying structure (for example, like number of hidden units or initial values of weights for LVQ networks). In this section, it is shown that a {System B} committee is very stable compared to other face classification systems. Three configurations are tested on the 'faces94' database. Each configuration is tried 10 times with both the {System A} and {System B} committees. The recognition rates are shown in tables 6.1, 6.2 and 6.3. The tables also show the parameters for each configuration. The results in tables 6.1, 6.2 and 6.3 indicate that the {System B} committee is very stable compared to the {System A} committee. Table 6.4 shows the recognition rates of 12 LVQ classifiers each trained on the same 30 classes used in table 6.3 from the 'faces94' set. The LVQ parameters (number of training epochs, number of hidden units, and learning rate) were randomly selected. It is clear that the stabilities of LVQ classifiers and {System A} committee are very low compared to the stability of {System B} committee.

Table 6.1. Recognition rates of {System A} and {System B} committees (first configuration)

| Trial | Configuration 1<br>M = 24, m = 18,<br>V = 15, L = 1 | |
|---|---|---|
| | {System A} | {System B} |
| 1 | 91.250 | 99.167 |
| 2 | 96.667 | 99.167 |
| 3 | 96.667 | 99.583 |
| 4 | 97.917 | 99.583 |
| 5 | 95.000 | 100.000 |
| 6 | 95.417 | 99.583 |
| 7 | 95.833 | 99.583 |
| 8 | 97.500 | 99.583 |
| 9 | 93.750 | 99.583 |
| 10 | 94.167 | 99.583 |
| Mean | 95.417 | 99.542 |
| Standard Deviation | 1.993 | 0.236 |

Table 6.2. Recognition rates of {System A} and {System B} committees (second configuration)

| Trial | Configuration 2<br>M = 48, m = 12,<br>V = 0, L = 2 | |
|---|---|---|
| | {System A} | {System B} |
| 1 | 10.833 | 99.375 |
| 2 | 15.208 | 98.750 |
| 3 | 9.792 | 99.375 |
| 4 | 12.917 | 99.792 |
| 5 | 11.875 | 98.542 |
| 6 | 10.208 | 99.792 |
| 7 | 8.750 | 99.167 |
| 8 | 6.875 | 98.542 |
| 9 | 9.583 | 98.125 |
| 10 | 6.458 | 99.583 |
| Mean | 10.250 | 99.104 |
| Standard Deviation | 2.653 | 0.581 |

Table 6.3. Recognition rates of {System A} and {System B} committees (third configuration)

| Trial | Configuration 3<br>M = 30, m = 5,<br>V = 0, L = 10 | |
|---|---|---|
| | {System A} | {System B} |
| 1 | 22.667 | 99.333 |
| 2 | 22.333 | 100.000 |
| 3 | 59.333 | 99.667 |
| 4 | 19.333 | 99.000 |
| 5 | 10.333 | 99.000 |
| 6 | 5.333 | 99.000 |
| 7 | 10.000 | 99.667 |
| 8 | 23.667 | 99.333 |
| 9 | 21.667 | 99.667 |
| 10 | 28.000 | 99.000 |
| Mean | 22.267 | 99.367 |
| Standard Deviation | 14.898 | 0.367 |

Table 6.4. Recognition rates of LVQ classifiers

| Trial | Recognition rate % |
|---|---|
| 1 | 94.667 |
| 2 | 78.333 |
| 3 | 79.667 |
| 4 | 86.333 |
| 5 | 95.333 |
| 6 | 52.667 |
| 7 | 90.000 |
| 8 | 87.000 |
| 9 | 97.667 |
| 10 | 99.000 |
| 11 | 84.333 |
| 12 | 78.667 |
| Mean | 85.306 |
| Standard Deviation | 12.596 |

### *6.2.2 The Main Experiments*

Table 6.5 shows the results of 43 different configurations for both {System A} and {System B} applied to (M = 60) classes from the 'faces94' set. The columns are labeled as follows:

- m: Number of classes per bootstrap
- V: Overlap parameter
- L: Number of layers (Number of LVQ classifiers trained on the same bootstrap)
- N: Number of bootstraps
- s: Shift parameter (s = m - V)
- R: Number of LVQ classifiers trained on any single class (R = L m / s)
- K: Total number of LVQ classifiers in the committee (K = L N)
- C01: Percentage of correctly classified test patterns (face images from test set) by {System A} committee ( = recognition rate of {System A})
- C02: Average recognition rate of component LVQ classifiers on their respective bootstraps (recognition rate on only (m) classes)
- C03: Average recognition rate of LVQ classifiers on the whole test-set (recognition rate on all (M) classes)
- C04: Total number of weights in {System A} committee before pruning
- C05: Total number of weights in {System A} committee after pruning
- C06: Average number of weight comparisons in {System B} committee
- C07: Recognition rate of {System B} committee

The table rows are ordered based on the values in column C06 in ascending order. Many interesting properties of the system can be observed in the results of table 6.5. These properties are discussed in section 6.3

Table 6.5. Forty-three experiments using 60 Essex classes on {System A} and {System B} committees using different configurations

| Configuration | m | V | L | N | s | R | K | C01 | C02 | C03 | C04 | C05 | C06 | C07 |
|---|---|---|---|---|---|---|---|---|---|---|---|---|---|---|
| 1 | 30 | 0 | 2 | 2 | 30 | 2 | 4 | 37.833 | 69.170 | 34.580 | 687 | 135 | 170 | 98.833 |
| 2 | 10 | 0 | 2 | 6 | 10 | 2 | 12 | 13.500 | 80.750 | 13.460 | 396 | 136 | 203 | 100.000 |
| 3 | 30 | 0 | 4 | 2 | 30 | 4 | 8 | 52.333 | 57.870 | 28.940 | 636 | 177 | 210 | 97.833 |
| 4 | 15 | 0 | 3 | 4 | 15 | 3 | 12 | 22.333 | 54.560 | 13.640 | 763 | 137 | 233 | 100.000 |
| 5 | 15 | 0 | 4 | 4 | 15 | 4 | 16 | 33.833 | 67.000 | 16.750 | 1119 | 222 | 320 | 95.500 |
| 6 | 20 | 0 | 4 | 3 | 20 | 4 | 12 | 56.333 | 87.580 | 29.190 | 752 | 284 | 324 | 100.000 |
| 7 | 30 | 25 | 1 | 12 | 5 | 6 | 12 | 79.000 | 66.440 | 33.220 | 1914 | 301 | 347 | 99.667 |
| 8 | 30 | 15 | 3 | 4 | 15 | 6 | 12 | 75.000 | 70.670 | 35.330 | 972 | 304 | 354 | 99.167 |
| 9 | 10 | 0 | 4 | 6 | 10 | 4 | 24 | 34.167 | 69.620 | 11.600 | 1527 | 250 | 365 | 97.167 |
| 10 | 30 | 20 | 2 | 6 | 10 | 6 | 12 | 86.000 | 77.250 | 38.620 | 1035 | 334 | 374 | 100.000 |
| 11 | 12 | 0 | 4 | 5 | 12 | 4 | 20 | 43.000 | 94.000 | 18.800 | 711 | 287 | 378 | 99.667 |
| 12 | 30 | 0 | 6 | 2 | 30 | 6 | 12 | 81.333 | 75.610 | 37.810 | 2088 | 348 | 395 | 100.000 |
| 13 | 6 | 0 | 4 | 10 | 6 | 4 | 40 | 10.833 | 85.920 | 8.590 | 764 | 292 | 523 | 98.667 |
| 14 | 5 | 0 | 4 | 12 | 5 | 4 | 48 | 8.167 | 83.330 | 6.940 | 856 | 348 | 618 | 94.833 |
| 15 | 22 | 20 | 1 | 30 | 2 | 11 | 30 | 91.500 | 79.440 | 29.130 | 3618 | 641 | 678 | 99.833 |
| 16 | 6 | 0 | 6 | 10 | 6 | 6 | 60 | 23.333 | 73.970 | 7.400 | 885 | 434 | 716 | 99.333 |
| 17 | 15 | 12 | 2 | 20 | 3 | 10 | 40 | 77.000 | 89.120 | 22.280 | 3279 | 676 | 746 | 100.000 |
| 18 | 30 | 20 | 4 | 6 | 10 | 12 | 24 | 91.333 | 71.290 | 35.650 | 3807 | 800 | 854 | 99.500 |
| 19 | 24 | 20 | 2 | 15 | 4 | 12 | 30 | 89.833 | 75.620 | 30.250 | 2015 | 840 | 895 | 99.333 |
| 20 | 30 | 25 | 2 | 12 | 5 | 12 | 24 | 97.333 | 84.580 | 42.290 | 3240 | 861 | 896 | 99.833 |
| 21 | 30 | 0 | 12 | 2 | 30 | 12 | 24 | 87.000 | 85.240 | 42.620 | 2820 | 865 | 913 | 100.000 |
| 22 | 30 | 15 | 6 | 4 | 15 | 12 | 24 | 92.667 | 91.180 | 45.590 | 3765 | 1002 | 1044 | 100.000 |
| 23 | 15 | 0 | 18 | 4 | 15 | 18 | 72 | 86.667 | 79.120 | 19.780 | 4980 | 1098 | 1189 | 99.500 |
| 24 | 12 | 0 | 18 | 5 | 12 | 18 | 90 | 76.500 | 92.170 | 18.430 | 5960 | 1293 | 1387 | 99.833 |
| 25 | 30 | 20 | 5 | 6 | 10 | 15 | 30 | 95.667 | 85.800 | 42.900 | 1900 | 1372 | 1409 | 99.833 |
| 26 | 25 | 24 | 1 | 60 | 1 | 25 | 60 | 95.833 | 74.290 | 30.950 | 4577 | 1379 | 1445 | 99.500 |
| 27 | 30 | 28 | 2 | 30 | 2 | 30 | 60 | 95.333 | 63.090 | 31.540 | 4380 | 1408 | 1496 | 100.000 |
| 28 | 30 | 24 | 6 | 10 | 6 | 30 | 60 | 97.833 | 65.020 | 32.510 | 4160 | 1453 | 1527 | 100.000 |
| 29 | 30 | 27 | 3 | 20 | 3 | 30 | 60 | 97.167 | 65.330 | 32.670 | 4160 | 1472 | 1535 | 100.000 |
| 30 | 30 | 25 | 5 | 12 | 5 | 30 | 60 | 97.667 | 67.650 | 33.830 | 4100 | 1491 | 1566 | 100.000 |
| 31 | 30 | 20 | 10 | 6 | 10 | 30 | 60 | 96.500 | 68.390 | 34.200 | 4410 | 1526 | 1591 | 100.000 |
| 32 | 30 | 0 | 18 | 2 | 30 | 18 | 36 | 89.333 | 89.990 | 45.000 | 1776 | 1631 | 1683 | 100.000 |
| 33 | 20 | 0 | 18 | 3 | 20 | 18 | 54 | 84.167 | 87.120 | 29.040 | 2214 | 1631 | 1699 | 99.833 |
| 34 | 25 | 20 | 6 | 12 | 5 | 30 | 72 | 96.000 | 74.870 | 31.190 | 5810 | 1665 | 1714 | 100.000 |
| 35 | 24 | 20 | 5 | 15 | 4 | 30 | 75 | 97.500 | 85.300 | 34.120 | 10851 | 1898 | 1939 | 99.667 |
| 36 | 20 | 0 | 24 | 3 | 20 | 24 | 72 | 89.333 | 83.550 | 27.850 | 4320 | 1920 | 2003 | 100.000 |
| 37 | 36 | 24 | 12 | 5 | 12 | 36 | 60 | 98.833 | 64.170 | 38.500 | 6615 | 2135 | 2218 | 100.000 |
| 38 | 30 | 15 | 15 | 4 | 15 | 30 | 60 | 96.833 | 72.360 | 36.180 | 5172 | 2213 | 2286 | 100.000 |
| 39 | 26 | 24 | 2 | 30 | 2 | 26 | 60 | 96.833 | 86.300 | 37.400 | 5178 | 2252 | 2302 | 100.000 |
| 40 | 28 | 24 | 4 | 15 | 4 | 28 | 60 | 97.167 | 79.050 | 36.890 | 5429 | 2314 | 2370 | 100.000 |
| 41 | 30 | 0 | 30 | 2 | 30 | 30 | 60 | 97.333 | 84.390 | 42.200 | 5063 | 2365 | 2413 | 99.500 |
| 42 | 27 | 24 | 3 | 20 | 3 | 27 | 60 | 96.833 | 89.980 | 40.490 | 5398 | 2442 | 2487 | 100.000 |
| 43 | 30 | 0 | 36 | 2 | 30 | 36 | 72 | 96.333 | 85.090 | 42.550 | 2952 | 2762 | 2799 | 100.000 |
| Mean | | | | | | | | 75.798 | 77.517 | 30.253 | | | | 99.461 |
| Standard Deviation | | | | | | | | 28.700 | 9.990 | 10.750 | | | | 1.127 |

### *6.2.3 A Disadvantage*

The {System B} committee is mainly based on LVQ component classifiers that have whole images of faces as inputs. Since the system is not based on feature extraction, the nature of the training inputs must honestly represent the input space. If this condition is not present, the classification will fail when the input to be classified is not close enough to any of the training inputs. To illustrate this idea the ORL face database is used to train a {System B} committee. The experiment is conducted on M = 40 classes with m = 20, V = 15 and L = 1. Although the {System B} committee recognized 99.5% of the training inputs correctly, it only classified 91.5% of the testing inputs correctly. This is because the ORL database contains much variability between the training and testing images for the committee to handle (figure 6.3). On the other hand, the training and testing sets of the Essex database are near to each other. This explains the high performance of the {System B} committee in classifying the testing inputs as can be seen from table 6.5 and figure 6.4. This problem could be reduced if suitable feature extraction is applied to the input images before presented to the LVQ classifiers.

### *6.2.4 Classification Capabilities*

The last experiment illustrates the ability of the {System B} committee to handle a large number of classes at once. The experiment is conducted on 392 classes taken from the 'faces94', 'faces95', 'faces96' and 'grimace' datasets. The parameters of the committee were as follows: M = 392, m = 28, V = 21 and L = 2. The total number of weights is 9336 weights. The committee correctly classified 95.153% of the testing set (3730 from 3920 images). Each testing input required in the average 10252 comparisons with the committee's weights. A single face takes about 5 seconds to be classified on a 733MHz PIII machine with 256 MBytes RAM. The training process took about 8 hours on the same machine.

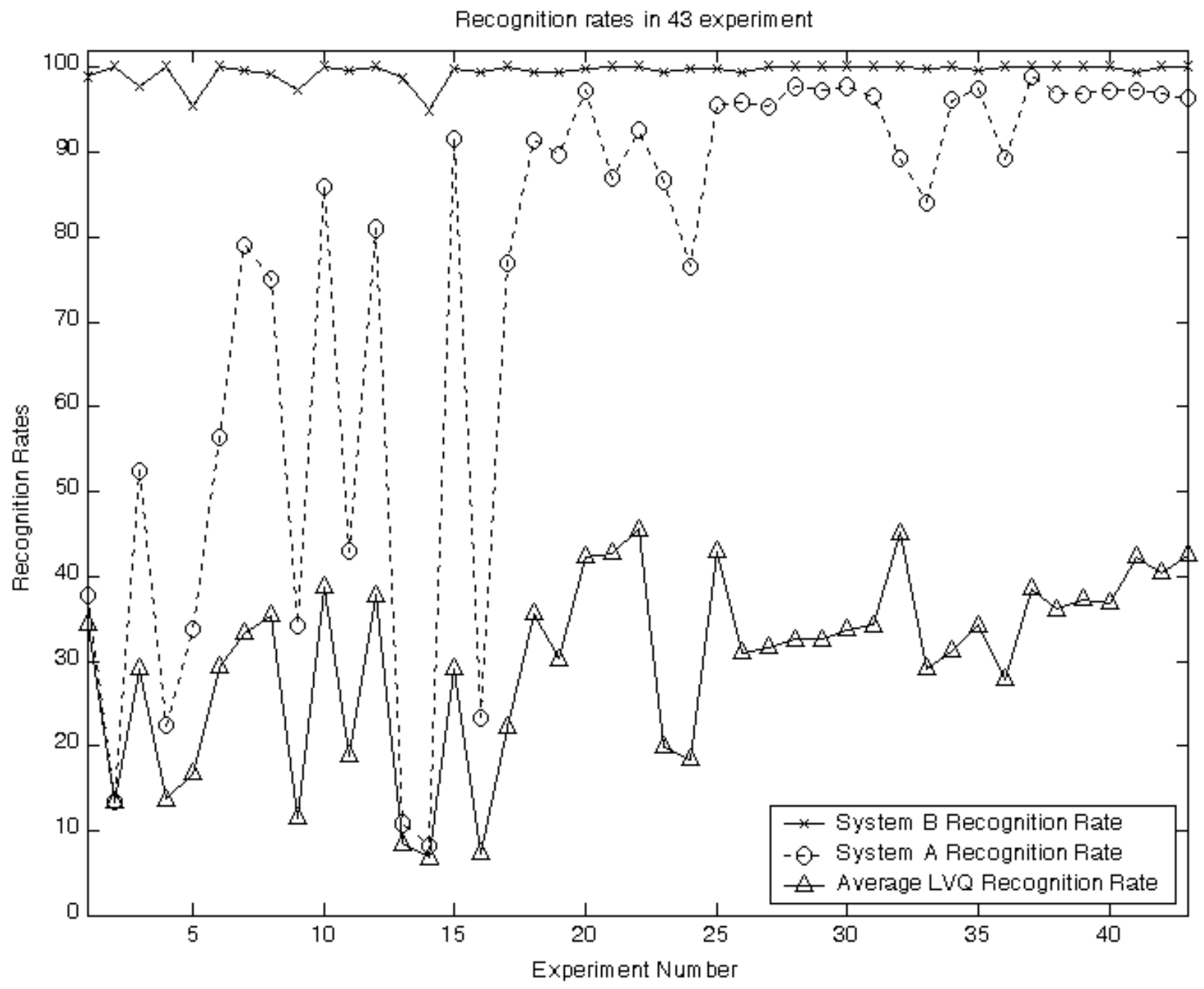

Figure 6.4. Recognition Rates of {System A} committees and their base LVQ nets during 43 experiments

### *6.3 Discussion of Results*

The results of section 6.2 reveal a lot about the properties of the system. For example, table 6.5 illustrates the main advantages of using combined systems. Column C01 and C07 are the recognition rates of {System A} and {System B} committees respectively (figure 6.4). The figure shows that although the {System A} committee uses simple voting for decision combination, its recognition rate is higher than the average LVQ recognition rates given in column C03. This confirms the claim that combining classifiers could enhance the recognition rate. On the other hand, the instability of the LVQ base classifiers makes the instability of the resulting {System A} committee worse as seen in the larger swings in figure 6.4. This confirms the claim that problems in the base classifiers could be multiplied in the combined system. In what follows a discussion of the results is provided to illustrate the important properties of the proposed combined system.

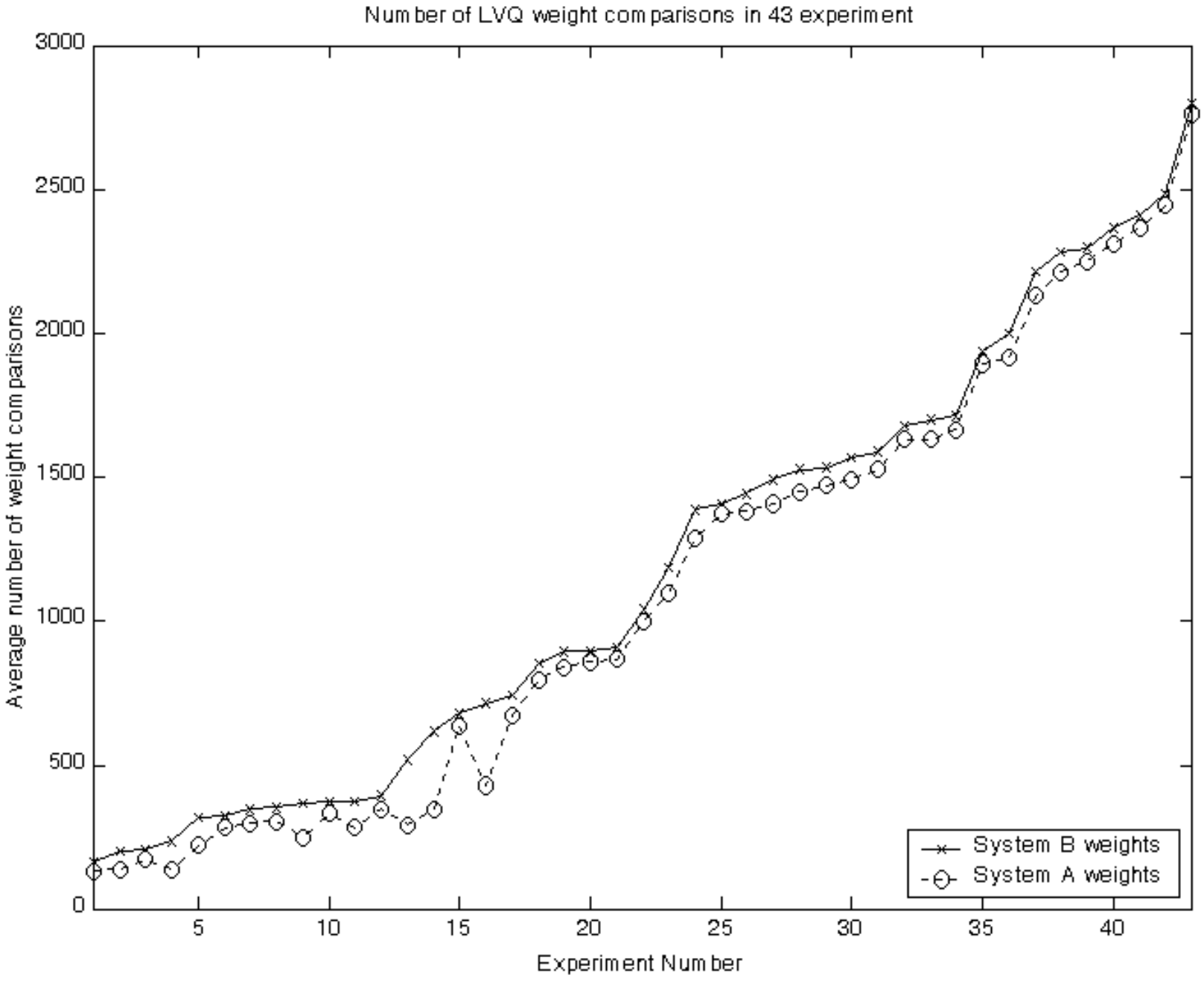

Figure 6.5. Average number of LVQ weight comparisons for a single input presented to {System B} and {System A} committees during 43 experiments

### *6.3.1 The Objectives*

The results show that the main objectives of this work are fulfilled. The first objective is to design a combined system capable of performing face recognition with high recognition rate. The results shown in sections 6.2.2, 6.2.3, and 6.2.4 and figure 6.4 confirm the system's ability to perform such task on the Essex and ORL face databases. The second objective is to design a combined system that requires the minimum number of parameters to be selected before training. The proposed system requires three parameters to be selected: m, V, and L.

Table 6.5 and figures 6.4 and 6.5 illustrate the effects of these three parameters on the performance of the resulting system. The recognition rate of the {System B} committee is almost unaffected by these parameters as seen from figure 6.4. The true effect is on the space and time requirements of the resulting system as seen from figure 6.5. When these parameters result in a high number of classifiers (column R in table 6.5), the resulting system requires the storage of a high number of LVQ weights (columns C04 and C05 in table 6.5 and figure 6.5). In addition, the average number of weight comparisons to any input image is also increased (column C06 in table 6.5 and figure 6.5) thus increasing training and classification time requirements. The third objective is to illustrate the possibility of designing a stable combined system based on unstable base classifiers. A look at table 6.4 and column C02 of table 6.5 and figure 6.4 reveals the instability of LVQ classifiers. The standard deviation of the recognition rates of LVQ classifiers is 12.596 in table 6.4 and is 9.99 in table 6.5. Comparing that to the {System B} committees in tables 6.1, 6.2, and 6.3 and column C07 of table 6.5 reveals the stability advantage of {System B} committee although it is originally based on unstable LVQ classifiers. The stability of the {System B} committee vs. the instability of the base LVQ classifiers are also apparent in figure 6.4

### *6.3.2 Space and Time Requirements*

Storage and time requirements of the proposed system are not measured in physical units (like bytes and seconds). Such absolute measurements are generally misleading. Perhaps another optimized implementation of this same algorithm on a faster computer with more memory will produce very different measurements of time and space. Nevertheless, a rough estimation of time and space requirements can be obtained from the number of LVQ weights generated during training or used during classification (columns C04, C05, and C06 in table 6.5 and figures 6.5 and 6.6). For instance, a comparison of columns C04 and C05 in table 6.5 (figure 6.6) highlights the effectiveness of classifier pruning. A large reduction in the number of weights is desirable to reduce classification time and space requirements. As pointed out in section 6.3.1, the values of the three system parameters (m, L, and V) largely affect time and space requirements since they affect the total number of generated LVQ nets and hence LVQ weights. As illustrated in section 6.3.3, the {System B} committee is generally more efficient in space and time than the {System A} committee. A high recognition rate obtained by the {System A} committee can be obtained by the {System B} committee using a reduced number of weights.

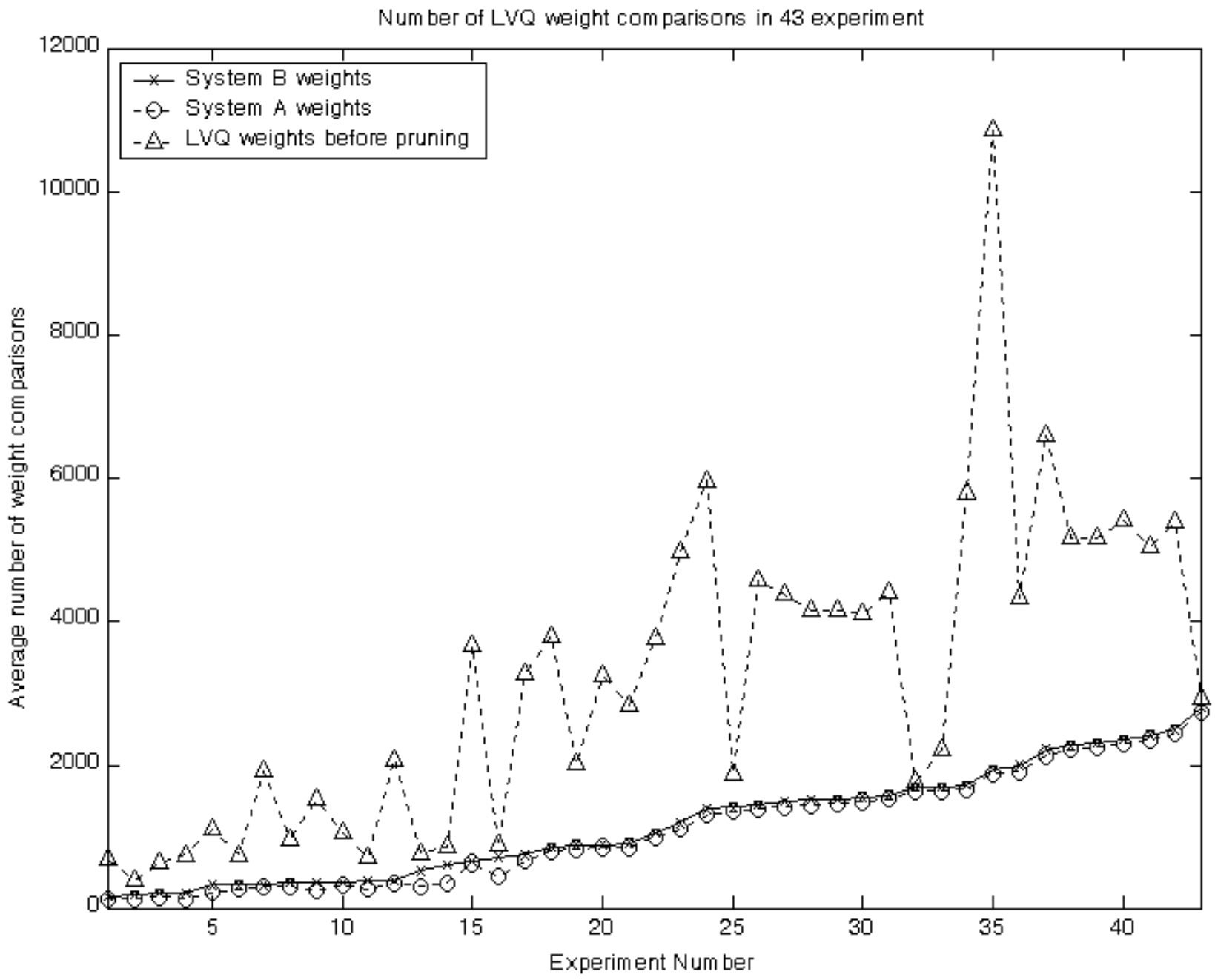

Figure 6.6. Effect of pruning on total number of LVQ weights

### *6.3.3 Comparing {System A} with {System B}*

Each LVQ classifier alone can correctly classify a small percentage from the overall dataset (column C03 in table 6.5 and figure

6.4). When combining these same classifiers using plurality voting in a {System A} committee, the recognition rate increases in many cases. This increase might not be sufficient to satisfy the desired recognition rate as seen from figure 6.4. This is where the FEC of {System B} appears to be very important. First, it requires no training as it depends on the LVQ weights already present in the {System A} committee. Second, it closes the large gap between the {System A} recognition rate and the desired recognition rate (columns C01 and C07 in table 6.5 and figure 6.4). In addition, {System B} committee is much more efficient during classification in both time and storage space. For example, the maximum recognition rate of {System A} is 98.833% in configuration 37 with 2135 LVQ weights in the committee (columns C01 and C05, row 37). This same recognition rate is reached by {System B} in configuration 1 using only 170 weights (columns C06 and C07, row 1). It should be noted that as any input is classified it is compared with all the weights of the {System A} committee and compared only to a portion of the weights of the FEC of the {System B} committee. The stability of the {System A} committee is very low compared to the stability of the {System B} committee as seen from the standard deviation of columns C01 and C07 and the swings in figure 6.4. It appears that voting is too simple to be used as a decision combination strategy for a complex problem as face recognition.

### *6.3.4 Comparisons with Other Systems*

In order to fairly compare this work to other combined systems one of two approaches should be taken. The first is to use the same face database used with other systems to evaluate the proposed system. This approach is only possible with the ORL database because no reported combined system that uses the Essex database was found. Other databases are too small, too simple, or too expensive to be used in the proposed system. In addition, many research groups construct their own database of faces. This leads to the difficulty of using the first approach. The second approach is to actually implement the reported work and test it using the Essex and ORL databases. Unfortunately, not all details are given in the papers and most of the systems are either unstable enough or require special types of input images (like [37] and [23]) to be fairly tested. Due to the previous difficulties, in what follows is a rough comparison with the combined systems presented in section 2.2.

In general, the main disadvantages of these systems are instability, design complexity, or low recognition rates. Compared to the system of [24], the proposed algorithm is much less in its recognition rate on the ORL database. Nevertheless, the recognition rate reported in [24] is unrealistic since it is based on images used during training despite called testing images. The system of [24] is very complex compared to the {System B} committee because the base RBF and LVQ classifiers require special parameters selection by trial and error to reach high recognition rates. This type of complexity cancels the benefits of its high recognition rate even if it is 99.5% as reported. The systems of [23] and [37] are tested on small face databases. Their reported recognition rates are therefore unreliable to assess their true strengths. Most of the algorithms in section 2.2 rely on base classifiers of different natures. In this system, only a single type of classifiers is used, the LVQ network. This simplifies greatly the design process of the system in several ways. First, the implementation process is unified. Second, the input images are conditioned only once to be used in exactly the same manner by all base classifiers. In addition, the decision combination strategy is simplified because all the bases classifiers produce the same type of response. Comparing the proposed system to boosting techniques in [36] and [35], it is apparent that the proposed system is suitable for parallel processing techniques. The system can be trained or used through a multi-processor machine or a computer networks to accelerate training and classification. This is because the base classifiers are trained and used independently from each other. In boosting, on the other hand, each new classifier is trained depending on the errors made by the previous classifiers. Hence it must be implemented on a serial machine not a parallel one. Another problem with some of the other systems is the system size problem. Some of these systems scale poorly with the increased number of classes. For instance, the system in [35] converts a C class recognition problem into a C (C – 1) / 2 two-class classification problems. If C equals 392, the system needs to solve 392 x 391 / 2 = 76636 two-class problems. Another example is the system in [37] where the size of the neural network heavily depends on the number of classes. The experiment in section 6.2.4 shows that the proposed system scales well to large classification problems. The system requires 10252 comparisons with LVQ weights in the average to classify a single input as belonging to one of 392 classes with a recognition rate around 95%. Finally, comparing the recognition rates of the proposed system in various experiments with the systems of section 2.2, the {System B} committee is similar to or better than most of these other systems.

### *6.3.5 Additional Observations*

The following observations are not directly related to face recognition or combined classifiers systems. They are more related to the field of neural networks. The disadvantages of using LVQ classifiers in pattern recognition problems are apparent. First, LVQ networks require the selection of many parameters: number of hidden units, number of training epochs, and learning rate. Referring to the LVQ training algorithm in section 5.2, each input is compared to all LVQ weights several times during training. For a classification problem with high input dimensionality and large number of classes, this training algorithm is impractical. In addition, the discussion of section 5.3.1 shows that not all LVQ weights are useful during classification. Even when such disadvantages are tolerated, the result may be undesirable because of the LVQ network instability that may produce low recognition rates. The {System B} committee can be considered an alternative training algorithm for LVQ networks. It uses less parameters, uses much efficient training strategy, produces a stable system, and only keeps necessary weights if pruning is used. From the neural networks point of view, the {System B} algorithm is better in training and applying LVQ neural networks than the traditional LVQ training algorithm of section 5.2.

### *6.4 Conclusions*

The system described in this work is capable of performing face recognition tasks with high performance. The ideas behind the system are simple ones. No feature extraction is required and the system performance is stable against its parameters. The system is suitable to be applied in applications such as human-computer interaction. Usually a person sits in front of his/her terminal in a rather fixed position with fixed artificial lighting. A few pictures for each user can be used to train the computer to recognize the person sitting in front of it. Other more difficult applications require many images to train the system to avoid the disadvantage described in section 6.2.3. The {System B} committee satisfies the main objectives of this thesis and compares generally well to other combined systems.

*Chapter 7*

# Conclusions and Future Work

### *7.1 Conclusions*

In this work, an algorithm for training a number of unstable classifiers to perform face recognition has been presented. The classifiers are LVQ neural networks and their decisions are combined using another specially designed LVQ neural network. This algorithm can be investigated from several angles. The first angle is from the field of multi-classifier systems point of view. Although the {System B} committee is constructed entirely using LVQ neural networks, this is not a restriction. Any type of classifiers can be used as the component classifier. The FEC LVQ neural net could be also replaced by any similar classifier. The only condition is that the FEC classifier must be capable of accepting and using the controlling inputs from the previous stage. This leads to the ability of considering the {System B} committee as a general scheme for constructing committee machines. It is seen in chapter 4 that the decision combination scheme is very important to achieve a high recognition rate. The FEC is capable of performing this combination with noticeable performance and design improvements. One performance improvement is making the overall system very stable. Another improvement is increasing the recognition rate of the system to very high levels (compared to voting for example). An important design improvement is reducing the designer’s need to select the best method for combining the committee’s decisions for a certain application.

Another conclusion is that a committee machine could already contain all the necessary information to reach a high recognition rate. Due to bad organization of the committee, the actual performance may be less than expected. This can be seen from that fact that the only difference between

{System A} and {System B} committees is in the FEC. This FEC is actually constructed from the information content of the {System A} weights. Nothing new is added to the {System A} committee to become a more powerful {System B} committee. The only new thing is information reorganization. This indicates the need for more attention to the design of good combination strategies. Another conclusion is that the cooperation of simple classifiers could perform much better than a single complex classifier. In addition, the stability of the {System B} committee suggests that there is little need to search for optimal system parameters and initial conditions. Combining classifiers in this way can produce reliable classification systems more easily. This would make it more practical to study the behavior of combined classifiers in many practical problems without having to worry about exact values of parameters or the initial state of the system before training.

The second angle is seeing the system as a face recognition system. The system is based on neural networks that accept whole images as inputs. This approach results in the disadvantage of section 6.2.3. The training images must be good representations of the actual images the system will handle after training. Thus, the system cannot be used as a high-accuracy face recognition system. In addition, the system cannot be reliably used within uncontrolled environments. Nevertheless, the system is suitable for some important applications like human computer interaction in multimedia applications. In such applications, usually, a person sits in front of his/her terminal in a rather fixed position with fixed artificial lighting. A few pictures for each user can be used to train the computer to recognize the person sitting in front of it. Other more difficult applications require many images to train the system to avoid the disadvantage described in section 6.2.3.

The third angle is regarding the {System B} committee as a huge LVQ classifier. The result of constructing a {System B} committee can be viewed as a large number of LVQ weights trained on the input patterns. If one tries to train the same number of weights using the conventional LVQ algorithm given in section 5.2, the required time will be much larger. This is because each new input pattern is compared to all LVQ weights in the conventional algorithm. On the other hand, the same input pattern will be compared to a small portion of the weights of the {System B} committee. Hence, the {System B} construction algorithm can be considered as a modification to the basic LVQ neural network training algorithm. This modification is capable of producing a large and stable LVQ network in reasonable time given a large number of classes and/or input patterns.

The important conclusion is that even in machine learning, good cooperation will lead to success.

### *7.2 Future Work*

The {System B} committee can be considered a general scheme. Other classification methods could be used as component classifiers instead of the LVQ neural networks. In addition, feature extraction can be added to enhance system performance on difficult databases. The {System B} committee consists of two stages. The possibility of adding more stages to the system should be considered. If the FEC misclassifies a significant number of patterns then another FEC can be added to further reduce classification error while relying on the input pattern, and the output of the first FEC. The generalization capabilities of the system should be studied more carefully. For the system to be practical, it should generalize well and methods for enhancing its generalization should be investigated. The {System B} construction algorithm could be used as an alternative to the basic LVQ algorithm. Its properties, limitations, and possible applications as a neural network should be studied both theoretically and experimentally. The face recognition system should be applied to a real life application in a controlled environment (for example in computer labs). The effect of using other classifiers in the committee or the FEC on the performance should be studied. Finally, the system should be used in other application beside face recognition to further study its strengths and weaknesses as a general pattern recognition system.

# Bibliography

[1] P. Jonathon Phillips, Alvin Martin, C.L. Wilson, Mark Przybocki, "An Introduction to Evaluating Biometric Systems", Natural Institute of Standards and Technology, February 2000

[2] Diego A. Socolinsky, Lawrence B. Wolff, Joshua D. Neuheisel, and Christopher K. Eveland, "Illumination Invariant Face Recognition Using Thermal Infrared Imagery", Computer Vision and Pattern Recognition, Kauai, December 2001

[3] Charles Beumier and Marc Acheroy, "Automatic Face Authentication from 3D surface", British Machine Vision Conference BMVC 98, University of Southampton UK, 1417, pp 449-458, September 1998

[4] George Bebis, Satishkumar Uthiram, Michael Georgiopoulos, "Face Detection and Verification Using Genetic Search", International Journal on Artificial Intelligence Tools, Vol. 9, No. 2, pp. 225-246, 2000

[5] Tin Kam Ho, in A. Kandel, H. Bunke, (eds.), "Multiple Classifier Combination: Lessons and Next Steps", Hybrid Methods in Pattern Recognition , World Scientific, 171-198, 2002

[6] Matthew A. Turk and Alex P. Pentland, "Face Recognition Using Eigenfaces", Proc. of IEEE Conf. on Computer Vision and Pattern Recognition, pp. 586-591, June 1991

[7] Carlos Eduardo Thomaz Raul Queiroz Feitosa Alvaro Veiga, "Design of Radial Basis Function Networks as Classifier in Face Recognition Using Eigenfaces", In proceedings of the 5th Brazilian Symposium on Neural Networks December 09 - 11, 1998

[8] Jun Zhang, Yong Yan, and Martin Lades, "Face Recognition: Eigenface, Elastic Matching, and Neural Nets", Proceedings of the IEEE, vol. 85, no. 9, pp. 1423-1435, September 1997

[9] S. Eickeler, S. Muller, G. Rigoll, "Recognition of JPEG Compressed Face Images Based on Statistical Methods", Image and Vision Computing Journal, Special Issue on Facial Image Analysis, 18(4):279-287, March 2000

[10] Ara V. Nefian and Monson H. Hayes III, "Face detection and recognition using hidden markov models", in International Conference on Image Processing, vol. 1, pp. 141-145, October 1998

[11] Ara V. Nefian and Monson H. Hayes III, "Face recognition using an embedded HMM", In Proceedings of the Audio- and Video-Based Biometric Person Authentication (AVBPA'99), Washington DC, USA, 1999

[12] F. Samaria and F. Fallside, "Automated face identification using hidden markov models", in Proceedings of the International Conference on Advanced Mechatronics, pp. 1-9, 1993

[13] Peter N. Belhumeur, Joao P. Hespanha, and David J. Kriegman, "Eigenfaces vs. Fisherfaces: Recognition Using Class Specific Linear Projection", IEEE Transactions on Pattern Analysis and Machine Intelligence, Vol. 19, No. 7, July 1997

[14] A. Jonathan Howell and Hilary Buxton, "Face Recognition Using Radial Basis Function Neural Networks", Proc. British Machine Vision Conference, 455-464. 32, 1996

[15] Guodong Guo, Stan Z. Li, and Kapluk Chan, "Face Recognition by Support Vector Machines", Proc. of the International Conferences on Automatic Face and Gesture Recognitionz, 196-201, 2000

[16] Dominique Valentin, Herve Abdi, Alice J. O'Toole, Garrison W. Cottrell, "Connectionist Models of Face Processing: A Survey", Pattern Recognition, vol. 27, no. 9, pp. 1209-1230, 1994

[17] Laurenz Wiskott, Jean-Marc Fellous, Norbert Kruger, and Christoph von der Malsburg, “Face Recognition by Elastic Bunch Graph Matching”, IEEE Transaction on Pattern Analysis and Machine Intelligence, 19(7): 775 - 779, 1997

[18] Gita Sukthankar, "Face recognition: A critical look at biologically inspired approaches", Technical Report CMU-RI-TR-00-04, Robotics Institute, Carnegie Mellon University, January, 2000

[19] Jie Yang, Hua Yu, William Kunz, "An Efficient LDA Algorithm for Face Recognition", The Sixth International Conference on Control, Automation, Robotics and Vision (ICARCV2000)

[20] W. Zhao, N. Nandhakurnar, "Linear discriminant analysis of MPF for face recognition", Proc. International Conference on Pattern Recognition, Brisbane, Australia, pp. 185188, 1998

[21] W. Zhao R. Chellappa P. J. Phillips, "Subspace linear discriminant analysis for face recognition", Center for Automation Research, University of Maryland, College Park, Technical Report CAR-TR-914, 1999

[22] P. Jonathon Phillips, Hyeonjoon Moon, Syed A. Rizvi, and Patrick J. Rauss, “The FERET Evolution Methodology for Face-Recognition Algorithms”, U.S. Army Research Laboratory, Technical Report NISTIR 6264, January 7, 1999

[23] Bernard Achermann and Horst Bunke, "Combination of Face Classifiers for Person Identification", Proceedings of the 13th IAPR International Conference on Pattern Recognition (ICPR), Vienna, Austria, Vol. III, pages 416-420, August 1996

[24] A. S. Tolba, A. N. Abu-Rezq, "Combined Classifiers for Invariant Face Recognition", Pattern Analysis and Applications, Vol. 3, No. 4, pp 289-302, December 2000

[25] Bernd Heisele, Purdy Ho, Tomaso Poggio, "Face Recognition with Support Vector Machines: Global versus Component-based Approach", in Proc. ICCV, Vancouver, Canada, vol. 2, pp. 688-694, 2001

[26] K. Jonsson, J. Matas, J. Kittler and Y. P. Li, "Learning Support Vectors for Face Verification and Recognition", Fourth IEEE International Conference on Automatic Face and Gesture Recognition 2000, pages 208-213, Los Alamitos, USA, March 2000

[27] P. Jonathon Phillips, “Support Vector Machines Applied to Face Recognition”, Advances in Neural Information Processing Systems II (pp. 803-809), MIT Press, 1999

[28] WenYi Zhao and Rama Chellappa, “Illumination-Insensitive Face Recognition Using Symmetric Shape-from-Shading”, Computer Vision and Pattern Recognition (CVPR'00)-Volume 1 June 13 - 15, 2000

[29] W. Zhao and Rama Chellappa, “Robust face recognition using symmetric shape-from-shading”, Technical Report CAR-TR-919, Center for Automation Research, University of Maryland, 1999

[30] WenYi Zhao and Rama Chellappa, “SFS based view synthesis for robust face recognition”, in 4th IEEE Conference on Automatic Face and Gesture Recognition, pp. 285 -292, Grenoble, France, 2000

[31] Laurenz Wiskott and Christoph von der Malsburg, “Face Recognition by Dynamic Link Matching”, In Proceedings of the International Conference on Artificial Neural Networks, ICANN'95, (Paris), pp. 347-352, Oct. 1995

[32] P. Kruizinga, N. Petkov, “Optical flow applied to person identification”, Proceedings of the 1994 EUROSIM Conference on Massively Parallel Processing Applications and Development, Delft, the Netherlands, (Elsevier, Amsterdam, 1994), 21-23 June, pp.871-878, 1994

[33] Terence Sim, Rahul Sukthankar, Matthew D. Mullin, and Shumeet Baluja, "High-Performance Memory-based Face Recognition for Visitor Identification", Technical report, Just Research, January 1999

[34] S. Gutta, J. Huang, B. Takacs, and H. Wechsler, "Face Recognition Using Ensembles of Networks", Proc. of 13th International Conference on Pattern Recognition (ICPR) , Vienna, Austria, 1996

[35] Guo-Dong Guo, Hong-Jiang Zhang, and Stan Z. Li, "Pairwise Face Recognition", In Proceedings of 8th IEEE International Conference on Computer Vision. Vancouver, Canada. July 9-12, 2001

[36] Guo-Dong Guo and Hong-Jiang Zhang, "Boosting for Fast Face Recognition", IEEE ICCV Workshop on Recognition, Analysis, and Tracking of Faces and Gestures in Real-Time Systems (RATFG-RTS'01) , 2001

[37] Fu Jie Huang, Zhihua Zhou, Hong-Jiang Zhang, and Tsuhan Chen, "Pose Invariant Face Recognition", Proceedings of the 4th IEEE International Conference on Automatic Face and Gesture Recognition, pp.245-250, Grenoble, France, 2000

[38] Thomas Fromherz, Peter Stucki, Martin Bichsel, "A Survey of Face Recognition", MML Technical Report No. 97.01, Department of Computer Science, University of Zurich, 1997

[39] Internet site: “http://www.uk.research.att.com/facedatabase.html”, valid until 30/6/2004

[40] Richard O. Duda, Peter E. Hart and David G. Stork, "Pattern Classification, 2nd Edition", Wiley Interscience, 1997

[41] Yu Hen Hu and Jenq-Neng Hwang (eds.), "Committee Machines", Published as a book chapter in "Handbook for Neural Network Signal Processing", CRC Press, 2001

[42] G. Valentini, F. Masulli, In R. Tagliaferri and M. Marinaro, editors, "Ensembles of Learning Machines", Neural Nets WIRN Vietri-2002, Series Lecture Notes in Computer Sciences, Springer-Verlag, vol. 2486, pp. 3-19, 2002

[43] L. Lam and C. Sue, "Application of majority voting to pattern recognition: an analysis of its behavior and performance", IEEE Transactions on Systems, Man and Cybernetics, 27(5):553–568, 1997

[44] L Xu, C Krzyzak, and C. Suen, "Methods of combining multiple classifiers and their applications to handwriting recognition", IEEE Transactions on Systems, Man and Cybernetics, 22(3):418–435, 1992

[45] C. Suen and L. Lam, "Multiple classifier combination methodologies for different output levels", In Multiple Classifier Systems. First International Workshop, MCS 2000, Cagliari, Italy, volume 1857 of Lecture Notes in Computer Science, pages 52–66. Springer-Verlag, 2000

[46] Y.S. Huang and Suen. C.Y., "Combination of multiple experts for the recognition of unconstrained handwritten numerals", IEEE Trans. on Pattern Analysis and Machine Intelligence, 17:90–94, 1995

[47] S. Cho and J. Kim, "Combining multiple neural networks by fuzzy integral and robust classification", IEEE Transactions on Systems, Man and Cybernetics, 25:380–384, 1995

[48] S. Cho and J. Kim. "Multiple network fusion using fuzzy logic", IEEE Transactions on Neural Networks, 6:497–501, 1995

[49] J.M. Keller, P. Gader, H. Tahani, J. Chiang, and M. Mohamed, "Advances in fuzzy integration for pattern recognition", Fuzzy Sets and Systems, 65:273–283, 1994

[50] D. Wang, J.M. Keller, C.A. Carson, K.K. McAdoo-Edwards, and C.W. Bailey, "Use of fuzzy logic inspired features to improve bacterial recognition through classifier fusion", IEEE Transactions on Systems, Man and Cybernetics, 28B(4):583–591, 1998

[51] S. Impedovo and A. Salzo, "A New Evaluation Method for Expert Combination in Multi-expert System Designing", In J. Kittler and F. Roli, editors, Multiple Classifier Systems. First International Workshop, MCS 2000, Cagliari, Italy, volume 1857 of Lecture Notes in Computer Science, pages 230–239. Springer-Verlag, 2000

[52] R.P.W. Duin and D.M.J. Tax, "Experiments with Classifier Combination Rules", In J. Kittler and F. Roli, editors, Multiple Classifier Systems. First International Workshop, MCS 2000, Cagliari, Italy, volume 1857 of Lecture Notes in Computer Science, pages 16–29. Springer-Verlag, 2000

[53] Ludmila I. Kuncheva, "A Theoritical Study on Six Classifier Fusion Strategies", IEEE Transactions on Pattern Analysis and Machine Intelligence, Vol. 24, No. 2. February 2002

[54] Dietterich T.G. "Machine Learning Research: Four Current Directions", AI Magazine, 18(4):97-136, 1999

[55] Laurene Fausett, "Fundamentals of Neural Networks, Architectures, Algorithms, and Applications", Printice Hall International, 1994

[56] Internet site: "http://cswww.essex.ac.uk/mv/allfaces/index.html", valid until 30/6/2004

| | |
|---|---|
| صاحب الرسالة | أحمد حسني عوض عيد |
| عنوان الرسالة | المصنّفات المركَّبة للتعرف اللامتغير على الوجه |
| الكلية | الهندسة |
| القسم العلمي | الهندسة الكهربية |
| موقع الكلية | بورسعيد |
| الدرجة العلمية | الماجستير |
| تاريخ المنح | ٣٠ / ٦ / ٢٠٠٤ |
| لغة الرسالة | الإنجليزية |
| هيئة الإشراف | أ.د./ عبد الحي أحمد سلام<br>د./ كامل أحمد الصيرفي |

## الموجز العربي

لا يوجد مصنف يمكنه منفردا التعامل بكفاءة مع المسألة المعقدة للتعرف على الوجه، لقد وجد الباحثون أن تجميع مخرجات عدد من المصنفات الأساسية يحسن من معدل تعرفها، و تنعكس نقاط ضعف المصنفات الأساسية على النظام المركب الناتج، في هذه الرسالة يتم اقتراح نظام مركب مبني على مصنفات أساسية غير مستقرة و منخفضة في معدل تعرفها، يتم تطبيق النظام للتعرف على صور لوجوه ٣٩٢ شخصا، يظهر النظام المقترح استقرارا ملحوظا و معدلا عاليا للتعرف باستخدام عدد قليل من المتغيرات أثناء التصميم، يوضح النظام إمكانية تصميم نظام مصنفات مركب قادر على الانتفاع بنقاط القوة في مصنفاته الأساسية مع اجتناب نقاط ضعفها.

# ملخص الرسالة

من بين كل مسائل التعرف على الأنماط تعتبر مسألة التعرف على الوجه واحدة من أصعبها، الطبيعة الخاصة لتلك المسألة تطلبت من الباحثين استقصاء الكثير من طرق التصنيف لحلها، لا يوجد مصنف باستطاعته منفردا الوصول لأداء جيد في كل الأشكال المختلفة لتطبيقات التعرف على الوجه، الإثبات الوحيد لوجود مثل هذا المصنف هو المخ البشري بإمكانياته الهائلة في التعرف على الأنماط، منذ ظهور فكرة التجميع المركب للمصنفات و قد أطلقت عددا ضخما من المحاولات لتطبيقها في العديد من مسائل التعرف على الأنماط، تعتمد هذه التقنية على تدريب عدد من المصنفات الأساسية بشكل منفصل على حل المشكلة، ثم يتم تجميع مخرجات المصنفات الأساسية بطريقة تجميع ما، بالرغم من عدم وجود نظام مركب قادر على إزالة كل صعوبات التعرف على الوجه إلا أن تقنية المصنفات المركبة أثبتت امتلاكها لصفات مثيرة للاهتمام قد تؤدي في النهاية إلى إزالة الكثير من معوقات الحصول على حل جيد لمثل هذه المسألة المعقدة، بالرغم من ذلك تنعكس النقائص الموجودة بالمصنفات الأساسية سلبا على النظام المركب الناتج من تجميعها، المصنفات الأساسية قليلة الاستقرار ذات التصميم المعقد و معدل التعرف المنخفض تؤثر سلبا في استقرار و تعقيد و أداء النظام المركب الناتج من تجميعها، في هذه الرسالة يتم اقتراح نظام مصنفات مركبة مبني على مصنفات أساسية غير مستقرة و منخفضة في معدل تعرفها، يتم تطبيق النظام للتعرف على قاعدة بيانات للوجوه تتكون من ٣٩٢ شخصا، يظهر النظام المقترح استقرارا ملحوظا و معدلا عاليا للتعرف باستخدام عدد قليل من المتغيرات أثناء التصميم، يظهر النظام المقترح أداء أفضل من معظم نظم المصنفات المركبة الموضحة في الأبحاث الأخرى من ناحية بساطته و استقراره و معدل تعرفه و استجابته لزيادة حجم المدخلات، يمكن تطبيق النظام المقترح على حاسب شخصي عادي و هو يناسب تطبيقات

**الوسائط المتعددة المعتمدة على التفاعل بين المستخدم البشري و جهاز الحاسب، كما يمكن تطبيق النظام باستخدام طرق المعالجة المتوازية للتعرف على عدد ضخم من الأشخاص. يوضح هذا البحث إمكانية تصميم نظام مصنفات مركب قادر على الانتفاع بنقاط القوة في مصنفاته الأساسية مع اجتناب نقاط ضعفها.**

**و تتكون الرسالة من سبعة فصول مرتبة كما يلي:**

**الفصل الأول: مقدمة لشرح المفاهيم الأساسية و التطبيقات العملية لمسألة التعرف على الوجه. و في هذا الفصل أيضا يتم عرض المفهوم الأساسي لتجميع المصنفات مع بيان مميزات هذه التقنية و معوقات استخدامها، كما يتم عرض المشكلة المطلوب حلها في البحث و الأهداف المطلوب تحقيقها من وراء البحث بالإضافة إلى محتويات و ترتيب الرسالة.**

**الفصل الثاني: يعرض الفصل الثاني البحوث السابقة في مجال التعرف على الوجه سواء باستخدام مصنف فردي أو نظام مصنفات مركبة، و يوضح هذا الفصل بشكل عام عيوب و مميزات نظم التصنيف المعتمدة على مصنف فردي في مجال التعرف على الوجه، ثم ينتقل بعد ذلك إلى عرض البحوث المعتمدة على نظم التصنيف المركبة للتعرف على الوجه مع التركيز على بيان العيوب الظاهرة بكل منها.**

**الفصل الثالث: و يعرض المفاهيم و المصطلحات الأساسية في مجال التعرف على الأنماط، فمسألة التعرف على الوجه تعتبر حالة خاصة من مسائل التعرف على الأنماط، و لذلك فمعظم الصعوبات و المشاكل المتفرعة من مجال التعرف على الأنماط ستواجه أيضا الباحثين في مجال**

التعرف على الوجه، بالإضافة إلى أن النظام المركب المقترح في هذه الرسالة قد يصلح للتطبيق في مسائل أخرى للتعرف على الأنماط بجانب مسألة التعرف على الوجه.

**الفصل الرابع:** يهدف إلى استعراض الطرق المختلفة لتصميم نظم المصنفات المركبة، و يوضح أيضا مميزات تطبيق النظم المركبة في حل مسائل التعرف على الأنماط، و يبين بعض الأسباب التي تؤدي إلى ظهور هذه المميزات، و يستعرض بشكل مجمل بعض المسائل المفتوحة للبحث في تصميم نظم المصنفات المركبة للتعرف على الأنماط.

**الفصل الخامس:** يعرض بشكل مفصل تصميم النظام المركب المقترح للتعرف على الوجه، و يبين طريقة عمل النظام المقترح أثناء مرحلتي التدريب و التصنيف، و يوضح الفصل أيضا تصميم و خواص المصنفات الأساسية مع بيان نقاط قوتها و ضعفها، و يشرح الطرق المستخدمة في تجميع مخرجاتها، و يقارن التصميم العام للنظام المقترح بالنظم الأخرى المعروضة في الفصل الرابع.

**الفصل السادس:** الهدف منه عرض التجارب المستخدمة في اختبار النظام المقترح و التعرف على خصائصه، و يبدأ باستعراض البرمجيات المستخدمة في تطبيق النظام المقترح عمليا و قواعد البيانات المستخدمة في إجراء التجارب، ثم يوضح الفصل الغرض من كل مجموعة من التجارب و يعرض نتائجها، بعد ذلك يعرض الفصل مناقشة النتائج من ناحية تحقيق أهداف البحث و توضيح مميزات و عيوب النظام المقترح مقارنة بالنظم المعروضة في الفصل الثاني.

**الفصل السابع:** يعرض ما تم استخلاصه من البحث و يوضح بشكل عام ما يمكن عمله لتطوير و استخدام النظام المقترح فيما بعد.

جامعة قناة السويس

كلية الهندسة ببورسعيد

# *المصنّفات المركّبة للتعرف اللامتغير على الوجه*

رسالة مقدمة من

المهندس/ أحمد حسني عوض

بكالوريوس الهندسة الكهربية، ١٩٩٩ شعبة الحاسب الآلي

كلية الهندسة، جامعة قناة السويس

للحصول على

درجة الماجستير في الهندسة الكهربية

## أجيزت بواسطة

أ. د./ هاني سليم جرجس

أستاذ هندسة الحاسبات و الإلكترونيات

كلية الهندسة

جامعة أسيوط

أ. د./ ناضر حمدي محمد علي

أستاذ و رئيس قسم الإلكترونيات

بالأكاديمية العربية للعلوم و التكنولوجيا

و النقل البحري بالإسكندرية

أ. د./ عبد الحى أحمد سلام

أستاذ القوى الكهربية بقسم الهندسة الكهربية

كلية الهندسة، جامعة قناة السويس

٢٠٠٤

جامعة قناة السويس
كلية الهندسة ببورسعيد

# المصنّفات المركّبة للتعرف اللامتغير على الوجه

رسالة مقدمة من
المهندس/ أحمد حسني عوض
بكالوريوس الهندسة الكهربية، ١٩٩٩
شعبة الحاسب الآلي
كلية الهندسة، جامعة قناة السويس

للحصول على
درجة الماجستير في الهندسة الكهربية

## تحت إشراف

أ. د./ عبد الحى أحمد سلام
أستاذ القوى الكهربية بقسم الهندسة الكهربية
كلية الهندسة، جامعة قناة السويس

د./ كامل أحمد الصيرفي
أستاذ مساعد بقسم الهندسة الكهربية
كلية الهندسة، جامعة قناة السويس